\documentclass[preprint]{elsarticle}
\pdfoutput=1 
\usepackage{comment}

\usepackage{lineno,hyperref}
\usepackage{framed,multirow}
\usepackage{epstopdf}
\usepackage{amssymb}
\usepackage{latexsym}
\usepackage{makeidx}
\usepackage{graphicx}
\usepackage{caption}

\usepackage{subfigure}
\usepackage{epstopdf}
\usepackage{url}
\usepackage[utf8]{inputenc}
\usepackage{amsmath}
    \usepackage{amsfonts}
    \usepackage{amssymb}
\usepackage{color, colortbl}
\graphicspath {{figure/}}
\modulolinenumbers[5]

\usepackage{booktabs} 

\usepackage[ruled,vlined,linesnumbered]{algorithm2e}
\SetAlFnt{\small}
\SetAlCapFnt{\small}
\SetAlCapNameFnt{\small}
\SetAlCapHSkip{0pt}
\IncMargin{-\parindent}

\usepackage{lineno,hyperref}
\usepackage{framed,multirow}
\usepackage{epstopdf}

\usepackage{amssymb}
\usepackage{latexsym}
\usepackage{makeidx}
\usepackage{graphicx}
\usepackage{caption}
\usepackage{epstopdf}
\usepackage{url}

\usepackage{amsmath}
\usepackage{amsfonts}
\usepackage{color, colortbl}
\journal{Pattern Recognition}
\usepackage{etoolbox}

\makeatletter
\def\ps@pprintTitle{%
   \let\@oddhead\@empty
   \let\@evenhead\@empty
   \let\@oddfoot\@empty
   \let\@evenfoot\@oddfoot
}
\makeatother

\begin{document}

\begin{frontmatter}

\title{A novel classification-selection approach for the self updating of template-based face recognition systems}

\author{Giulia Orrù\corref{mycorrespondingauthor}}
\ead{giulia.orru@unica.it}
\cortext[mycorrespondingauthor]{Corresponding author}

\author{Gian Luca Marcialis\corref{cor2}}
\ead{marcialis@unica.it}
\author{Fabio Roli\corref{cor2}}
\ead{roli@unica.it}
\address{University of Cagliari, DIEE, Piazza d'Armi, Cagliari, Italy}

\begin{abstract}
\small
The boosting on the need of security notably increased the amount of possible facial recognition applications, especially due to the success of the Internet of Things (IoT) paradigm. However, although handcrafted and deep learning-inspired facial features reached a significant level of compactness and expressive power, the facial recognition performance still suffers from intra-class variations such as ageing, facial expressions, lighting changes, and pose.
These variations cannot be captured in a single acquisition and require multiple acquisitions of long duration, which are expensive and need a high level of collaboration from the users.
Among others, self-update algorithms have been proposed in order to mitigate these problems. Self-updating aims to add novel templates to the users' gallery among the inputs submitted during system operations. Consequently, computational complexity and storage space tend to be among the critical requirements of these algorithms. 
The present paper deals with the above problems by a novel template-based self-update algorithm, able to keep over time the expressive power of a \textit{limited} set of templates stored in the system database. The rationale behind the proposed approach is in the working hypothesis that  a dominating mode characterises the features' distribution given the client. Therefore, the key point is to select the best templates around that mode. We propose two methods, which are tested on systems based on handcrafted features and deep-learning-inspired autoencoders at the state-of-the-art. Three benchmark data sets are used. Experimental results confirm that, by effective and compact feature sets which can support our working hypothesis, the proposed classification-selection approaches overcome the problem of manual updating and, in case, stringent computational requirements.
\end{abstract}

\end{frontmatter}

{\small This is an original manuscript of an article published by Elsevier in Pattern Recognition on 27 November 2019\\
Available online: $https://doi.org/10.1016/j.patcog.2019.107121$}

\section{Introduction}
Face recognition (FR) is one of the most challenging tasks in pattern recognition due to intra-class and inter-class variations\footnote{https://www.bayometric.com/biometrics-face-finger-iris-palm-voice/}. 
According to \cite{6403227}, a set of templates that cover a large number of variations is needed to obtain a robust FR system. Moreover, the periodic re-enrolment can be perceived as intrusive and boring by users and, regardless, it is not the best solution since we are unable to predict all possible changes in the facial appearance even in the short-term. In addition, we cannot be sure that ``old'' templates may remain unrepresentative in the short-medium time: for example, in case of changes due to scratches, suntan, eyeglass type or beard style \cite{Uludag20041533}.
The requirement to have as many templates as possible is often in contrast with the problem of the storage space. For example, when templates must be conveyed in a smart-card or in a smart-phone, this hypothesis does not hold. An appropriate artificial template from a set of existing ones, called ``supertemplate'' \cite{Ryu2005,Jiang:2002:OFT:605089.605099}, could partially overcome this problem, but computing an effective supertemplate is not trivial, because it should embed both temporary and temporal variations. Another possibility is trying to rely on deep networks-based approaches \cite{8114708}, but even in this case, we must be able to incrementally train the whole net as novel samples and novel users are available. The network itself requires large storage space and often a long and not trivial design step and fine parameters tuning are necessary for large scale facial verification, option which is avoided by adopting the standard template-matching-based approach. It is possible to hypothesise that even using state-of-the-art auto-encoded features, as the FaceNet-based ones \cite{facenet}, the manual intervention of users is still necessary for short, medium and long time period. Moreover, several vulnerabilities have been shown to affect, for example, FaceNet, as recently reported in \cite{8338413}. 
It is long time that self-update systems have been proposed as alternative to the repeated enrolment since from the early publications in 2005 \cite{Balcan2005PersonII}.
A self-update system is able to update/add templates without the need of a supervisor, except for the starting step where it is hypothesised that a set of at least one template is stored in the gallery for each subject \cite{Roli2006}. This first set of templates can be collected in the standard way, by supervised enrolment. During system's operation, the submitted images/templates which are close to the template(s) are stored into a buffer and, usually off-line, added to the subject's gallery \cite{1699908}. This is done by estimating an ``updating'' threshold more stringent than the acceptance threshold in order to avoid or limit the possible introduction of impostors. 

Two limitations basically affect current self-update algorithms: (1) they require the progressive addition of samples into the gallery, thus eventual storage space problems are not taken into account; (2) this incremental insertion often leads to the addition of intra-class variations so large that they may cause the insertion of impostors in the user's gallery, despite the stringent updating threshold.
The second item opens a serious breach from the security point of view, whilst the first one leads to the need of a sort of selection or replacement algorithm \cite{Freni2008,Pagano2015} in case that the memory space is not enough. These reasons explain why, to the best of our knowledge, no self-update algorithms have been implemented and integrated yet in real face verification applications.

Therefore, in this paper, as follow-up of \cite{1247}, we faced with the self-update problem by considering a very small number of templates per client. Besides the advantage of meeting eventual and stringent hardware requirements of mobile devices, the limited number of templates may also reduce the probability of introducing impostors into the users' gallery. Our work starts from the counter-intuitive hypothesis that intra-class variations, although large, make the feature space such that it may be considered as smoothly partitioned according to each user. The samples statistically furthest from the dominating mode are totally overlapped with samples of other subjects. This is supported by some experimental evidence showing that, actually, if initial templates are characterised by the neutral expression at a controlled lighting condition, self-update algorithms tend to attract similar faces from the same users, gradually ``drifting'' to other users when expressions or environmental conditions are too far from the initial ones \cite{Roli2006,RattaniDual}.
On the basis of this hypothesis, we introduce two basic methodologies relying on the classification-selection paradigm basically proposed in \cite{Uludag20041533}.

With regard to our previous work \cite{1247}, we updated and increased the evidence reported in that early publication by: (1) clearly explaining the rationale behind our approach, that was only drafted in that early publication; (2) proposing two more algorithms; (3) performing a large set of experiments which simulate conditions near and far from the system's working hypothesis. This allowed to clarify when the proposed methods can work and when not; (4) comparing our algorithms' performance with that of other existing self-updating approaches \cite{Roli2006,Pagano2015,RattaniDual}, in order to assess their possible advantages and drawbacks with respect to the state of the art. 

The paper is organised as follows. Section 2 makes an overview of the current literature, in order to collocate and motivate our proposal and the prior works, in particular, with which our methods are compared. Section 3 describes our methods in detail. The experimental methodology and experiments are shown in Section 4. Conclusions are drawn in Section 5.

\section{State of the art}
We introduce here a common notation useful to explain the rest of this manuscript. The initial set of templates of $k$ users is named \textit{gallery template} $GT=\lbrace t_1,t_2,...,t_N\rbrace $.\\
$GT$ can also be grouped as $GT=\{ GT^1, ..., GT^k\}$, where $GT^i$ is the gallery template of the user $i$, being $i \in \{1,...,k\}$. Each template of $GT$ is associated with a certain subject by the set of labels $L=\lbrace l_1, ..., l_N \rbrace$, where $l_i \in \{1,...,k\}$.

Let $U=\lbrace U_1, ...,U_n \rbrace$ be the so-called \textit{batches} \cite{Rattani2013AMD}. Each element of $U$ is a set of faces collected during system's operations. It is assumed that the batch $U_i$ was collected before $U_{i+1}$. Thus $U$ is an ordered sequence of samples sets collected at different time. Worth noting, $U_i=\{ u_{i1}, ..., u_{iw}\}$, where $w=|U_i|$ and $u_{ij}$ is the \textit{unlabelled} sample $j$ of $U_i$. We pointed out the term ``unlabelled'' because no gallery's subject is associated to this sample.

As a first step, a self-update algorithm estimates the \textit{updating} threshold $t^*$ from $GT$ and, when the batch $U_i$ is available for a certain claimed user, the distances between each $u_{ij} \in U_i$ and any $t_h \in GT$ are computed. Samples of $U_i$ whose distance is less than $t^*$ are added into the user's gallery and the pseudo-label set $PL_i=\{pl_{i1},...,pl_{iw}\}$ is generated. $pl_{ij} \in \{1, ...,k\}$ is the label of the $GT$'s nearest sample to $u_{ij}$, among the ones whose distance is less than $t^*$. All the $U_i$ samples not satisfying the updating threshold constraint for at least one template of $GT$ are disregarded. The details of this basic approach are reported in Algorithm 1.

The semi-supervised learning theory basically inspired these methods \cite{Roli2008}.
In general, self-update do not only rely on the addition of novel templates but, where present, re-setting the system's parameters, for example, the matcher's ones or the feature extraction algorithm's ones.

The first self-update algorithms for face recognition \cite{Liu20031945,IncrPcaZhao} were based on the application of the Principal Component Analysis (PCA) \cite{eigenFace}. 

In \cite{Roli2006}, the PCA is applied at each batch by including the pseudo-labelled samples for incrementally recomputing the eigenfaces. 
Algorithms as the one above can not distinguish between samples that contain redundant information \cite{Roli2006,Phillips:2005:OFR:1068507.1069015}, thus they exhibited the limitations we mentioned in Section 1.
Filtering redundant information can mitigate the growth of facial galleries. Of course, this implies a definition of what ``redundancy'' is. Among others, the ``context sensitive'' method \cite{Pagano2015} which combines a standard self-update approach with a change detection module, tries to insert a new template only if it is not redundant.
The redundancy definition is based on the detection of changes in illumination conditions and, for this reason, a global luminance quality index ($GLQ$) is calculated for both input data and template.
 
Therefore, this method allows to add templates that present many intra-class variants to update the gallery.

Other self-update biometric methods adopted a two-staged approach. In the first stage, the input samples are pseudo-labelled. The second stage selects the best samples.
Ref. \cite{RattaniDual} proposes a classification-selection technique by defining a risk minimisation technique over the definition of a posteriori probability of inserting a genuine user.
This probability is represented by computation of the adjacency matrix among samples based on the related match-scores. The posterior probability estimation comes from the evaluation of the minimum energy as function of the adjacency matrix.
The second stage allows selecting, among samples pseudo-labelled as genuine users (that is, given $u \in U_i$ $\exists l \in \{1, ..., k\} : u$ can be pseudo-labelled with $l$), those that have more information based on the risk minimisation criteria and Bayesian risk theory. 
\begin{algorithm}[H]
\caption{Traditional self-update}\label{trAlg}
\small
\begin{itemize}
  \item \footnotesize $\text{Let } GT=\lbrace t_1,...,t_N \rbrace \text{ be the initial Gallery template} $\BlankLine
  \item \footnotesize$ \text{Let } L=\lbrace l_1,...,l_N \rbrace \text{ be the initial Gallery labels, where } l_i \in \lbrace user_1,...,user_k\rbrace $\BlankLine
  \item \footnotesize $\text{Let } U=\lbrace U_1,...,U_n \rbrace \text{ be the set of batches of unlabelled samples}$\BlankLine
  \item \footnotesize $\text{where } U_j=\lbrace u_{j1},...,u_{jw} \rbrace \text{ and } w=|U_j|$ \BlankLine
  \item \footnotesize Let $classify(U_j,GT,L)$ be the function that returns the pseudo-labels of the batch $U_j$ for each element\BlankLine
\end{itemize}
\footnotesize Estimate the update threshold $t^*$ using $GT$ and $L$\BlankLine
\For {$j =1$ to $n$}{
\footnotesize $PL_j=classify(U_j,GT,L)$\BlankLine
\footnotesize where $PL_j=\lbrace pl_{j1},...,pl_{jw}\rbrace$ \BlankLine
\For{$e=1$ to $w$}{
 \If{$distance(u_{je},GT)<t^*$}{
\footnotesize $ 	GT_{new}=GT\cup u_{je}$\BlankLine
\footnotesize $ 	L_{new}=L\cup lp_{je}$\BlankLine
 }
}
\footnotesize$GT = GT_{new}$\BlankLine
\footnotesize$L = L_{new}$\BlankLine
\footnotesize Estimate the update threshold $t^*$ using $GT$ and $L$ \BlankLine
}
\end{algorithm}

In summary, it is possible to categorise the biometric self-update systems as follows: traditional self-update approaches, described in Alg. \ref{trAlg}, and classification-selection-based approaches, described in Alg. \ref{ourAlg}.

Unfortunately, even the latter methods cause an increase in the galleries size. As a consequence, the impostors' management can be problematic in the medium-long period. 
Moreover, as stated in Section 1, there is no guarantee that the most effective samples are added and that a supervised template selection done at pre-set time intervals cannot do it better.

Therefore, we propose two approaches aimed to keep constant the number of templates per user without loss of effectiveness. This gallery size limit allows to control the computational complexity of the system and to realistically assess the adaptation ability of the system, with a few, non-redundant, set of templates per user. The pre-set number of templates is referred as $p$ in Alg. \ref{ourAlg}. This algorithm generalises all classification-selection methods; where $p$ is not set \textit{a priori}, no limitations are given in terms of galleries size.\\

\begin{algorithm}[H]
\caption{Classification-selection self-update}\label{ourAlg}
\small
\begin{itemize}
  \item \footnotesize $\text{Let } GT=\lbrace t_1,...,t_N \rbrace \text{ be the initial Gallery template} $\BlankLine
  \item \footnotesize$ \text{Let } L=\lbrace l_1,...,l_N \rbrace \text{ be the initial Gallery labels, where } l_i \in \lbrace user_1,...,user_k\rbrace $\BlankLine
  \item \footnotesize $\text{Let } U=\lbrace U_1,...,U_n \rbrace \text{ be the set of batches of unlabelled samples}$\BlankLine
  \item \footnotesize $\text{where } U_j=\lbrace u_{j1},...,u_{jw} \rbrace \text{ and } w=|U_j|$ \BlankLine
  \item \footnotesize Let $classify(U_j,GT,L)$ be the function that returns the pseudo-labels of the batch $U_j$ for each element\BlankLine
  \item\footnotesize Let $p$ be the maximum number of templates per client\BlankLine
  \item\footnotesize Let $select(GT,L,p)$ be a function that selects $p$ samples per user from $GT$\BlankLine
\end{itemize}

\footnotesize Estimate the update threshold $t^*$ using $GT$\BlankLine
\For {$j =1$ to $n$}{
\footnotesize $PL_j=classify(U_j,GT,L)$\BlankLine
\For{$e=1$ to $w$}{
 \If{$distance(u_{je},GT)<t^*$}{
\footnotesize $ 	GT_{new}=GT\cup u_{je}$\BlankLine
\footnotesize $ 	L_{new}=L\cup lp_{je}$\BlankLine
 }
}
\footnotesize $GT,L=select(GT_{new},L_{new},p)$\BlankLine
\footnotesize Estimate the update threshold $t^*$ using $GT$ \BlankLine
}
\end{algorithm}

\section{Proposed methods}
We summarised that early classification-selection approaches \cite{Pagano2015,RattaniDual} tried to manage the pointed out limitations of self-updating. However, none of these approaches considers that the limited availability of storable samples can also be viewed as an advantage: the smaller the gallery size, the smaller the system's complexity and the probability that an impostor can be added too. This is suggested by the experimental evidence reported in \cite{Roli2006}, where impostors are inserted after some updating iterations or cycles \cite{Rattani2013AMD,RattaniDual}. Therefore, there is room for methods able to exploit this apparent limitation and convert it into an advantage.

What we wrote is against that evidence for which the more the templates, the more the system's accuracy. However, the recent efforts aimed to provide compact and effective facial representations can be of significant help in supporting the above observations. 

First of all, let us hypothesise that the chosen facial features are in a space where it is possible to apply and define a \textit{distance}; for example, the Euclidean distance or the L1 distance.

\begin{figure}[htbp]
\centering%
\includegraphics[scale=0.18]{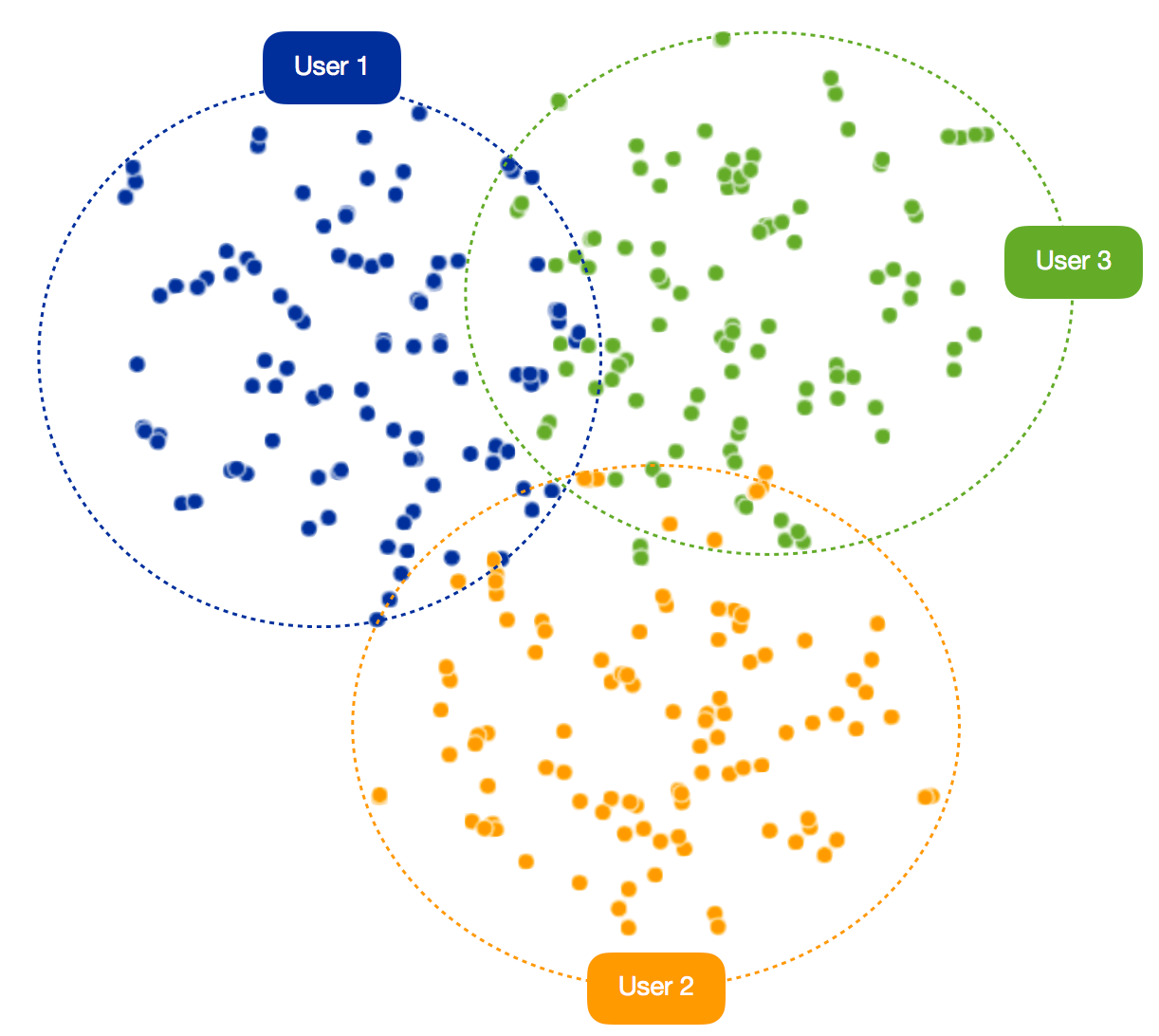}  
\caption{2D graphic representation of the sample distribution on a data set of 3 users under the hypothesis presented. Each cluster is associated with a user. The clusters are partially overlapping.}
\label{fig:overlap}
\end{figure}

Fig. \ref{fig:overlap} is our starting point, where the samples from three users are depicted. A circle drafts the distribution of the features for each of them.  

Ideally, assuming that a face is represented by a feature space $x$ sampled from the appearance of $k$ enrolled subjects:

\begin{equation}
p(x) = \sum_i^k p(x \mid l(x)=i) P(l(x)=i)
\label{eq:clusterFormula0}
\end{equation}
Where $l(x)$ is a labelling function such that $l(x) \in \{1,...,k\}$.

Since it is not possible to be aware of all possible samples' labels \textit{a priori}, we may hypothesise that $p(x)$ depends on an analogous number $k$ of possible clusters of \textit{unlabelled} samples $CL=\{CL_1,...CL_k\}$:

\begin{equation}
p(x)= \sum_i^k p(x \mid x \in CL_i) P(x \in CL_i)
\label{eq:clusterFormula}
\end{equation}

Fig. \ref{fig:overlap} shows what we mean. The depicted circles are the possible projections of $p(x \mid x \in CL_i)$, which are overlapped in some regions. This can be further modelled by the contribution of all known subjects. In other words, each $CL_i$ can be seen as a set of $k$ subsets $CL_{i}=\{ CL_{i1}, …, CL_{ik} \}$, where each sample in $CL_{ij}$ is labelled as the subject $j$, that is, $\forall x \in C_{ij}, l(x)=j$.
Accordingly:
\begin{equation}
p(x \mid x \in CL_i)= \sum_j^k p(x \mid x \in CL_i, x \in CL_{ij}) P(x \in CL_{ij} \mid x \in CL_i)
\label{eq:clusterFormula2}
\end{equation}
Being $CL_{ij} \subseteq CL_i$,  we have $P(x \in CL_{ij} \mid x \in CL_i) = P(x \in CL_{ij})$. If we hypothesise a large majority of samples of the subject $i$ falling in $CL_i$, we have that $P(x \in CL_{ii})$ is much more than any other $P(x \in CL_{ij})$; Eq. \ref{eq:clusterFormula2}  can be rewritten as:
\begin{equation}
p(x \mid x \in CL_i) \approx p(x \mid x \in CL_{ii}) 
\label{eq:clusterFormula3}
\end{equation}
Each cluster concurring to the generation of $p(x)$ is dominated by the mode $p(x \mid x \in CL_{ii})$.

On the basis of this modelling, the individual contribution of user $i$ to the whole feature space is given by $\cup_j CL_{ji}$, which corresponds to:

\begin{equation}
p(x \mid l(x)=i)= \sum_j^k p(x \mid x \in CL_{ji}) P(x \in CL_{ji}) P(x \in CL_j)
\label{eq:clusterFormula4}
\end{equation}

Therefore, templates should be selected from $\cup_j CL_{ji}$. In this paper, we introduced two ways of detecting such templates: in the first one, we hypothesise that $p(x \mid x \in CL_i)$ is further characterised by one centroid called $c_i$ which approximates the centroid of $CL_{ii}$ according to Eq. \ref{eq:clusterFormula3}; in the second one, we use the patterns located in $\cup_j CL_{ji}$, that is, over the region characterised by the probability in Eq. \ref{eq:clusterFormula4}. In both cases, each partition $\cup_j CL_{ji}$ is estimated by the pseudo-labelling step used in standard self-update. 
The core of our methodology consists in the template selection phase, indicated generically in the Alg.\ref{ourAlg} with the $select$ function. We propose two different criteria in Sections \ref{sec:clustering}-\ref{sec:editing}. The different behaviour of the self-update system that comes out from each criterion is hypothesised and verified by experiments in Section 4.

 \begin{figure}[htbp]
 \centering%
 \subfigure[]{
 \includegraphics[width=0.32\textheight]{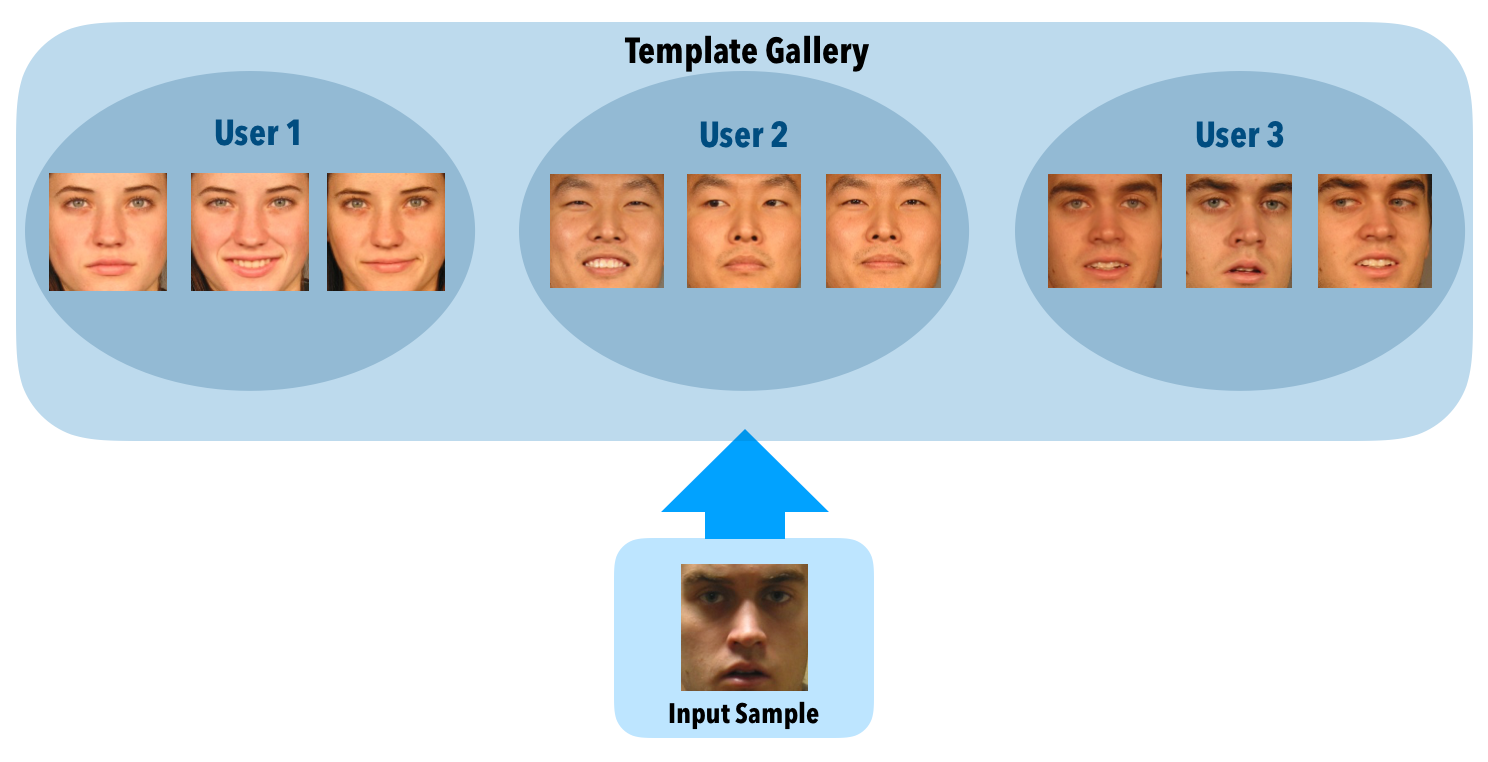} 
  \label{fig:exampleupdateInput}
 }
 \subfigure[]{
 \includegraphics[scale=0.23]{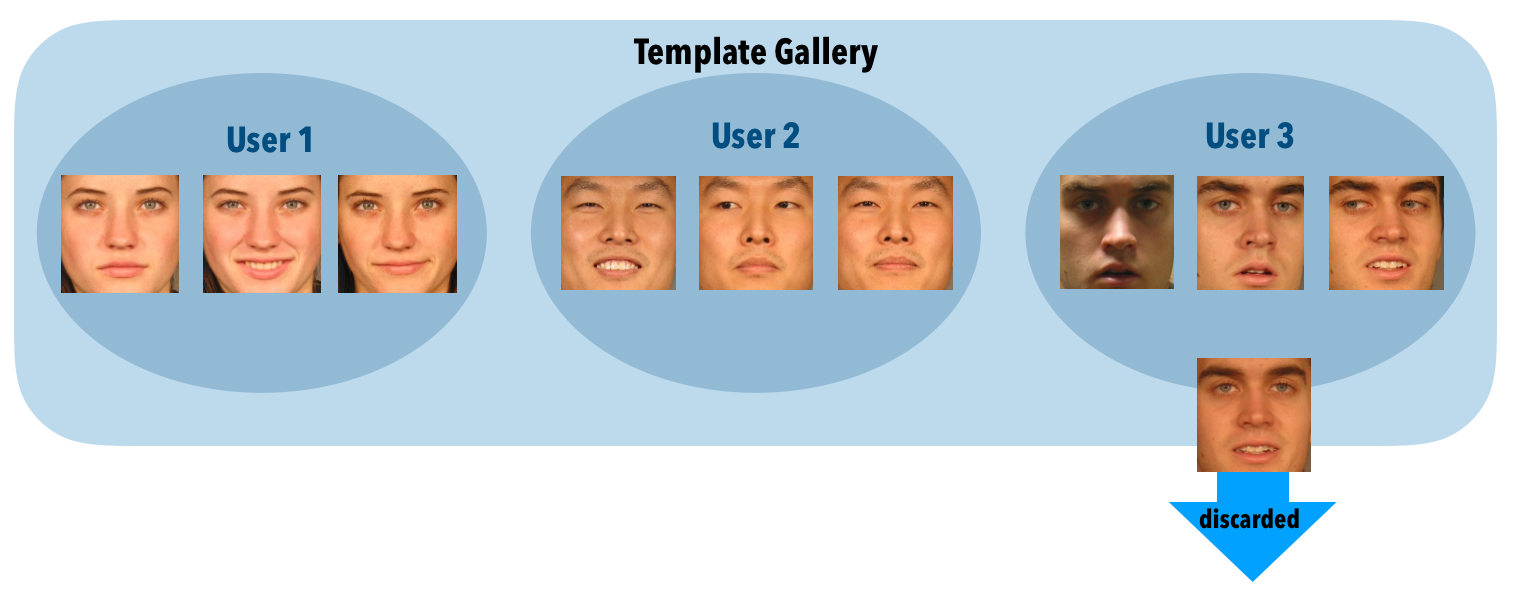} 
  \label{fig:exampleupdateDiscard}
 }\vspace{-5pt}

 \caption{Example of  gallery update on a system with three templates per user limit.}
 \label{fig:exampleupdate}
 \end{figure}

\subsection{Clustering-based classification-selection}
\label{sec:clustering}

Let us suppose that the cluster $CL_{i}$ is characterised by a centroid $c_i$. For example, if the generating function of facial samples is Gaussian for each subject, the centroid corresponds to $p(c_i \mid c_i  \in  CL_i)>p(x \mid x \in CL_i)$, $\forall x \neq c_i$. By moving away from $c_i$, the distribution gradient is gradually negative; thus, the points near to $c_i$ are also geometrically close to each other, as well as their probability of occurrence is high.
According to the above modelling, the most probable occurrences of possible templates of users $i$ are around $c_i$, since they are representative of $CL_{ii}$, that is, the main mode of $CL_i$. The intra-class variations far from this point are less frequent and, anyway, potentially overlapped by samples of other clusters.

This is suggested by state-of-the-art results where it is evident that first pseudo-labelled samples added to a certain subject's gallery are geometrically and statistically quite close to his starting templates (\textit{e.g.} \cite{Roli2006}); then, they gradually tend to go far away from that templates, by describing less frequent intra-class variations: for example, from small lighting variations to relevant facial expressions variations. Consequently, they can be misclassified with similar variations of other subjects. This behaviour is clearly reported in \cite{Roli2006}.

Worth mentioning, Eq. \ref{eq:clusterFormula} still admits a (high) degree of overlapping among facial samples, as acknowledged by the research community. On the other hand, Eq. \ref{eq:clusterFormula} hypothesises the existence of specific templates characterising the dominating mode. Therefore, the updating step must look for those samples instead of trying capturing the largest intra-class variations which may lead to including impostors in the client's gallery. If this holds for long-term variations too, the user's gallery should change gradually because its samples follow the centroid variations, which are the most significant ones.

Following the observations above, the first method is based on the detection of the ``centroid'' of each user's distribution as expected from the appearance of $p(x|x\in CL_i)$ in Eq. \ref{eq:clusterFormula}. 
According to Alg. \ref{trAlg}, each $u\in U_i$ is classified by using the updating threshold $t^*$, thus obtaining the pseudo-labels $PL_i$. As a matter of fact, this corresponds to the estimation of each $\cup_j C_{ji}$, that is, $GT_{new}^i$ in Alg. \ref{kmeansAlg}. Next, the algorithm selects the $p$ samples closest to the centroid, as described in Alg.\ref{kmeansAlg}.
The hypothesis of Gaussian generating function allows to adopt the well-known K-Means algorithm. This searches for natural clusters at the feature level. The cluster ensemble approaches have been successfully applied to different research areas \cite{BOONGOEN20181,6762944} and are appropriate to address the problem of template selection.

\begin{figure}[htbp]
\centering%
\includegraphics[scale=0.18]{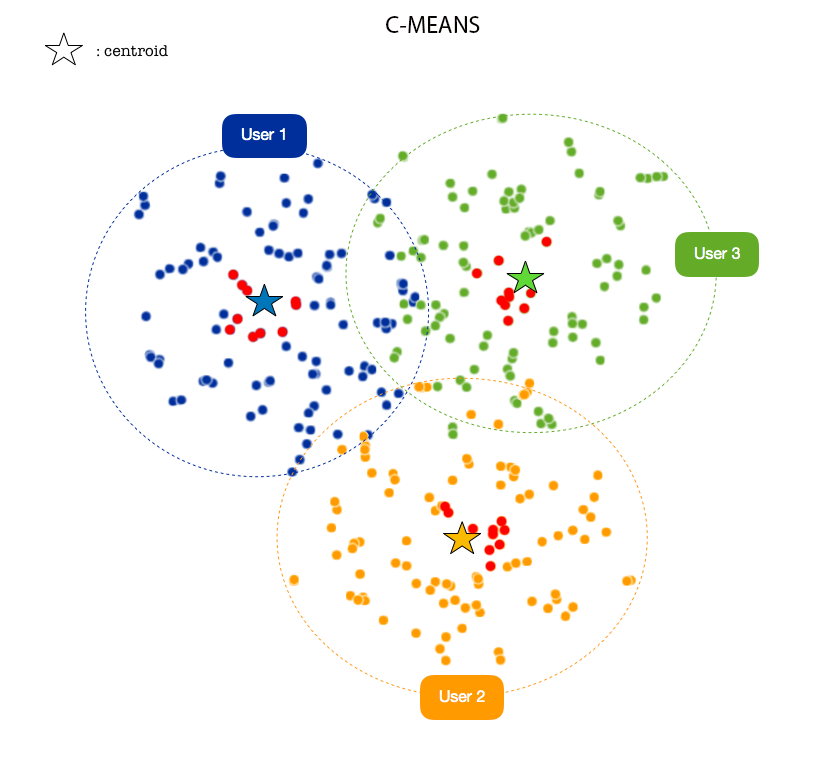}  
\caption{2D graphical representation of the application of the K-Means algorithm on a data set of 3 users. The stars symbolise cluster centroids. The samples marked in red are those selected by the algorithm.}
\label{fig:kmeansdistr}
\end{figure}

We applied this algorithm in order to estimate each $CL_i$ of Eq. \ref{eq:clusterFormula}. Then, $CL_{ii}$ is the subset of $CL_i$ with the largest number of samples labelled and pseudo-labelled with $i$ (Fig. \ref{fig:kmeansdistr}). Finally, $c_i$ is computed. This agrees with the meaning of Eqs. \ref{eq:clusterFormula2}-\ref{eq:clusterFormula3}.
The most ``representative'' samples per user with (pseudo-)label $i$ can be then selected around $c_i$. 

If the modelling of Eq. \ref{eq:clusterFormula2} is not satisfied, two problems may arise:
\begin{enumerate}
\item A user may be associated with more than one cluster: in this case, a selection criterion for the most representative ``mode'' should be selected, or kept both. This also means that some users could not have ``their'' reference mode, and this leads to the impossibility of updating their galleries.
\item A cluster may be associated with more than one user: in this case, a selection criterion may be decided for the users to whom the cluster must refer.
\end{enumerate}

In Section \ref{sec:experiments}, we forced the proposed algorithms to work under different environmental conditions, represented by three different data sets, in order to verify if such cases might occur and why.\\
\begin{algorithm}[H]
\caption{$select(GT_{new},L_{new},p)$ by K-Means}\label{kmeansAlg}
\small
\begin{itemize}
\item \textbf{Input:} $GT_{new}=\lbrace GT_{new}^1,..,GT_{new}^k\rbrace, L_{new}=\lbrace L_{new}^1,..,L_{new}^k\rbrace, p $ \text{ where } $GT_{new}^i$ is the template gallery of user $i$ with label $L_{new}^i$ 
\item \textbf{Output:} $GT=\lbrace GT^1,..,GT^k\rbrace$,$L=\lbrace L^1,..,L^k \rbrace$ \text{ where } $GT^i$ is the selected template gallery of user $i$ with label $L^i$
\item Let $kmeans(T,L,k)$ be a function that partitions $N$ observations in $T$ into $k$ clusters  $CL=\lbrace CL_1,..CL_k \rbrace$  \text{ with labels } $LCL=\lbrace LCL_1,..LCL_k \rbrace$ from $L$ \text{ and centroids }  $C=\lbrace c_1,..c_k\rbrace$, \text{ where } $N=|T|=|L|$
\end{itemize}
\footnotesize $[CL,LCL,C]=kmeans(GT_{new},L_{new},k)$\BlankLine
\For{ $i=1$ to $k$}{
\For{ $j=1$ to $k$}{
\footnotesize $S_j=\{l \in LCL_j | l=i$\}
}
\footnotesize $u=\mbox{arg}\max_{j}|S_j|$ \text{ where} $|...|$ \text{denotes the set cardinality}\BlankLine
\footnotesize $GT^i=\varnothing$\\
\For{ $h=1$ \text{to} $p$}{
\footnotesize$ x=\mbox{arg}\min_{y \in GT_{new}^u}(\|y-c_i\|^2 )$\BlankLine
\footnotesize$GT^i=Gt^i\cup \{x\}$\BlankLine
\footnotesize$GT_{new}^u=GT_{new}^u \setminus \{x\}$
}
}
\end{algorithm}
\subsection{Classification-selection by editing algorithms}
\label{sec:editing}

The second approach is based on relaxing the hypothesis of Gaussian generating function. This leads to the problem of how detecting the centroid. 
A possible alternative criterion is to find $p$ samples of $\cup_j CL_{ji}$ that are, averagely, the closest each other for the user $i$. This could lead to avoid the case of bad estimation of the centroid due to lack of data, since we search over a space larger than that of $CL_{ii}$. Worth noting, $\cup_j CL_{ji} \equiv GT_{new}^i$ in  Alg. \ref{replAlg}. Therefore, we adopted the editing algorithm for reducing the training set size in K-NN-based classification tasks \cite{LUMINI2006495} and also used for supervised template updating \cite{Freni2008}, named MDIST (Alg. \ref{replAlg}).

\begin{figure}[htbp]
\centering%
\subfigure{
\includegraphics[scale=0.19]{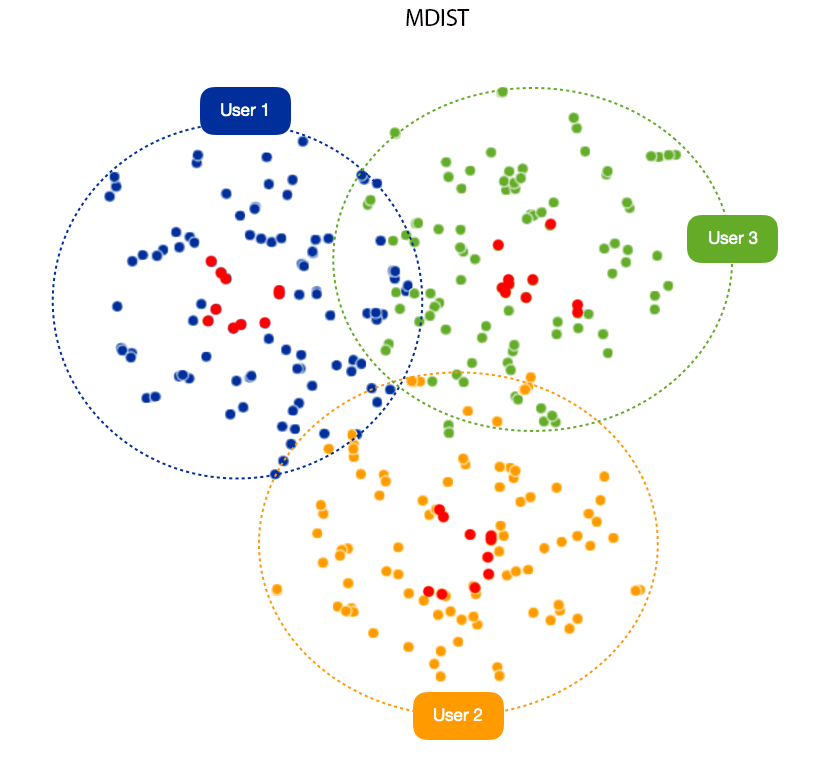}  
\label{sub:mdist}
}
\subfigure{
\includegraphics[scale=0.19]{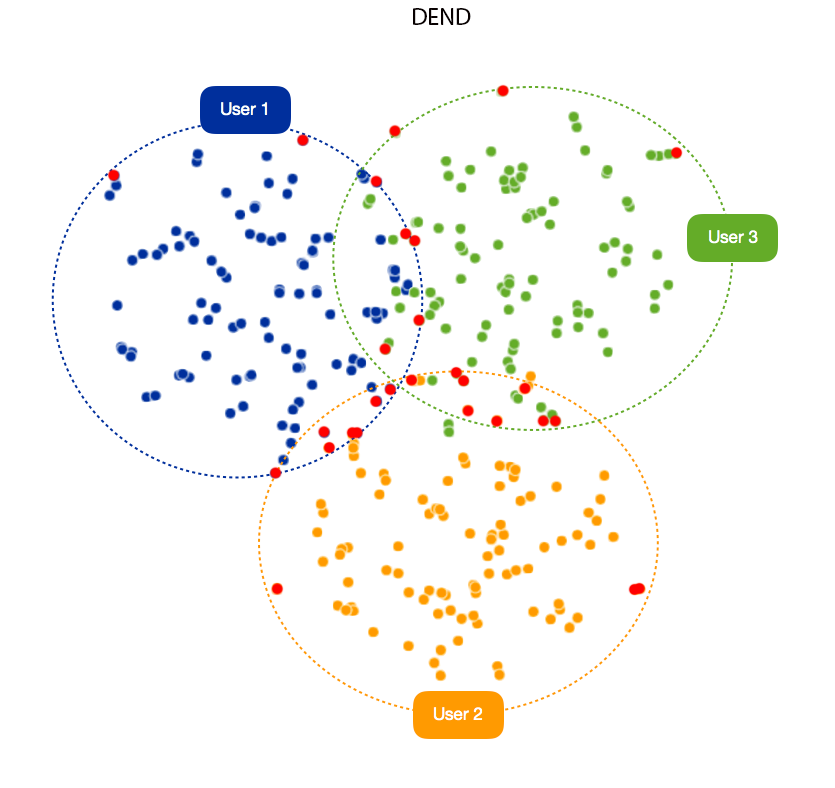}   
\label{sub:dend}
}
\caption{2D graphical representation of the application of the MDIST and DEND algorithms on a data set of 3 users. The samples marked in red are those selected by the algorithm.}
\end{figure}

In order to have a counterproof of our hypothesis's likelihood, depicted by Eq. \ref{eq:clusterFormula3}, we also implemented the DEND algorithm, which is complementary to MDIST \cite{Freni2008}. In this case, the pseudo-labelled and/or labelled samples which are averagely the furthest each other are selected, because they are potentially clusters $CL_{ji}$, thus far from the dominating mode.

Again, the system is two-staged: (1) the input samples are pseudo-labelled; (2) the Euclidean distances of the feature vectors of all samples are calculated and $p$ templates are selected. The MDIST algorithm selects the $p$ templates with the smallest average distance (Fig. \ref{sub:mdist}), that is, templates averagely close in the features space, therefore, with small but relevant intra-class variations; the DEND algorithm selects  the $p$ templates with the most significant average distance (Fig. \ref{sub:dend}), that is, templates with intra-class variations such that they are, according to our hypothesis, statistically and geometrically overlapped to those of other users.

This behaviour is exemplified in Figs. \ref{sub:mdist}-\ref{sub:dend}. Therefore, we expect that MDIST has a similar performance of K-Means, whilst DEND should confirm that looking for large intra-class variations means to move away from the main mode.

An important future evaluation could concern the integration of the K-Means and MDIST methods. In fact, it is possible to consider more stringent rules for the template selection: among the closest samples to each centroid, we could also apply the constraint that their related distances are minimised.

It is important to note that the choice of the threshold $t^*$ influences the proposed methods.
A selective threshold in the classification step, as the so-called zeroFAR operational point (FAR=0\%), assures that no impostors are considered as genuine samples, drastically reducing the possibility for an impostor to pass the classification phase. In this case, however, genuine faces with high intra-class variations are rejected in the first phase.
A less stringent threshold, on the other hand, increases the possibility that an impostor passes the classification phase but allows the presence of a greater number of genuine samples with many intra-class variations. This also agrees with our working hypothesis.
For this reason, the threshold must be set according to the type of selection that follows the classification. The MDIST method (as well as K-means method) selects the samples closest each other; the DEND, which selects the samples on the edge of the clusters, it is better to have a more accurate classification at the expense of the variability of the biometric trait. In our experiment, we set a relatively stringent updating threshold, thus meeting the characteristics of both algorithms.

\begin{algorithm}[H]
\caption{$select(GT_{new},L_{new},p)$ by MDIST}\label{replAlg}
\small
\begin{itemize}
    \item \textbf{Input:} $GT_{new}=\lbrace GT_{new}^1,..,GT_{new}^k\rbrace, L_{new}=\lbrace L_{new}^1,..,L_{new}^k\rbrace, p $ \text{ where } $GT_{new}^i$ is the template gallery of user i with label $L_{new}^i$ 
    \item \textbf{Output:} $GT=\lbrace GT^1,..,GT^k\rbrace, L=\lbrace L^1,..,L^k \rbrace$ \text{ where } $GT^i$ is the selected template gallery of user i with label $L^i$
\end{itemize}

\For {$i =1$ to $k$}{

\footnotesize $GT^i=\{\textbf{x} \subset GT_{new}^i,|\textbf{x}|=p:\mbox{arg}\min_{\textbf{y} \subset GT_{new}^i,|\textbf{y}|=p} (\sum_{h=1}^{p}\sum_{b=1}^{p}(y_h-y_b)^2)\}$\BlankLine

}

\end{algorithm}

\section{Experimental results}
\label{sec:experiments}

\subsection{Data set}
\label{sec:data set}
Three data sets are used for our experiments: Multimodal-DIEE, FRGC and a subset of the set of data LFW.

\begin{figure}[htbp]
\centering%
\subfigure{
\includegraphics[scale=0.185]{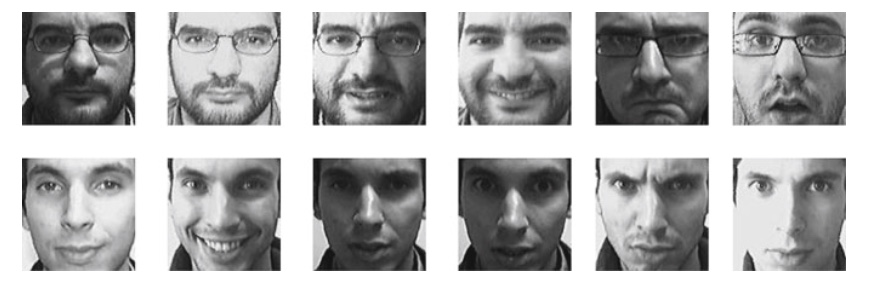}  
\label{sub:diee}
}
\subfigure{
\includegraphics[scale=0.185]{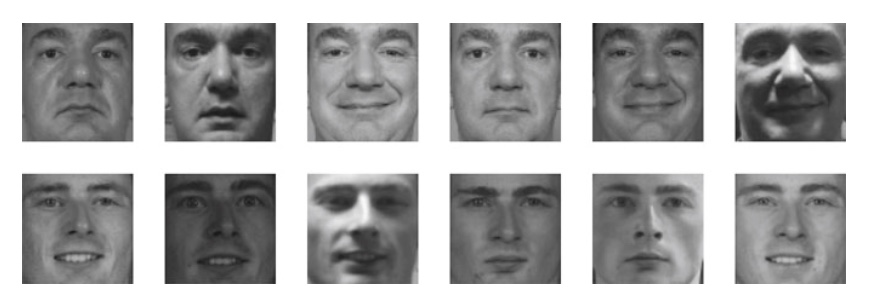}   
\label{sub:frgc}
}
\subfigure{
\includegraphics[scale=0.26]{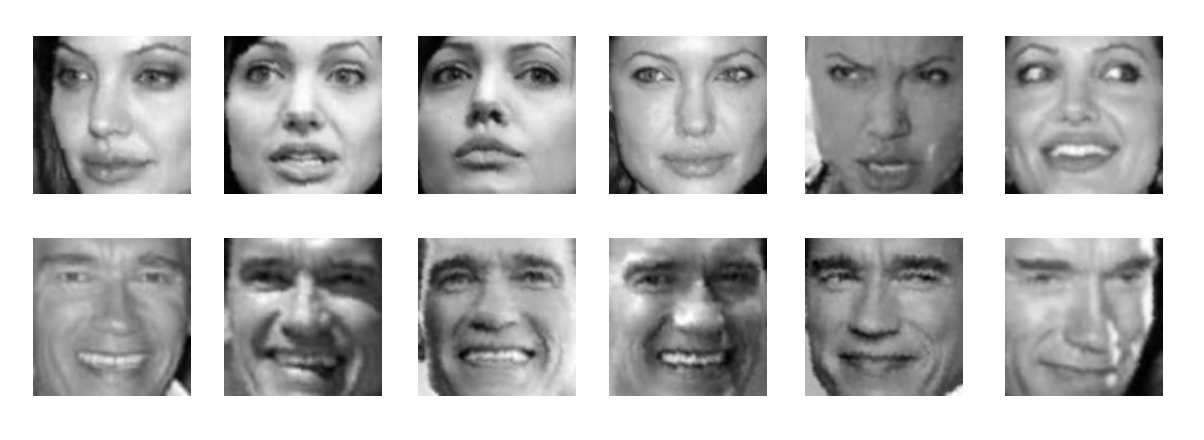} 
\label{sub:lfw}
}
\caption{Example of variation of faces in Multimodal-DIEE (first set), FRGC (second set) and LFW data set (third set).}
\end{figure}

The Multimodal-DIEE data set \cite{Rattani2013AMD} (Fig. \ref{sub:diee}) was acquired in our laboratory. The data set is composed of 59 users, 60 faces per user. During this period, about 1.5 years, 6 acquisition sessions were done. Faces were captured in the frontal pose, at the same distance from the camera. Some variations in lighting conditions are present. 

The FRGC \cite{Phillips:2005:OFR:1068507.1069015} was acquired by the University of Notre Dame. It includes the faces of 222 users acquired on 16 sessions (Fig. \ref{sub:frgc}). Some of these sessions contain uncontrolled captures. The faces have variations of expressions and lighting. For the experiments, 187 users have been selected and about 100/200 faces per user. 

The LFW data set \cite{LFWTech} consists of more than 13,000 web images collected for a total of 5749 users.
For 4069 people there is only a single image in the database. For many other users, the images are not enough to be used in self-update experiments (several batches need to be able to evaluate the evolution of the data set over time). For this reason, a subset of 16 people with about 15/30 faces per user for a total of 390 faces was selected from the data set. Due to the nature of the data set, images are acquired in a completely uncontrolled manner.
The faces in the data set have variations of pose, lighting, age and expression (see Fig. \ref{sub:lfw}). 

As usual, the pre-processing step precedes the feature extraction process by using the BSIF algorithm or FaceNet, ResNet50 and SeNet50 auto-encoding networks \cite{BSIF,facenet,He2015DeepRL}. Pre-processing is based on the following stpdf: faces are rotated in order to align eyes, guaranteeing a pre-set interocular distance, scaled, cropped and saved in grayscale. The images are scaled to a 100x100 size for the Multimodal-DIEE and LFW data sets and 150x150 for the FRGC data set, for the BSIF application and to 160x160 size for the FaceNet, ResNet50 and SeNet50 applications. For the BSIF extraction we normalized images using DoG (Difference of Gaussian)\cite{dog}. This algorithm is a band-pass filter, that highlights the facial features in order to mitigate the illumination problems. After the pre-processing phase: 
\begin{itemize}
    \item the BSIF algorithm with 10 filters of size 9x9 pels is applied. The BSIF algorithm \cite{BSIF} calculates a binary code string for the pixels of a given image. The image resulting from the BSIF application is then subdivided into 6x6 non-overlapping regions. For each region, the histogram is calculated. Finally, the histograms are concatenated in a feature vector for a comprehensive description of the face. BSIFs are the state-of-the-art on facial representations based on handcrafted textural features;
    \item the FaceNet auto-encoding network \cite{facenet} is used to derive a 128B feature vector. FaceNet-based feature vectors are the state-of-the-art on facial representations by auto-encoding methods based on deep learning approaches. We used an open-source implementation based on TensorFlow \footnote{https://github.com/davidsandberg/facenet} trained on the model 20170512-110547. This model has been trained on the MS-Celeb-1M dataset \cite{guo2016msceleb};
    \item the ResNet50 auto-encoding network \cite{He2015DeepRL} is used to derive a 2048B feature vector. The pre-trained model is trained on MS-Celeb-1M dataset and then fine-tuned on VGGFace2 dataset \cite{Cao2017VGGFace2AD}.
        \item the SeNet50 auto-encoding network \cite{Hu2017SqueezeandExcitationN} is used to derive a 2048B feature vector. The pre-trained model is trained on MS-Celeb-1M dataset and then fine-tuned on VGGFace2 dataset \cite{Cao2017VGGFace2AD}.
\end{itemize}

\begin{figure}
\centering
\includegraphics[width=\textwidth]{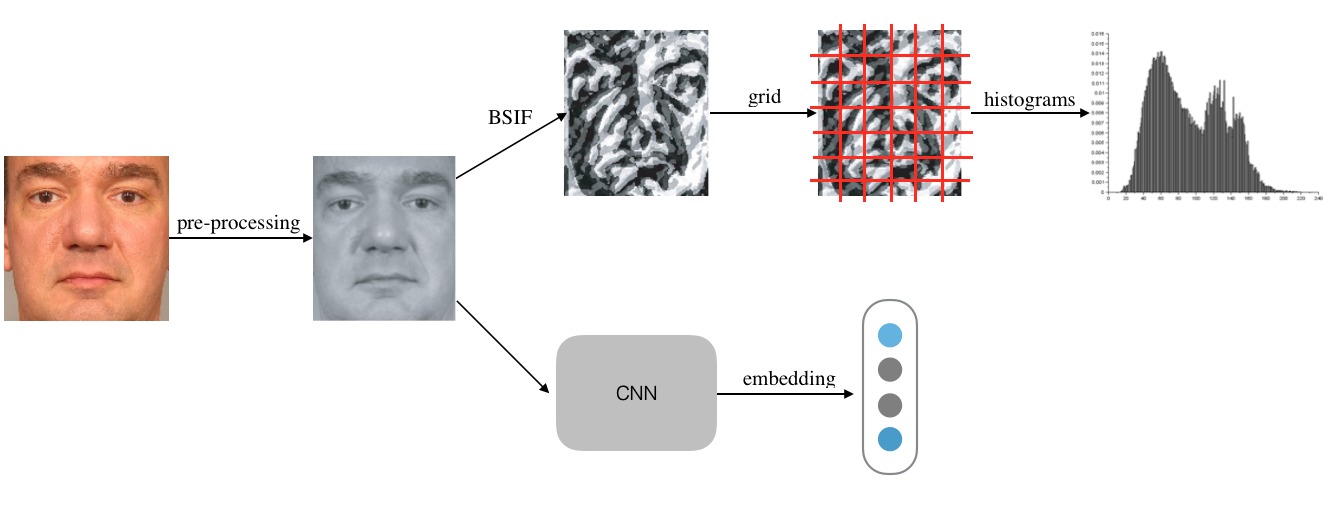}
\caption{Block system of features extraction.}
  \label{fig:featuresExtraction}
\end{figure}

\subsection{A preliminary view on the feature space efficacy}

This Section is aimed to point out how the handcrafted and auto-encoded features spread the genuine users' and the impostors' matching scores over the three data sets. Such analysis can give us some insights on the feature space efficacy and the degree of adherence of our hypothesis to real data.

\begin{figure}
  \centering
  \includegraphics[width=.495\textwidth]{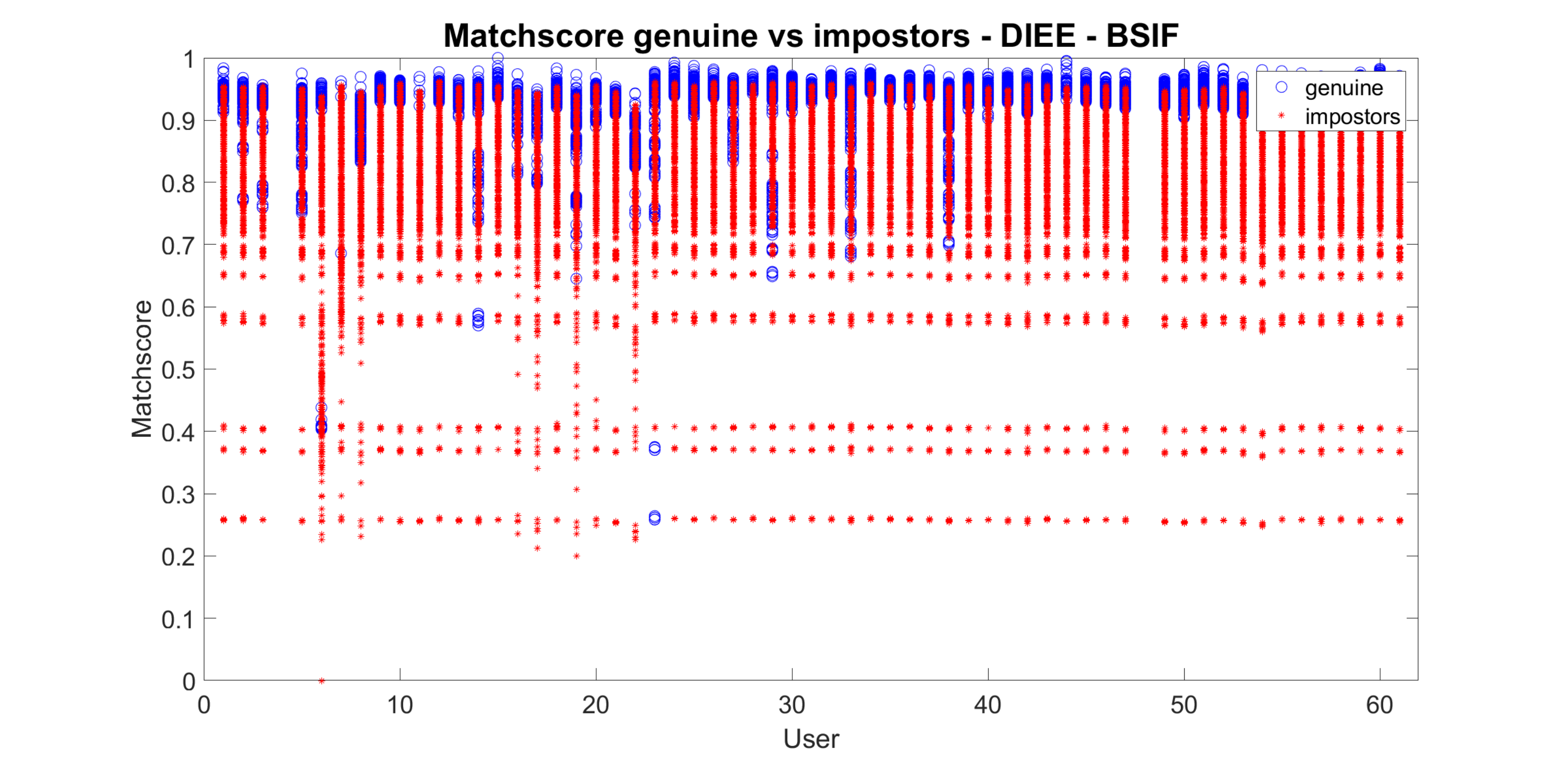}
  \includegraphics[width=.495\textwidth]{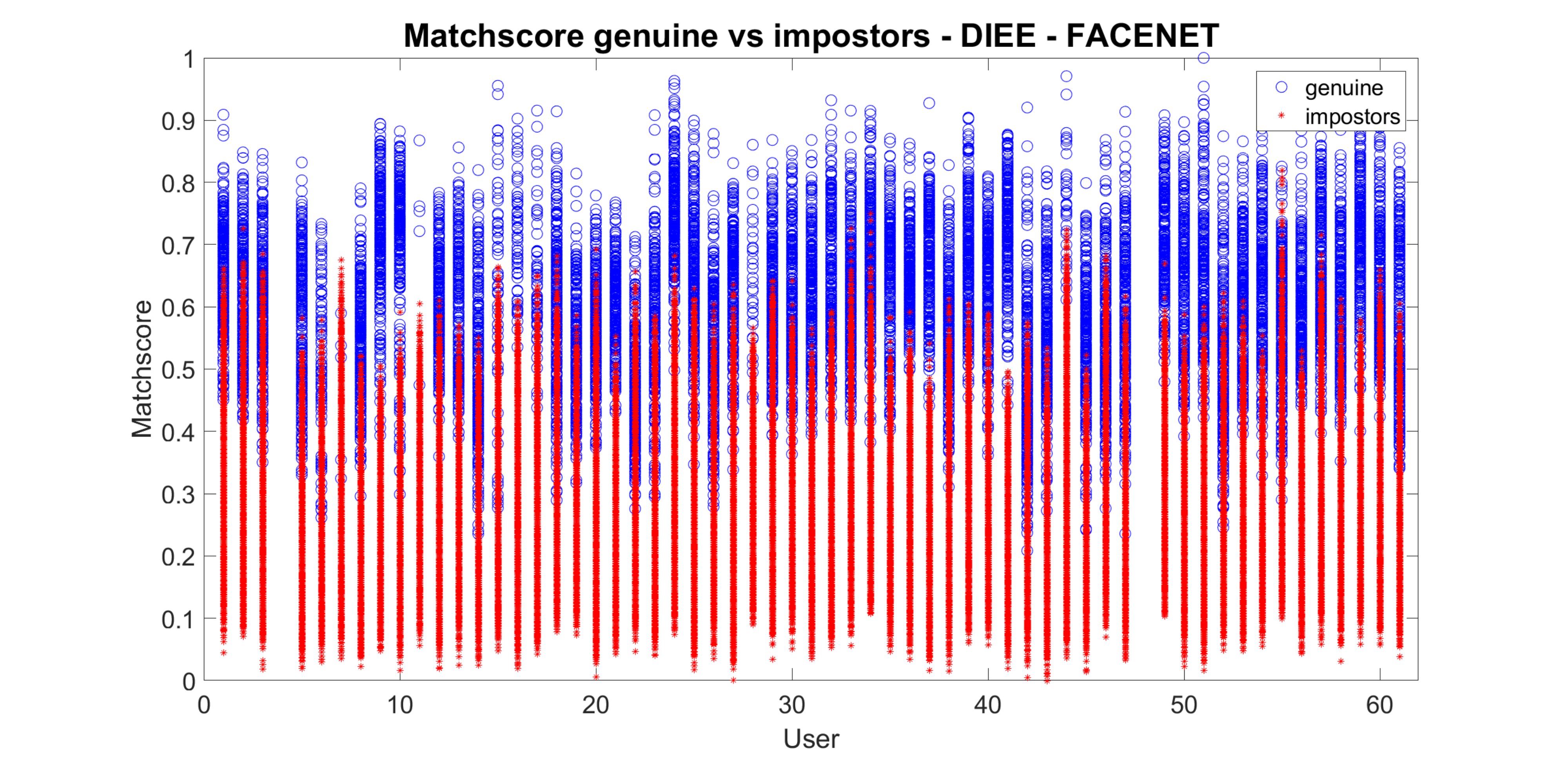}
  \includegraphics[width=.495\textwidth]{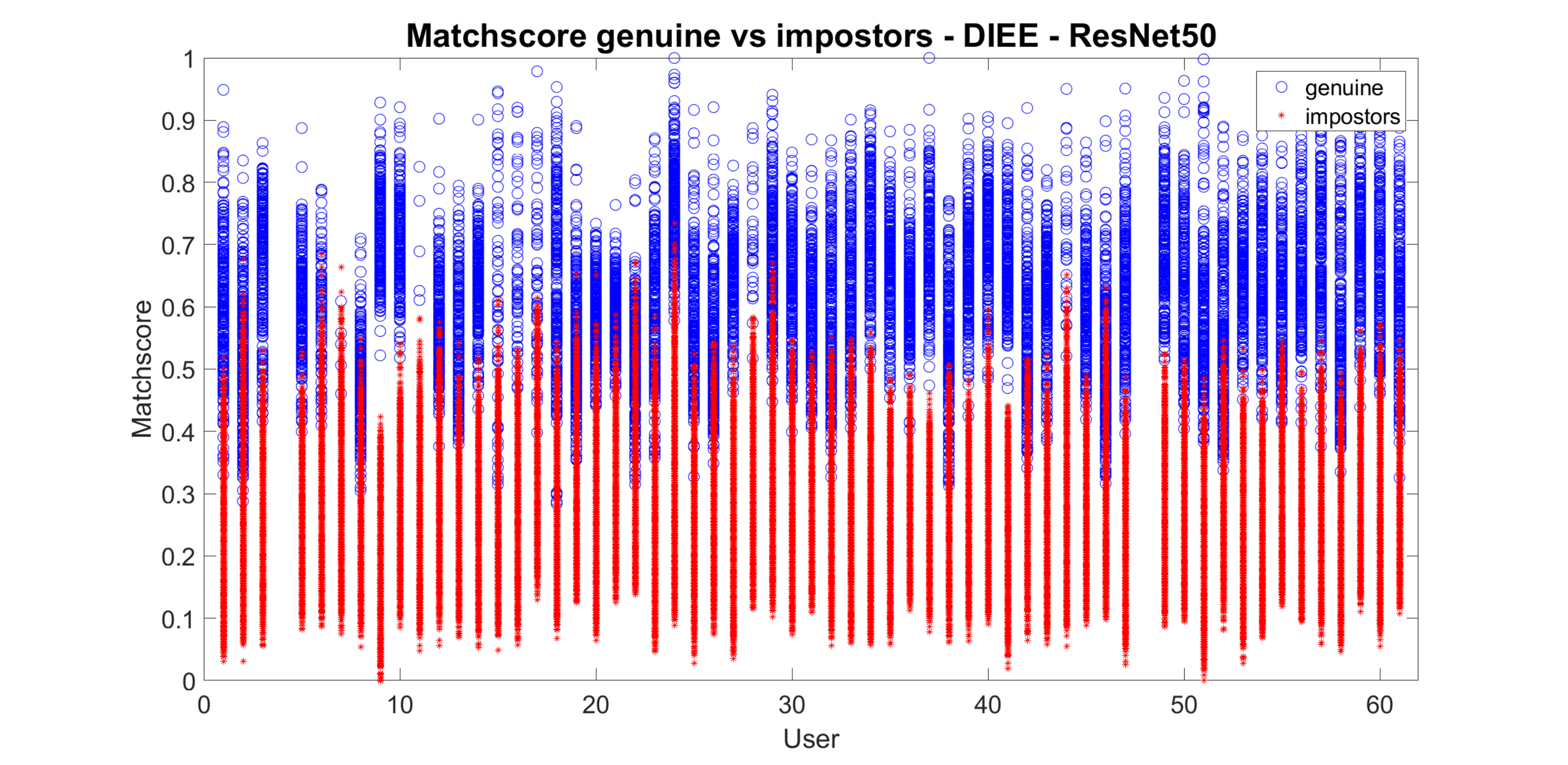}
  \includegraphics[width=.495\textwidth]{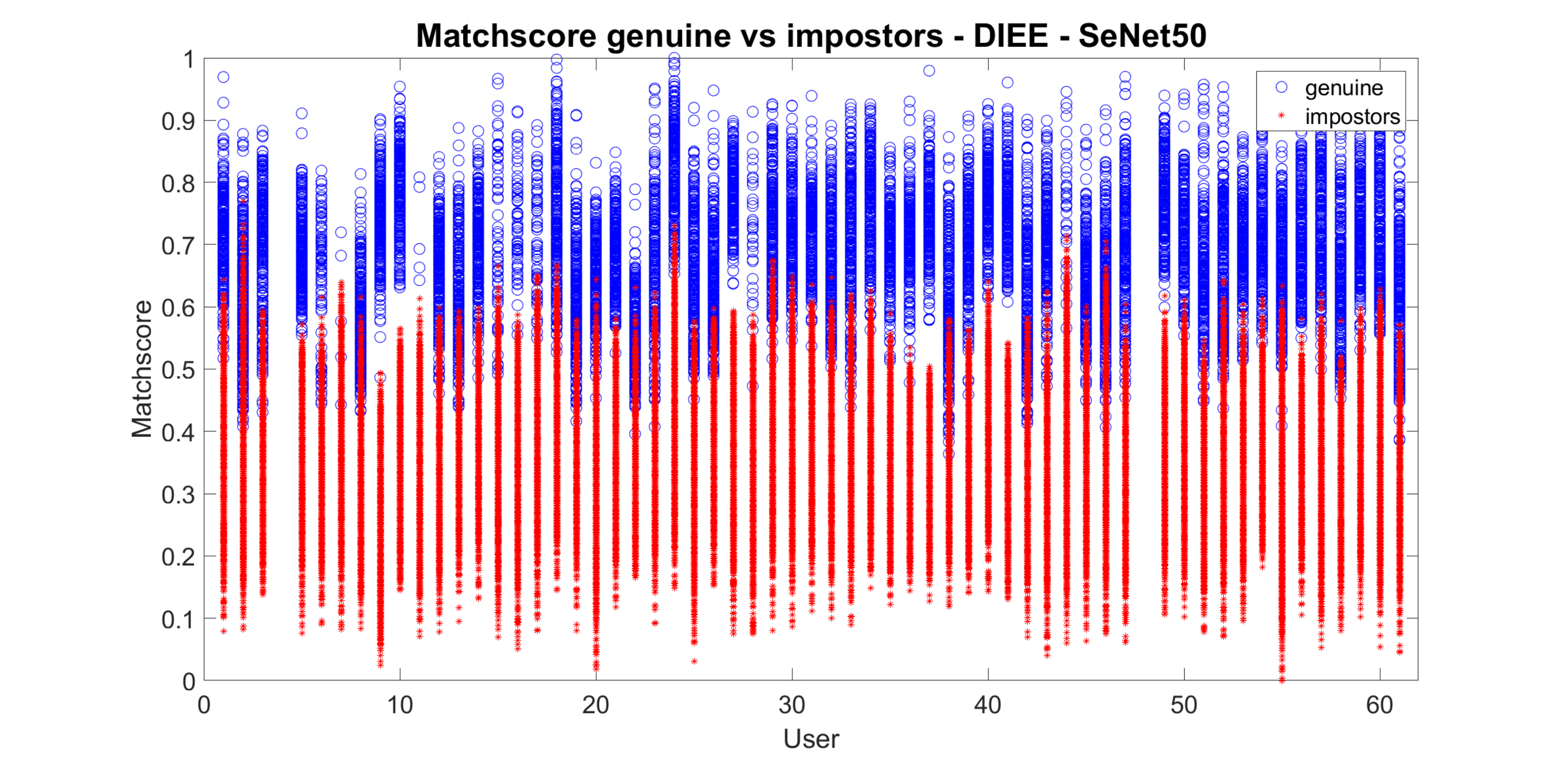}
  \caption{BSIF, FaceNet, ResNet50 and SeNet50 matching scores for the DIEE data set. The plots report in the x axis the subject identifier, whilst in the y axis the matching scores among the first five templates and the other samples are reported. Genuine users' scores are depicted in blue, the impostors' in red. Goal of these plots is to show the degree of adherence of the data with respect to the basic working hypothesis described in Section \ref{sec:clustering}.}
    \label{fig:dieebsif}
\end{figure}

\begin{figure}
  \centering
  \includegraphics[width=.495\textwidth]{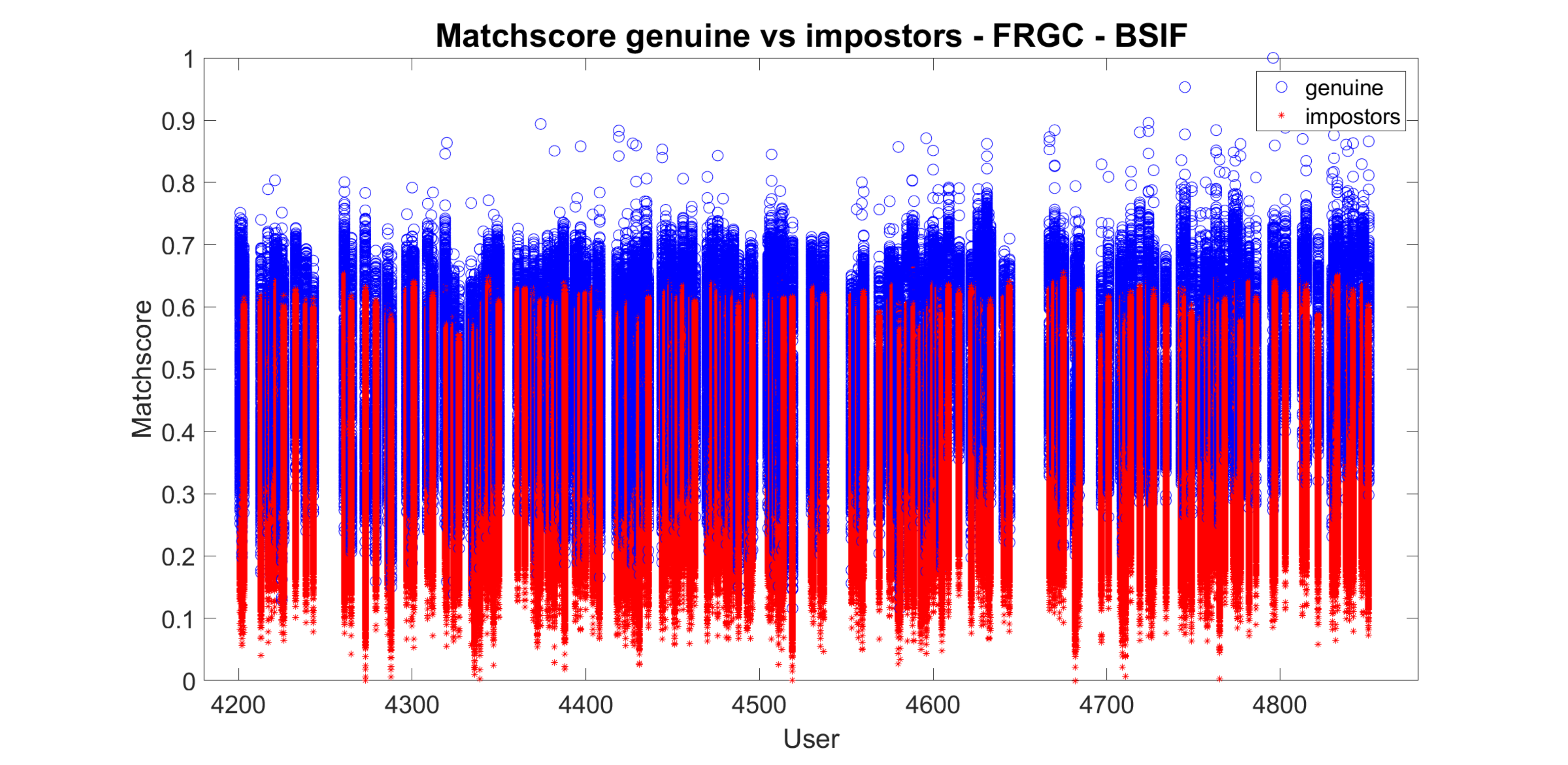}
  \includegraphics[width=.495\textwidth]{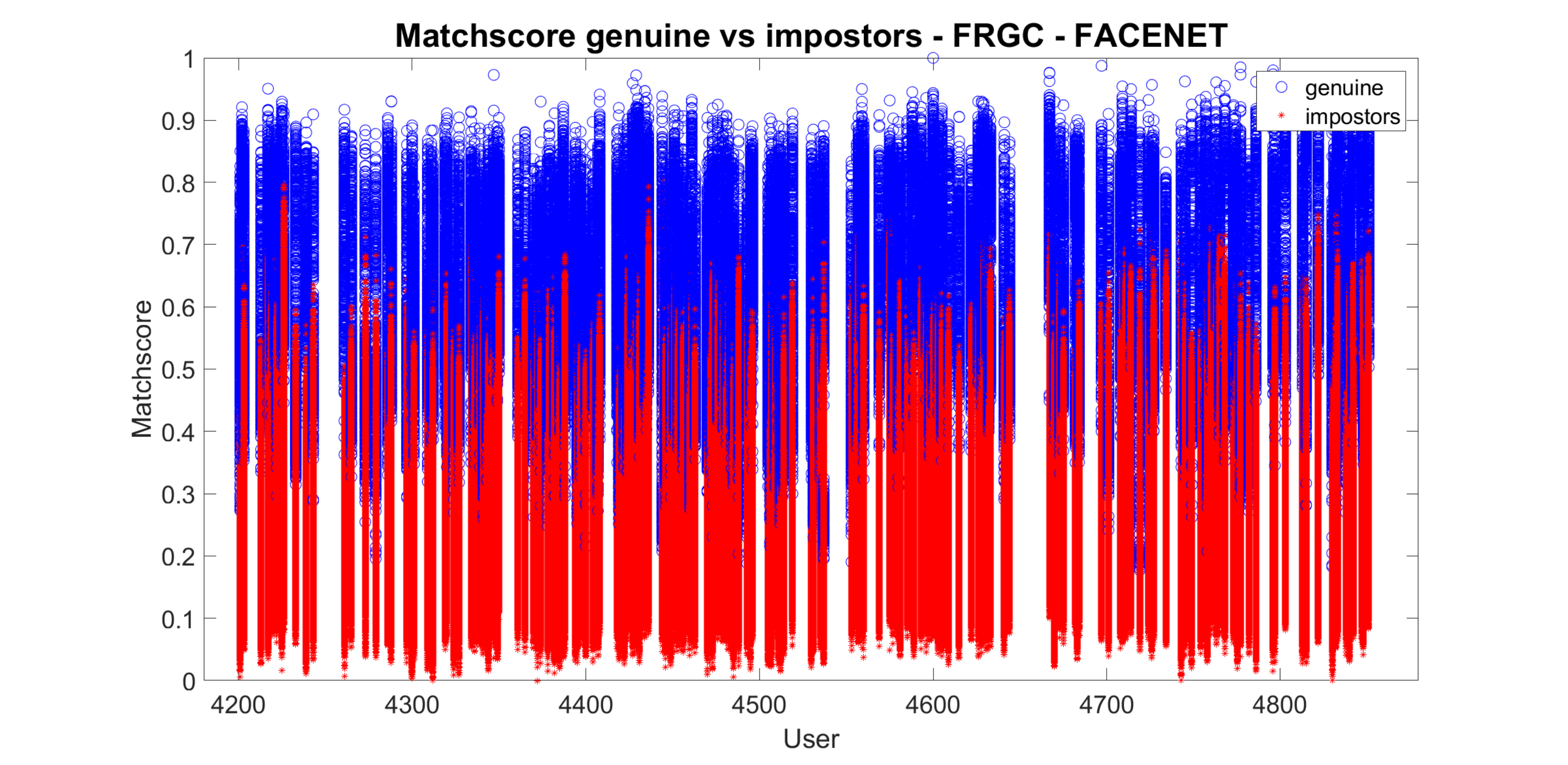}
    \includegraphics[width=.495\textwidth]{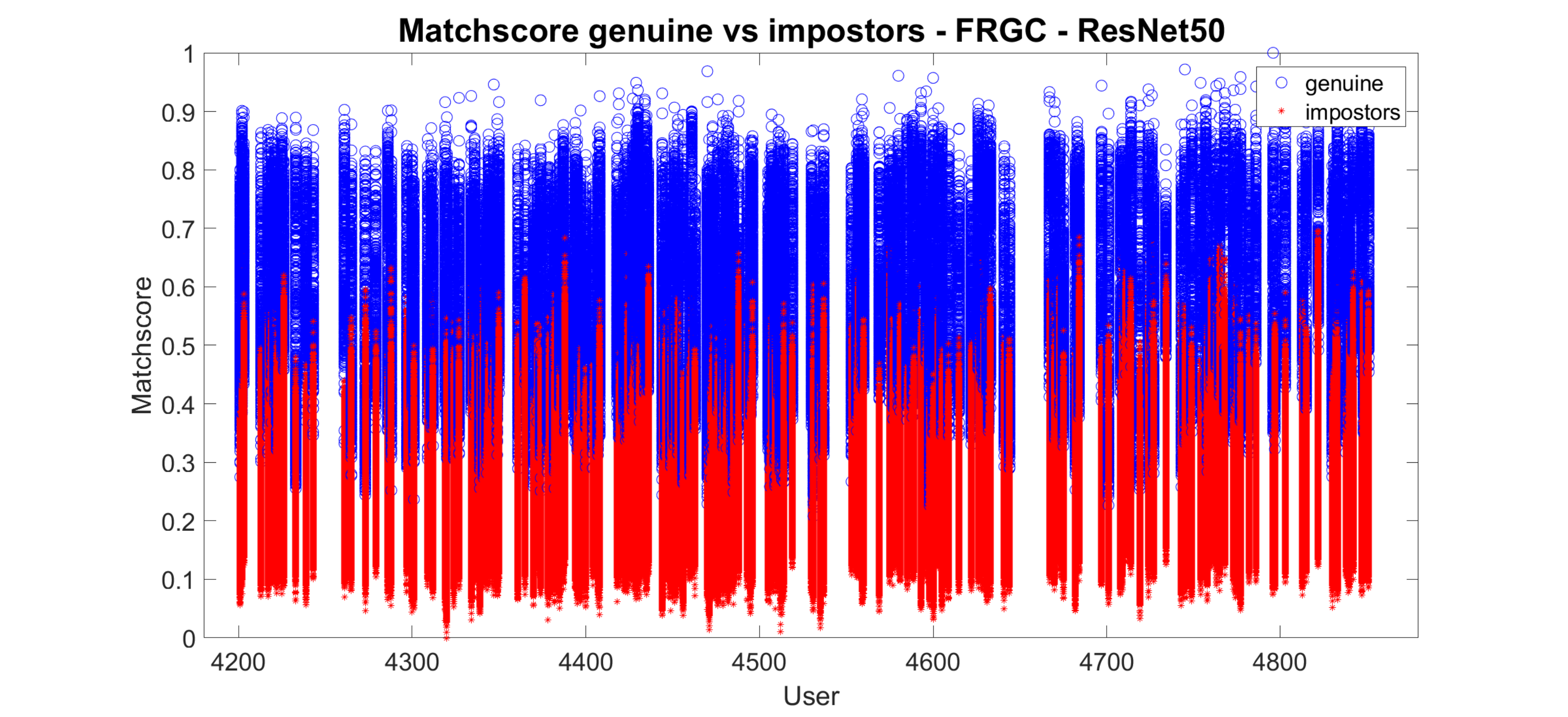}
        \includegraphics[width=.495\textwidth]{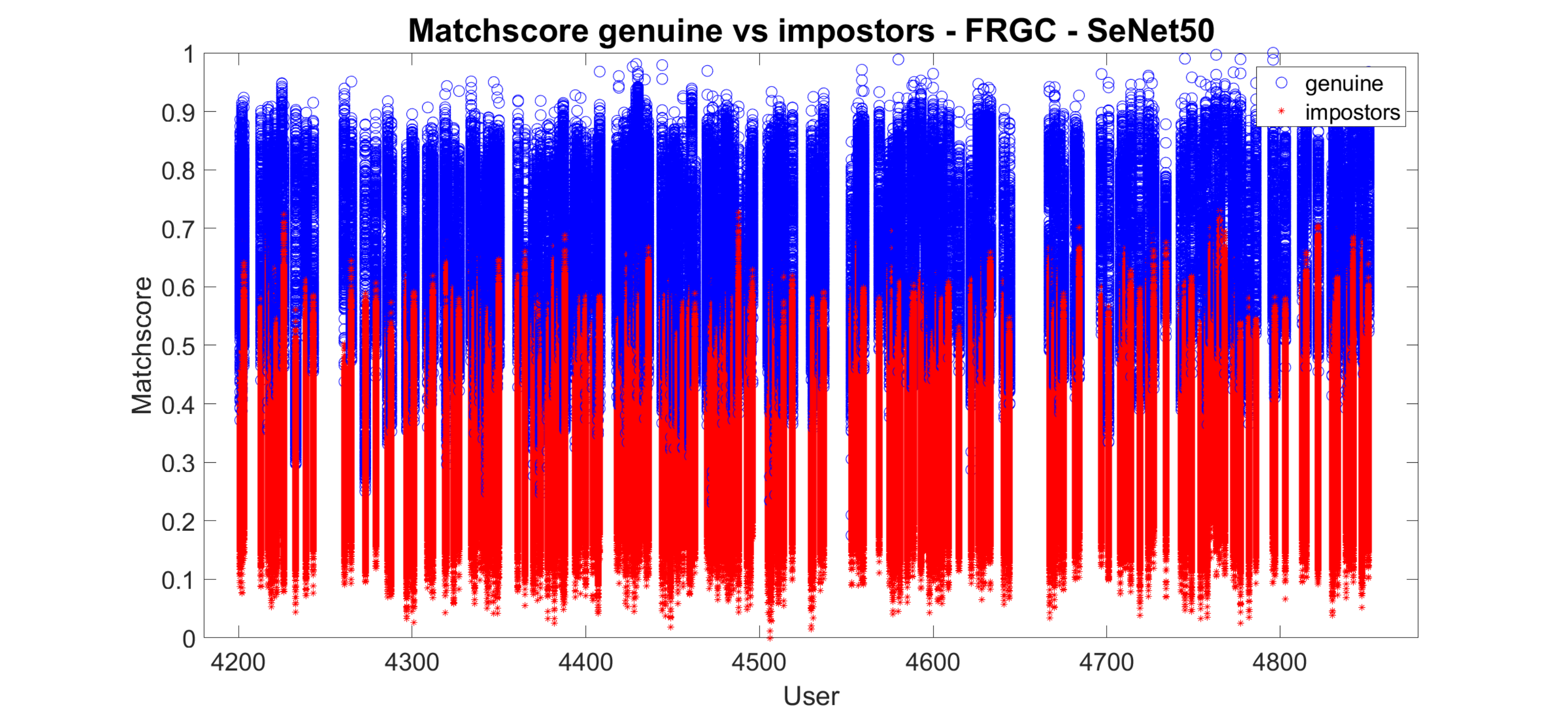}
  \caption{BSIF, FaceNet, ResNet50 and SeNet50 matching scores for the FRGC data set. The plots report in the x axis the subject identifier, whilst in the y axis the matching scores among the first five templates and the other samples are reported. Genuine users' scores are depicted in blue, the impostors' in red. Goal of these plots is to show the degree of adherence of the data with respect to the basic working hypothesis described in Section \ref{sec:clustering}.}
    \label{fig:frgcbsif}
\end{figure}

\begin{figure}
  \centering
  \includegraphics[width=.495\textwidth]{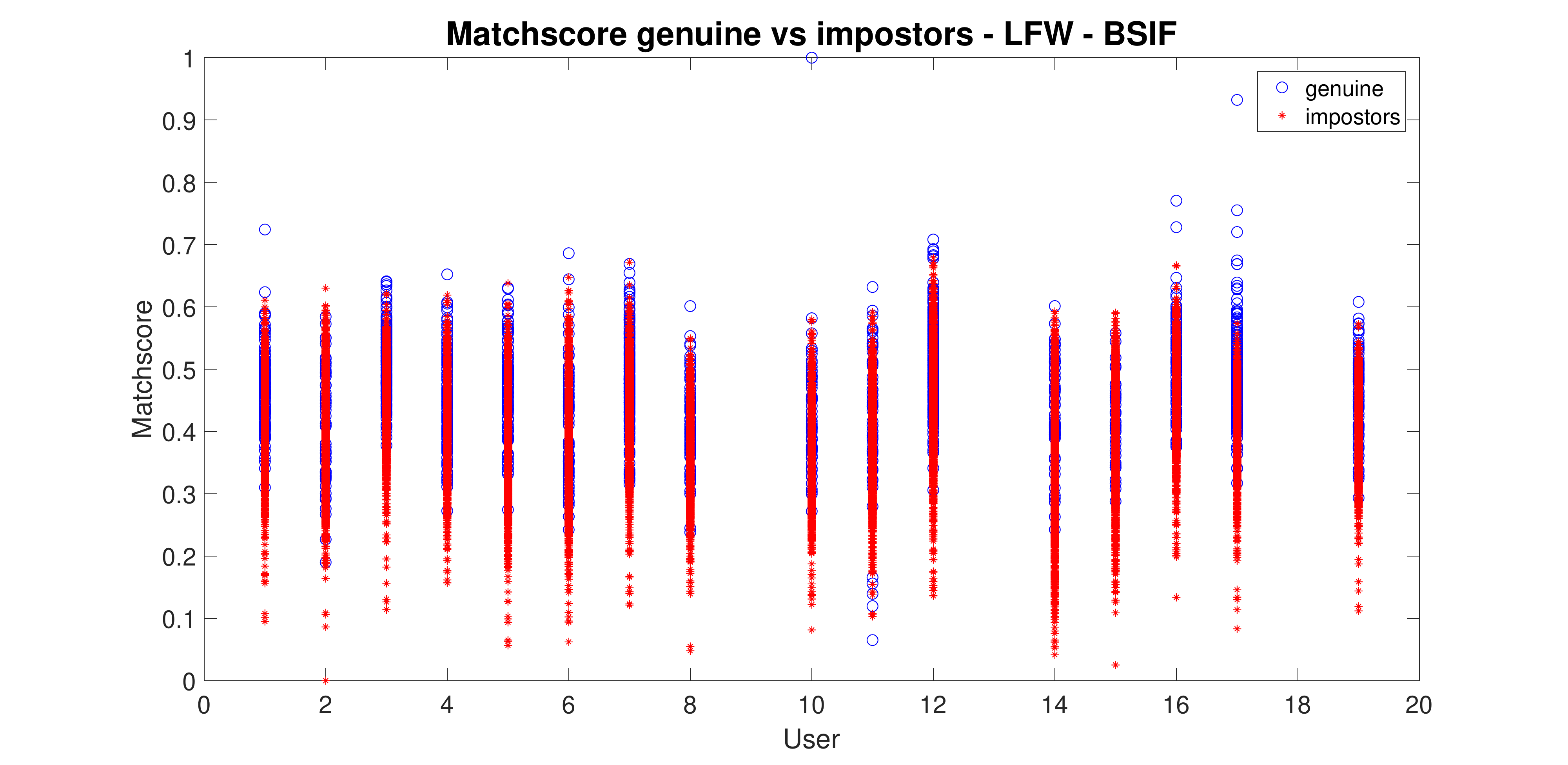}
  \includegraphics[width=.495\textwidth]{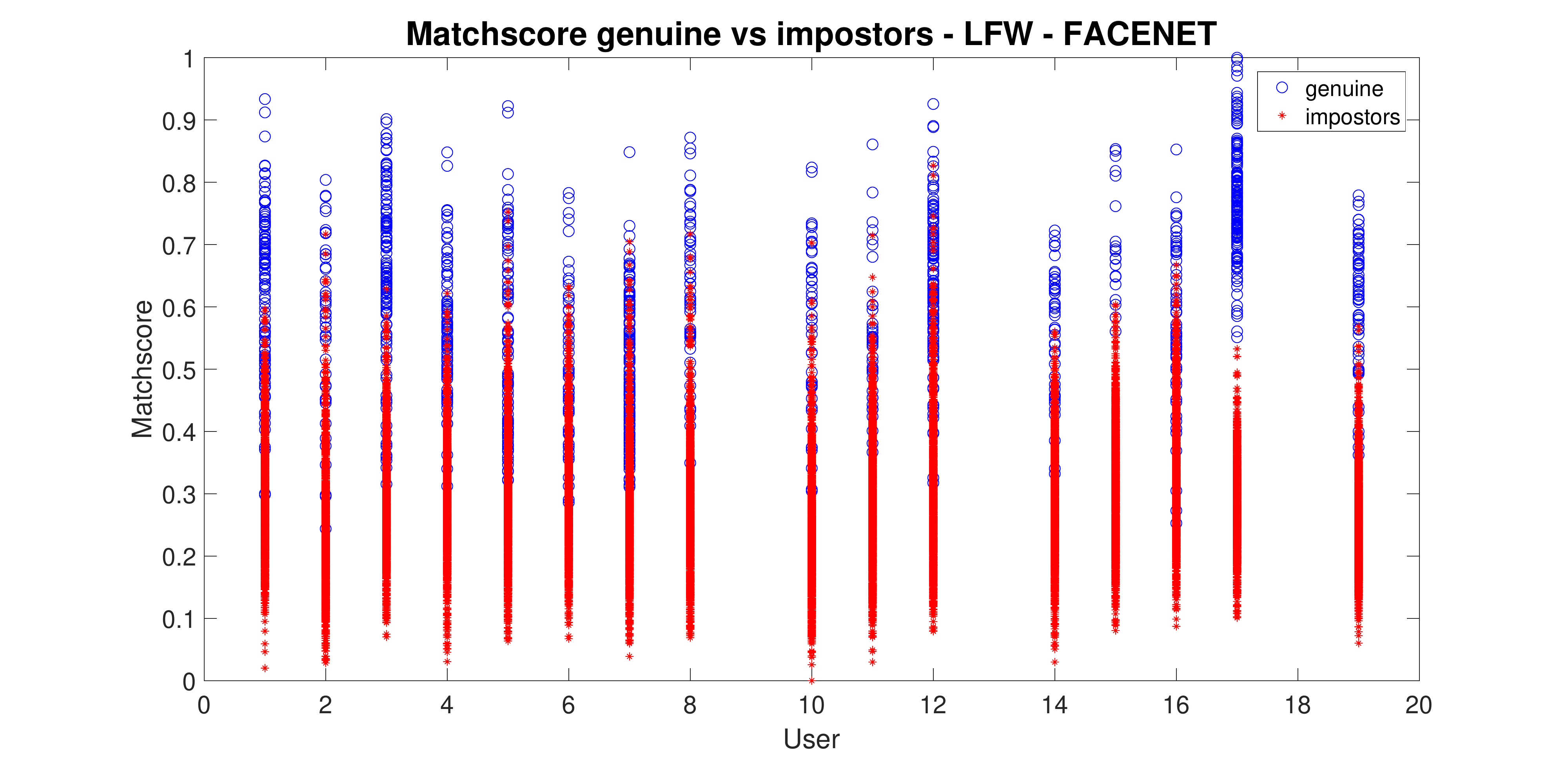}
    \includegraphics[width=.495\textwidth]{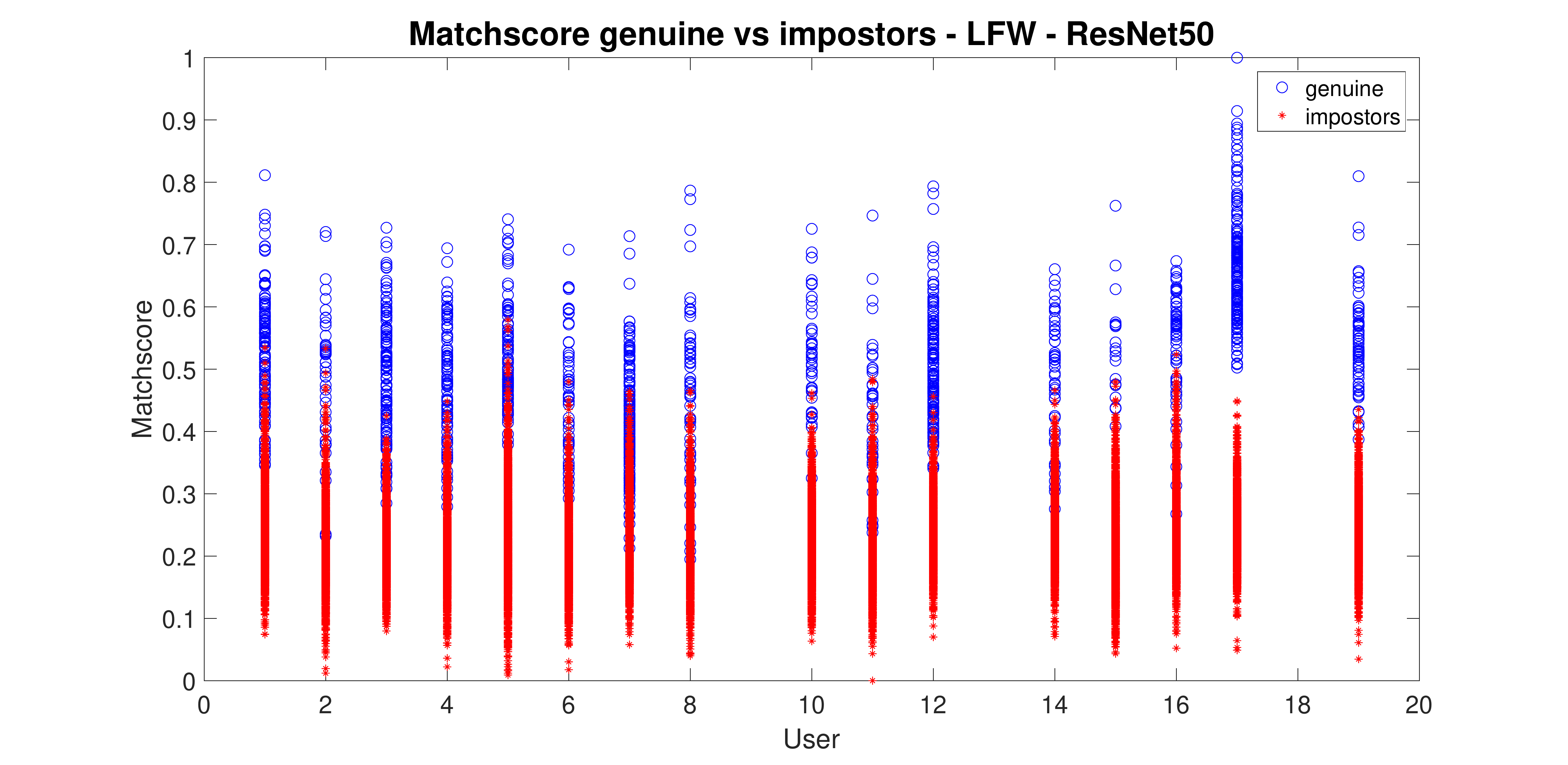}
        \includegraphics[width=.495\textwidth]{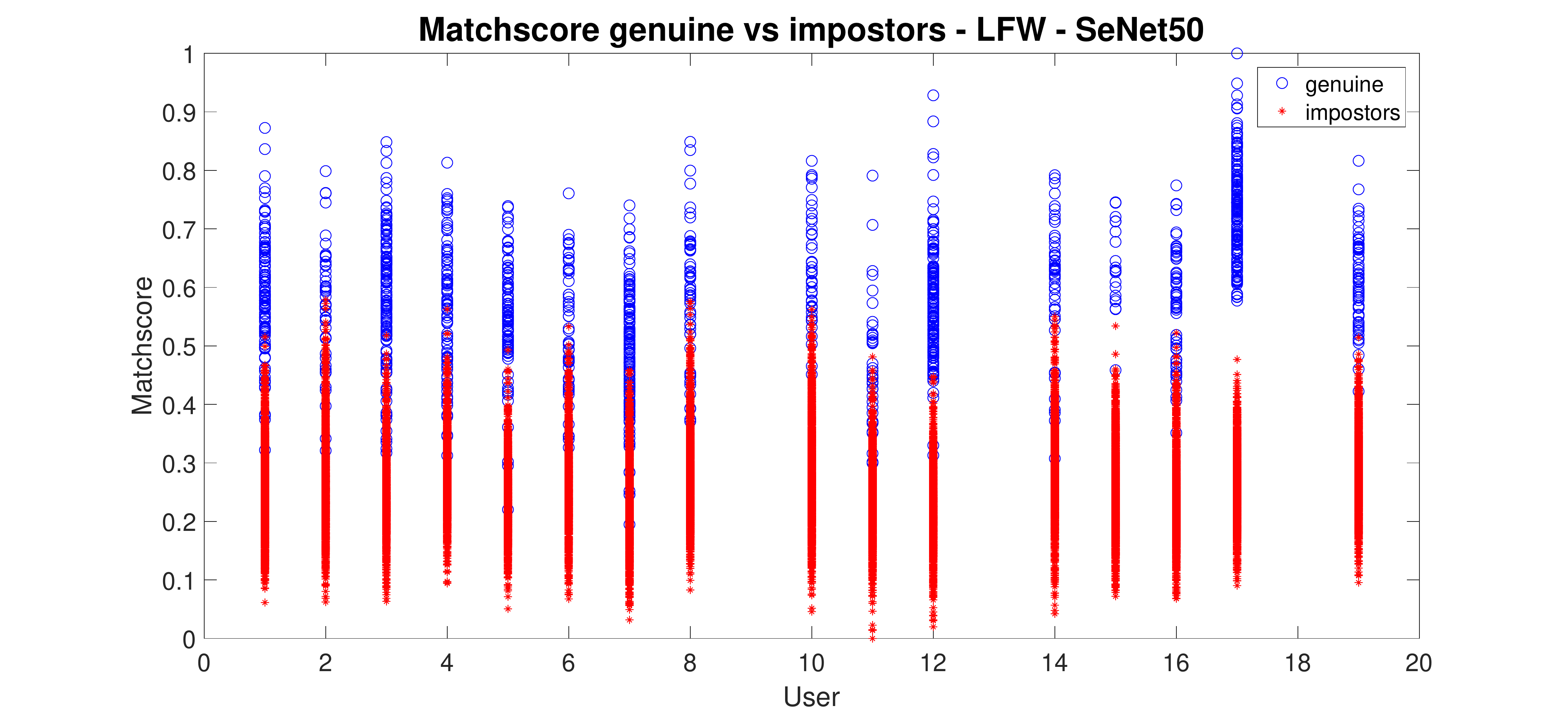}
  \caption{BSIF, FaceNet, ResNet50 and SeNet50 matching scores for the LFW data set. The plots report in the x axis the subject identifier, whilst in the y axis the matching scores among the first five templates and the other samples are reported. Genuine users' scores are depicted in blue, the impostors' in red. Goal of these plots is to show the degree of adherence of the data with respect to the basic working hypothesis described in Section \ref{sec:clustering}.}
    \label{fig:lfwbsif}
\end{figure}

We reported these scores sets per subject in Figs. \ref{fig:dieebsif}-\ref{fig:lfwbsif}, where the $x$ axis is the subject identifier. The genuine users' scores are highlighted in blue and the impostors' in red, respectively. 

The Multimodal-DIEE shows an accentuated ``separation'' between the genuine users' and the impostors' scores if we focus on the autoencoded features especially (Fig. \ref{fig:dieebsif}). BSIF tends to overlap both classes of users. We should consider that, despite their overlap, the probability of occurrence of the genuine samples in the BSIF plot exhibits a more accentuated peek than that of the probability of occurrence of the impostors in the same region. A similar effect, over a large user population, is related to subjects of the FRGC plots (Fig. \ref{fig:frgcbsif}).
DIEE and FRGC data sets may represent a partially controlled environment for face recognition; they differ in the acquisition modality of the intra- and inter-class variations. Moreover, the FRGC's user population is much larger than that of the Multi-modal DIEE, which was acquired over a larger time-span.

Finally, the BSIF feature space leads to a strong overlapping degree between the genuine users' and the impostors' distributions in the LFW plots (Fig. \ref{fig:lfwbsif}). This may be expected since the LFW data set was collected by a search over the Internet, thus representing a fully uncontrolled facial recognition environment. Instead, the representations through ResNet50 and SeNet50 are more powerful and are able to separate better the two distributions.

To sum up, all the data sets are a good challenge for both handcrafted and auto-encoded features. Being their performance far from a high accuracy, the benefits of self-updating should be pointed out, as well as the plausibility of our working hypothesis.

\subsection{Experimental protocol}

In this Section, we describe the experimental protocol and the performance measurements used in this paper.
Each data set was randomly divided into seven batches. The first batch was used as the initial gallery (training set), the last was used as test set in order to evaluate the performance of the system and the other five were used as adaptation sets in order to simulate periodic system's update. According to \cite{Roli2006}, an independent test set is useful to have the same reference for all update cycles.
It has also been proposed the use of the batch $i+1$ as test set of the batch $i$  \cite{Rattani2013AMD}. Since not all used data sets have temporally labelled acquisitions, we followed the independent test set approach.\\ 
The initial number of templates in the gallery is defined by the parameter $p$. The value $p$ also defines the number of samples per user present in each batch. 
For the Multimodal-DIEE and FRGC data sets we used $p \in \{5, 6, 7\}$, and $p=4$ for the LFW data set due to its small size.\\
In order to evaluate the performance, the Equal Error Rate (EER) and the percentage of impostors wrongly added were calculated. 
\\
Each experiment was repeated 10 times and the results are averaged over those runs.
All the experiments were performed with a desktop PC with operating system Windows 7 Professional 64bit, 32 GB RAM using MATLAB v.R2013a.

\subsection{Results}

In this Section, we show the performance of the proposed and the other state-of-the-art methods using handcrafted (BSIF) and auto-encoded (FaceNet, ResNet50 and SeNet50) features.\\For Multimodal-DIEE and FRGC data sets we tested three possible values for $p$, 5, 6 and 7, which represent different constraints in terms of storage size. The results did not differ significantly for different $p$ values. Reported results refer to $p=6$.
We plotted the performance measurements against the batch numbers. For the EER plots the performance of a system without update was reported (in gray).

\begin{figure}[htbp]
\centering%
\subfigure{
\includegraphics[scale=0.142]{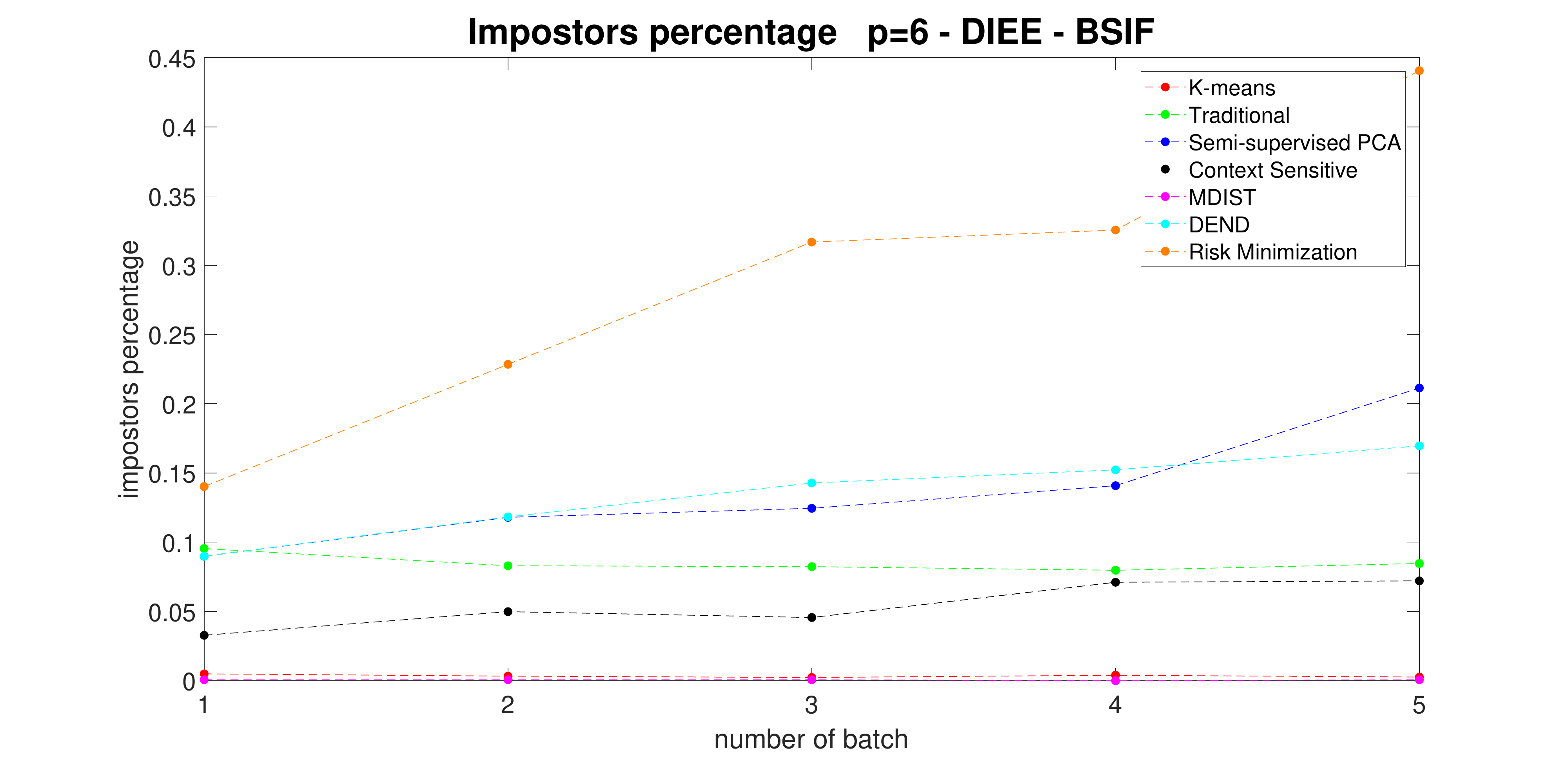}
}
\subfigure{
\includegraphics[scale=0.142]{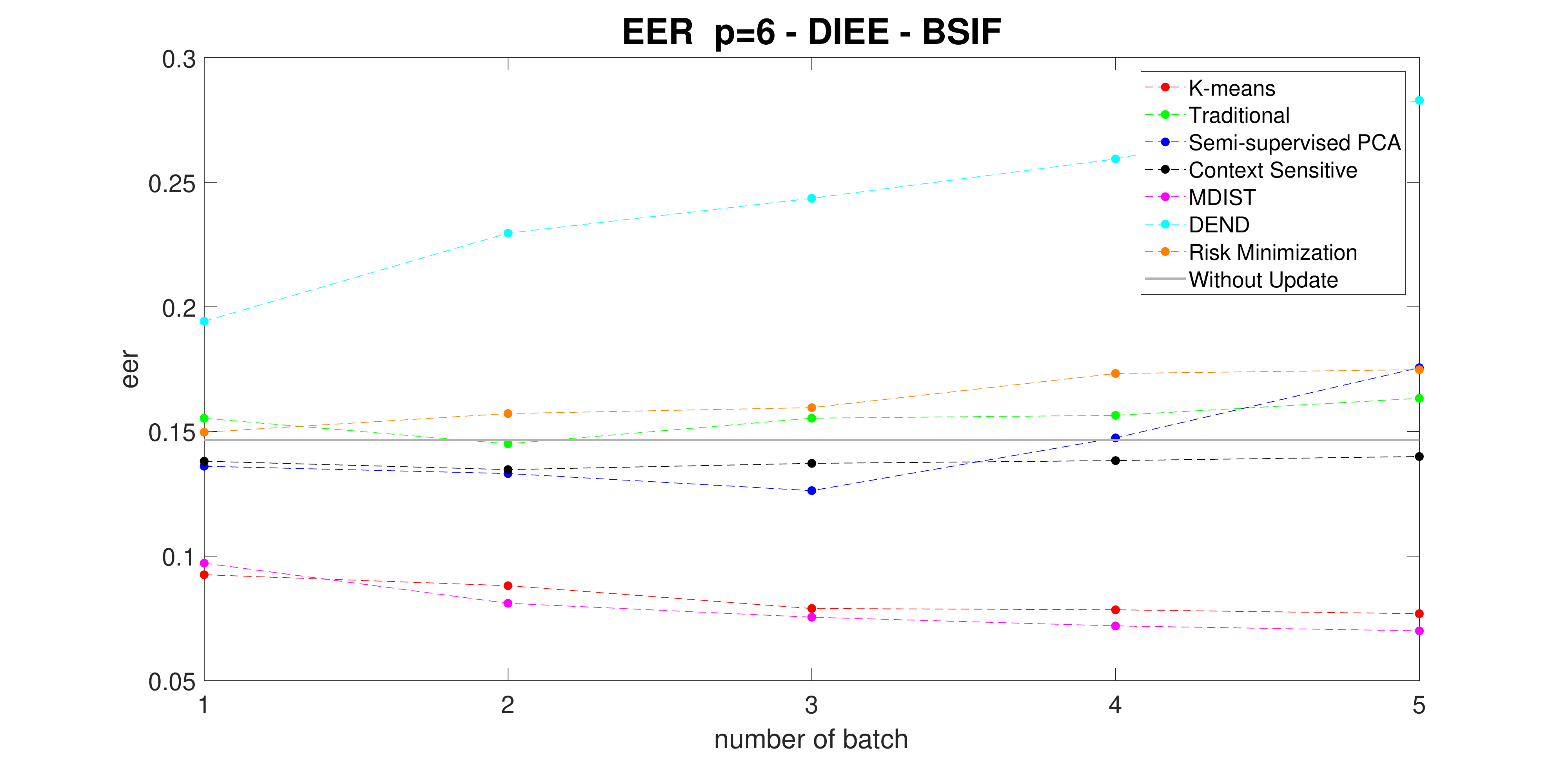} 
}
\subfigure{
\includegraphics[scale=0.142]{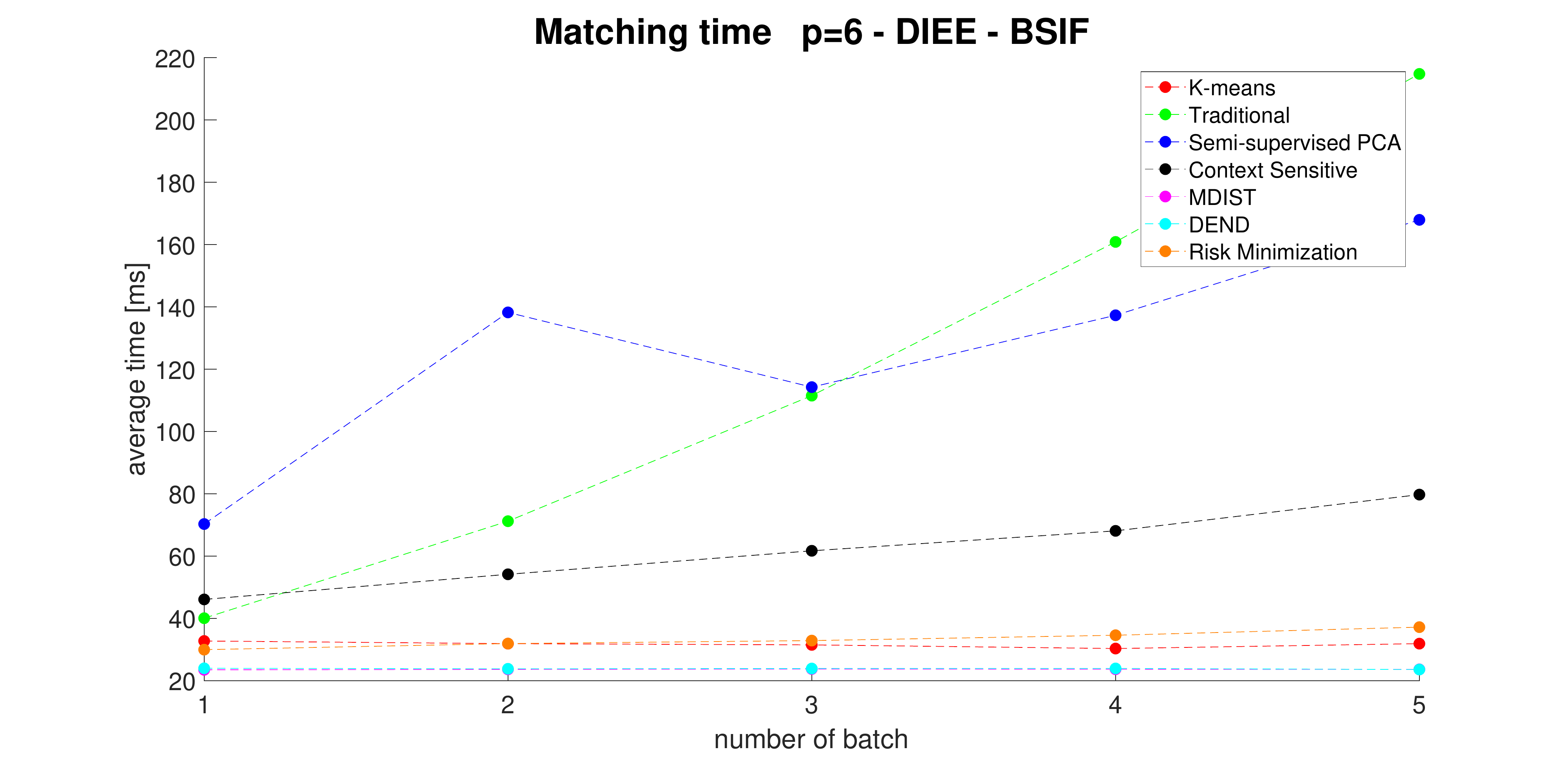}}
\caption{EER, percentage impostors and matching time comparison among the state of the art and the new proposed method with p=6 for Multimodal-DIEE using BISF handcrafted features. On the x axis is shown the number of the batch and on the y axis the performance index.}
\label{diee6}
\end{figure}

\begin{figure}[htbp]
\centering%
\subfigure{
\includegraphics[width=.495\textwidth]{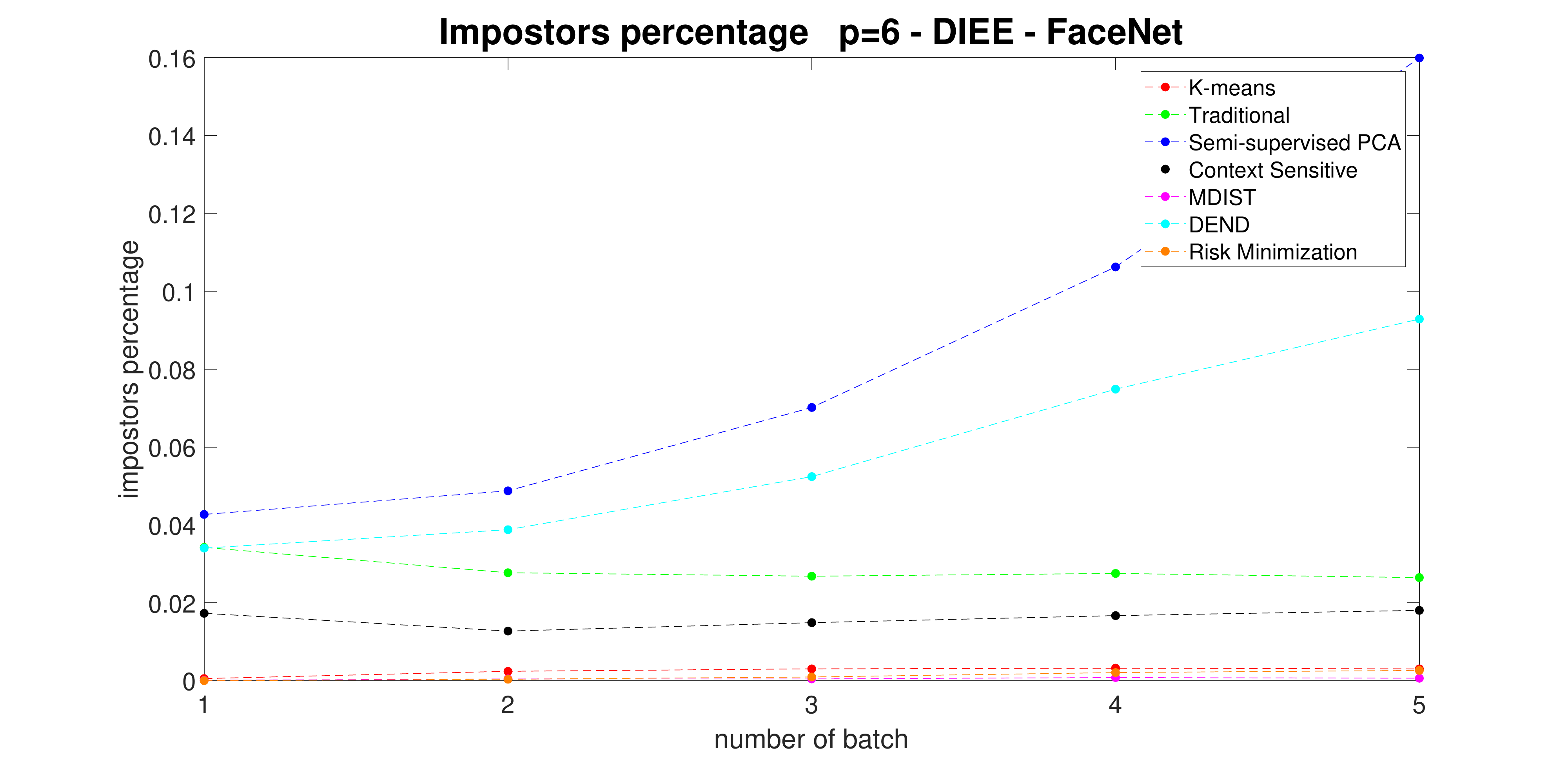}
}
\subfigure{
\includegraphics[scale=0.142]{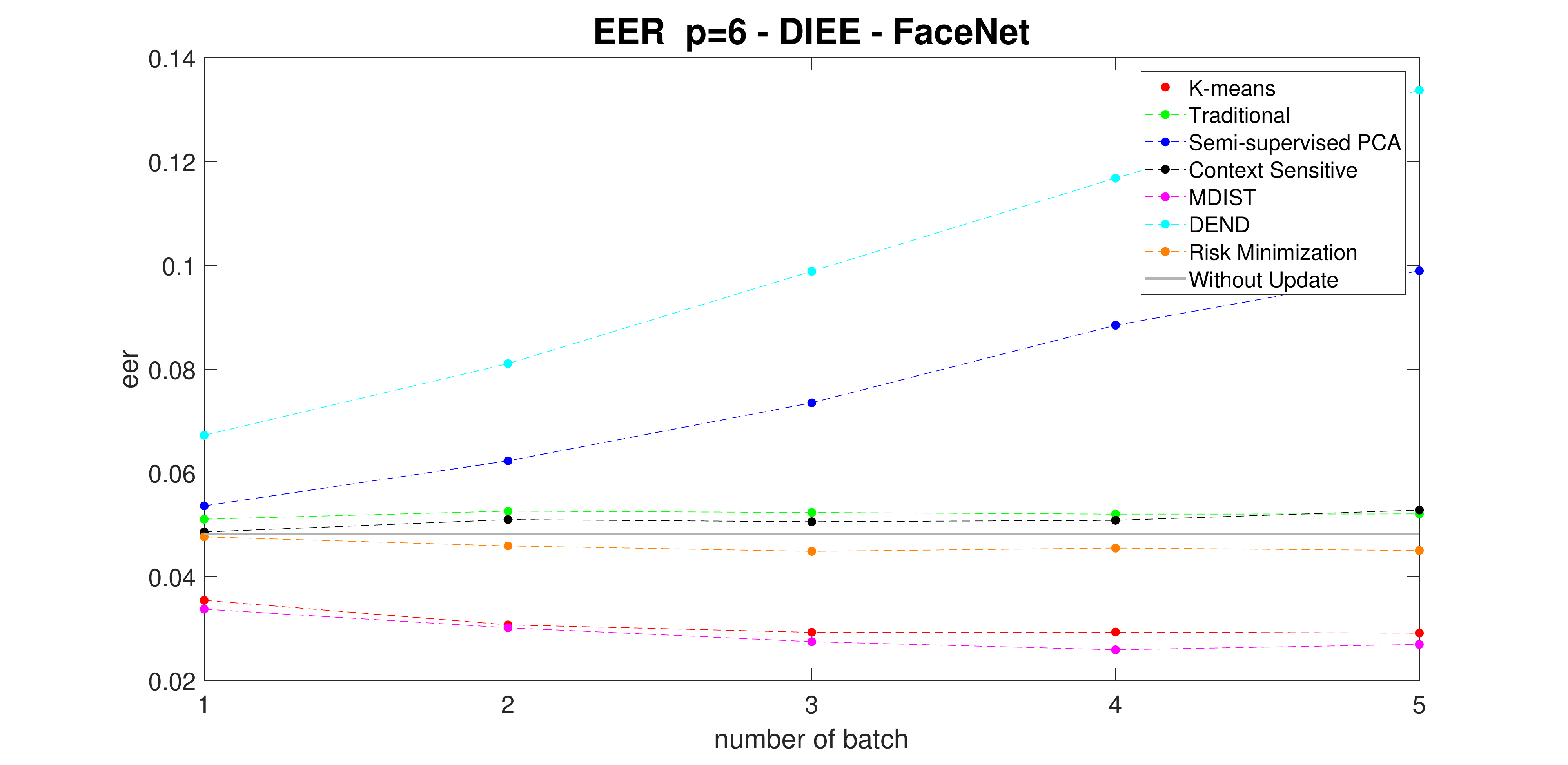}
}
\subfigure{
\includegraphics[scale=0.142]{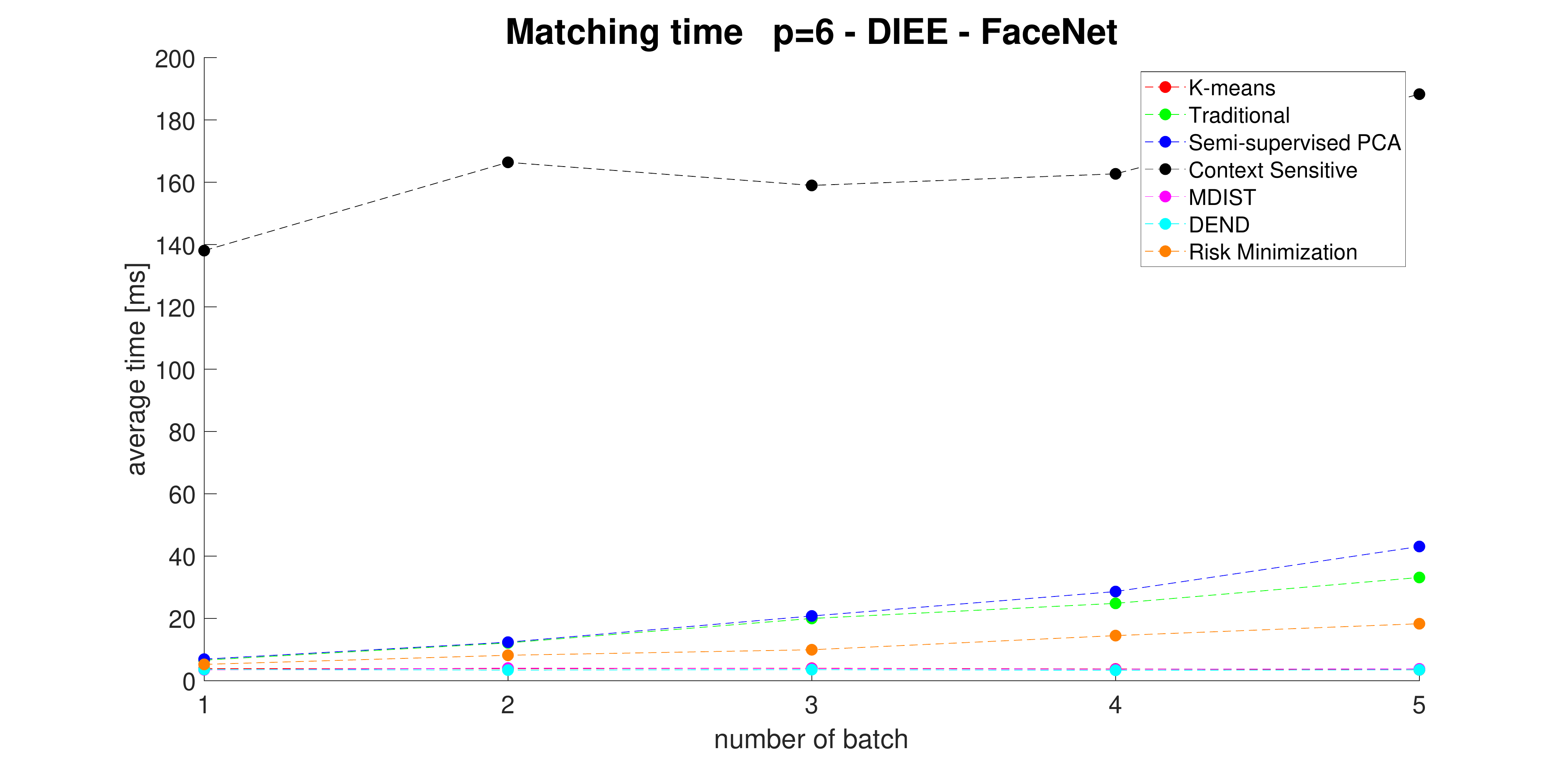}}
\caption{EER, percentage impostors and matching time comparison among the state of the art and the new proposed method with p=6 for Multimodal-DIEE using FaceNet auto-encoded features. On the x axis is shown the number of the batch and on the y axis the performance index.}
\label{diee6face}
\end{figure}

\begin{figure}[htbp]
\centering%
\subfigure{
\includegraphics[scale=0.142]{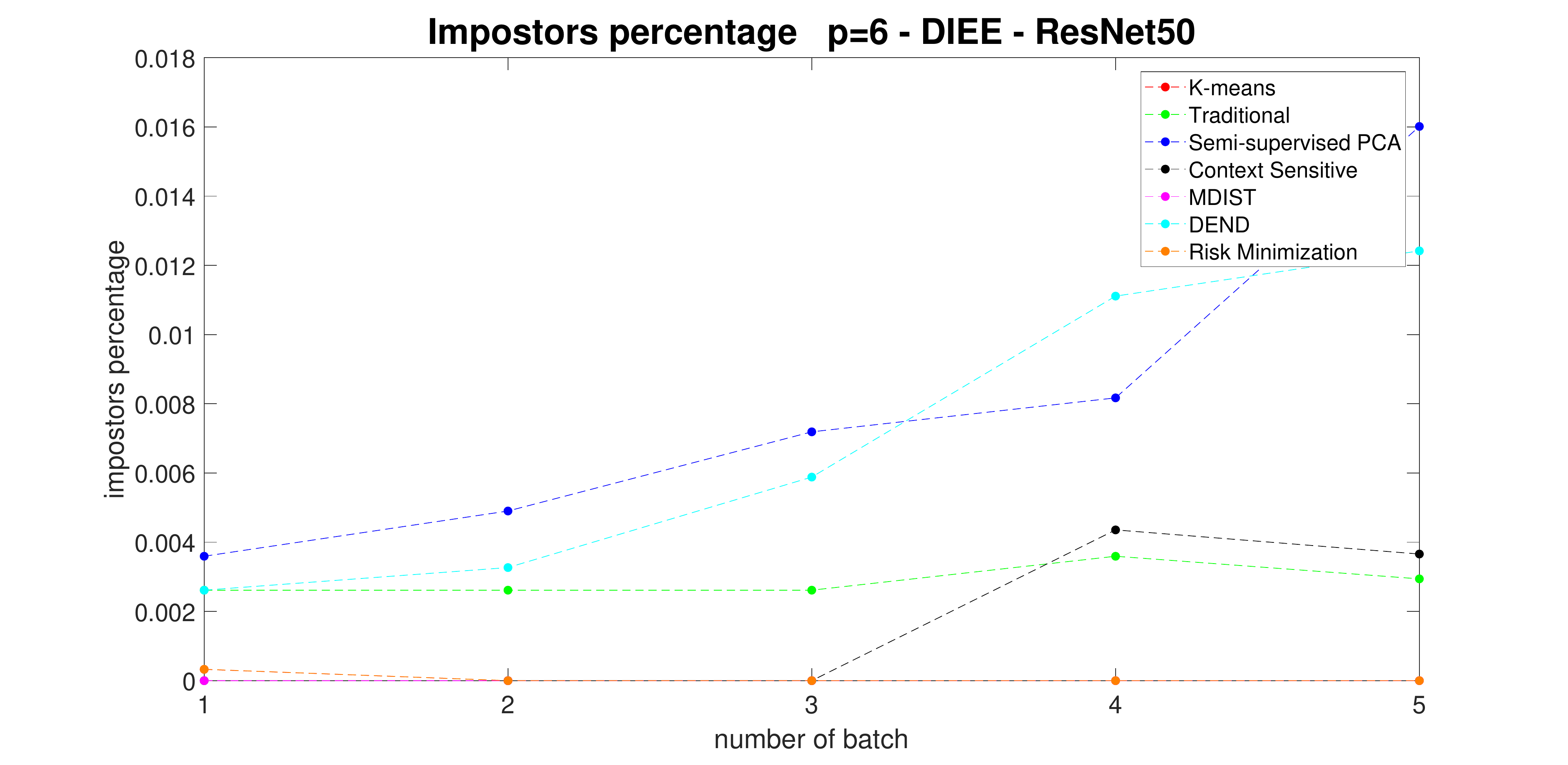}
}
\subfigure{
\includegraphics[scale=0.142]{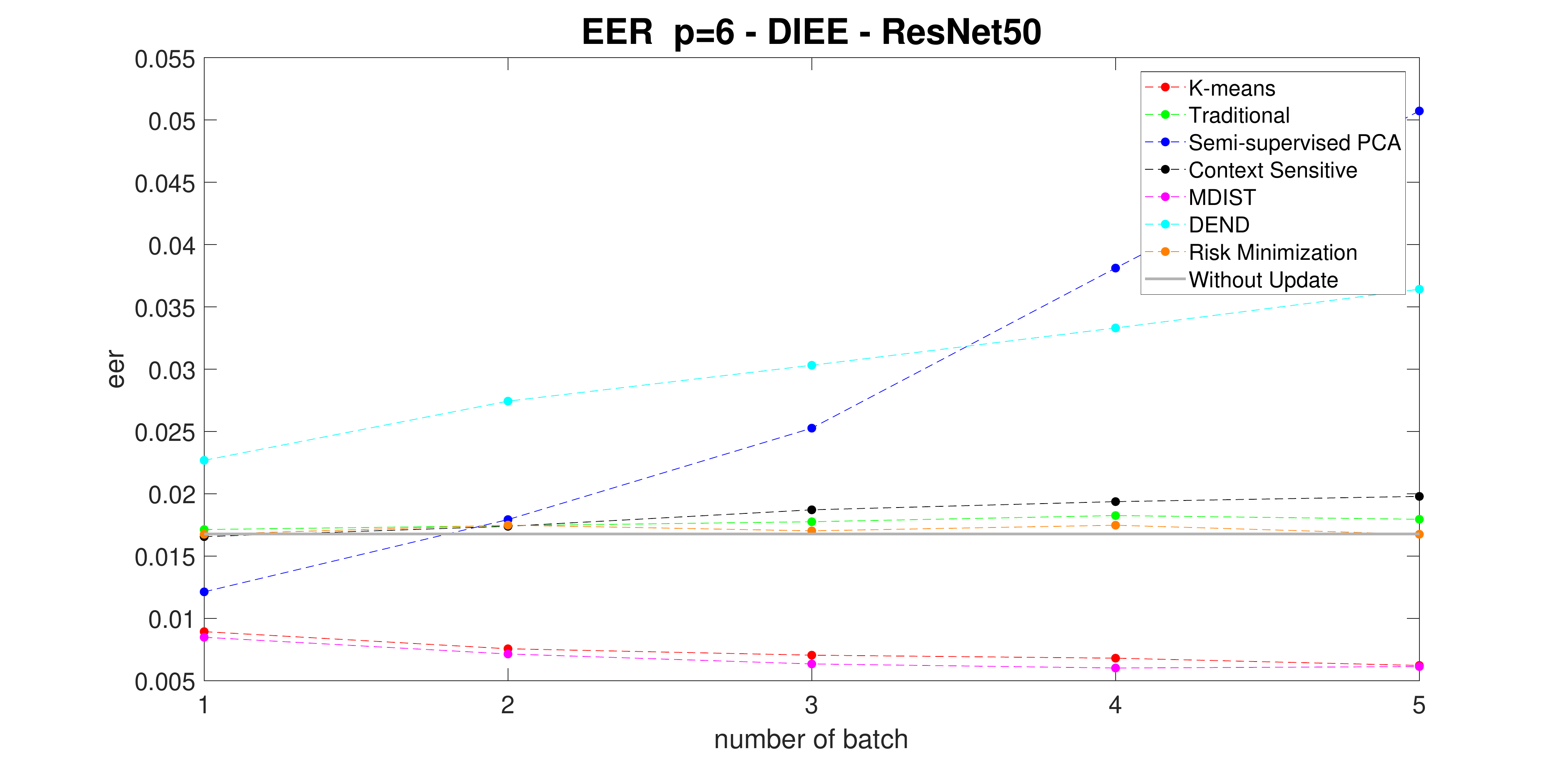} 
}
\subfigure{
\includegraphics[scale=0.142]{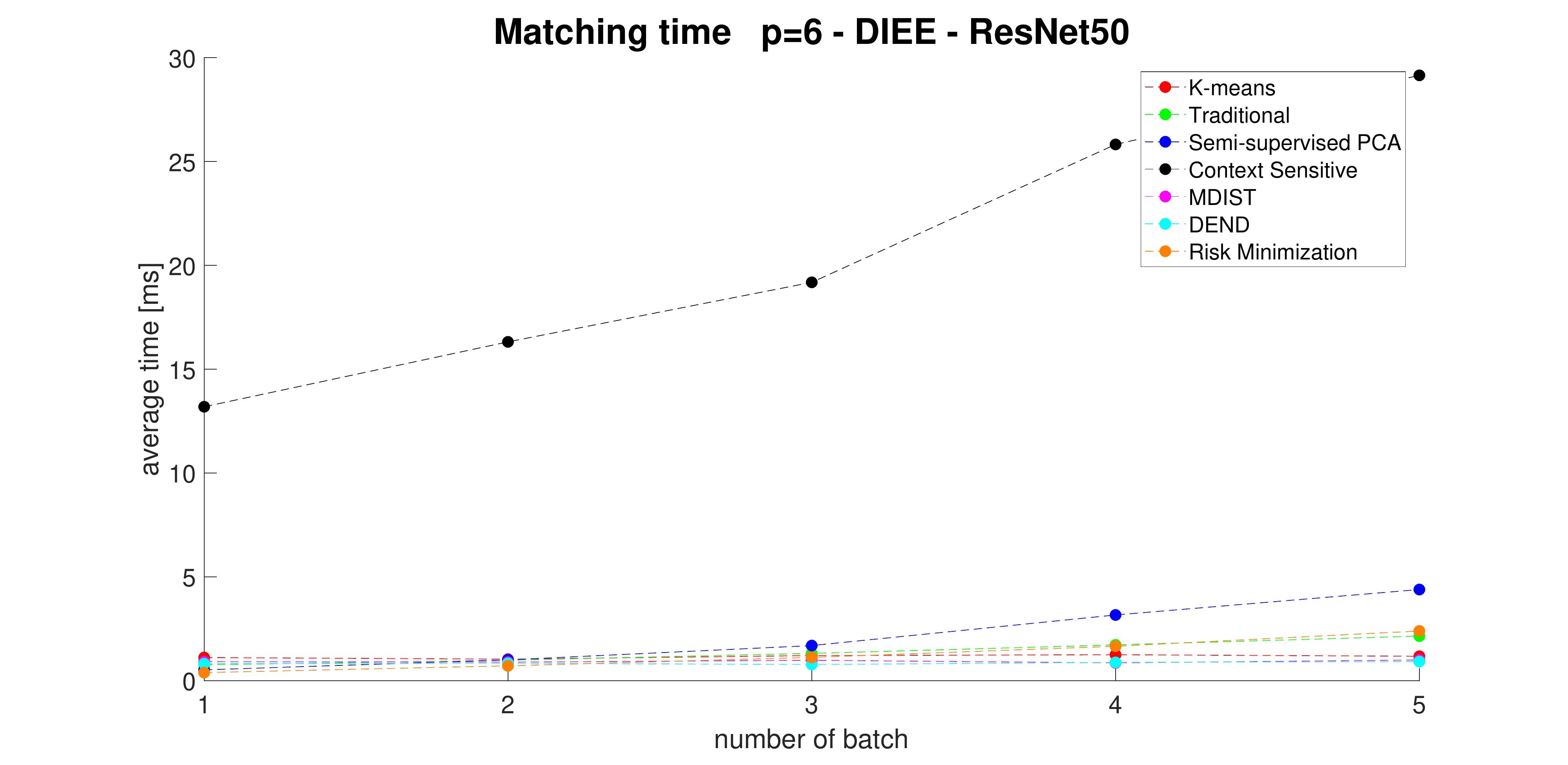}}
\caption{EER, percentage impostors and matching time comparison among the state of the art and the new proposed method with p=6 for Multimodal-DIEE using ResNet50 auto-encoded features. On the x axis is shown the number of the batch and on the y axis the performance index.}
\label{diee6resnet}
\end{figure}

\begin{figure}[htbp]
\centering%
\subfigure{
\includegraphics[scale=0.142]{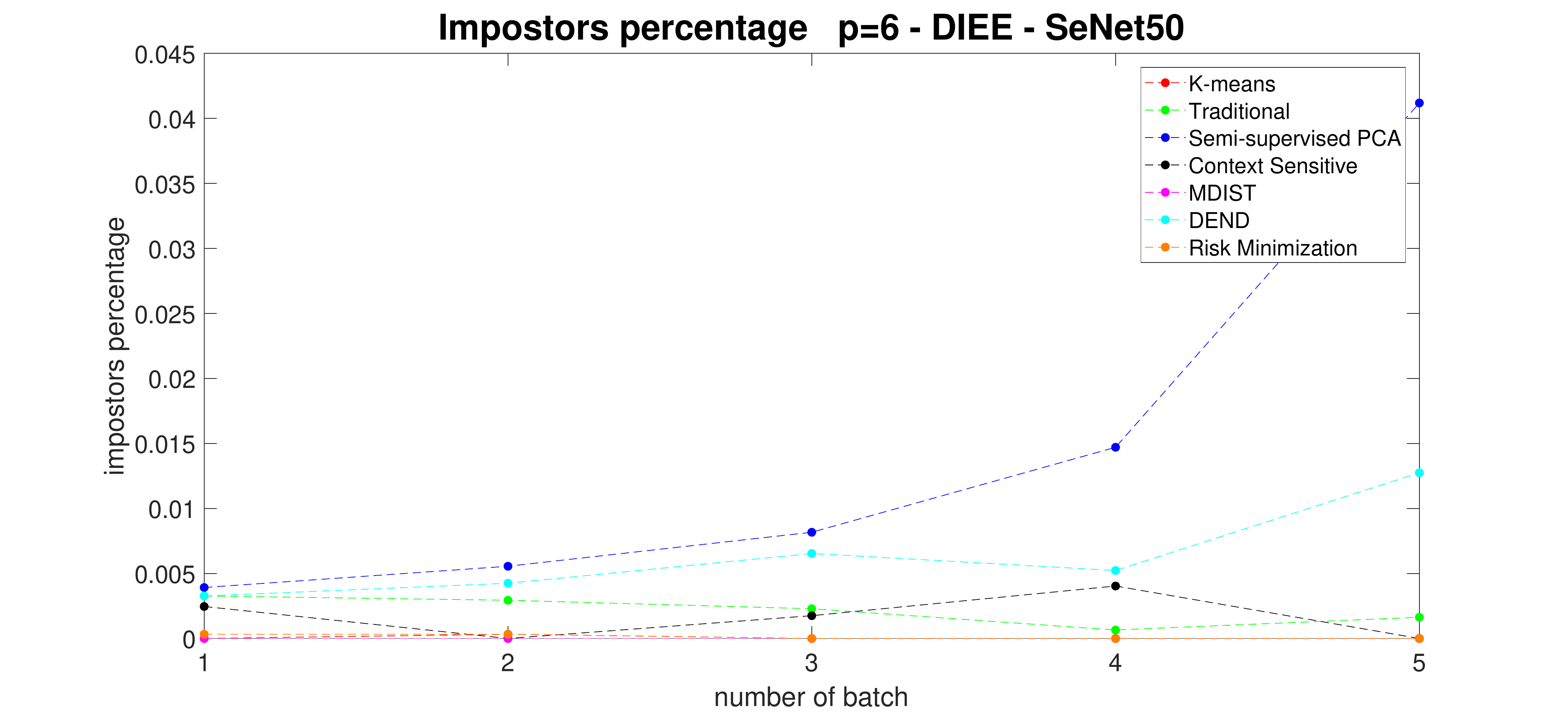}
}
\subfigure{
\includegraphics[scale=0.142]{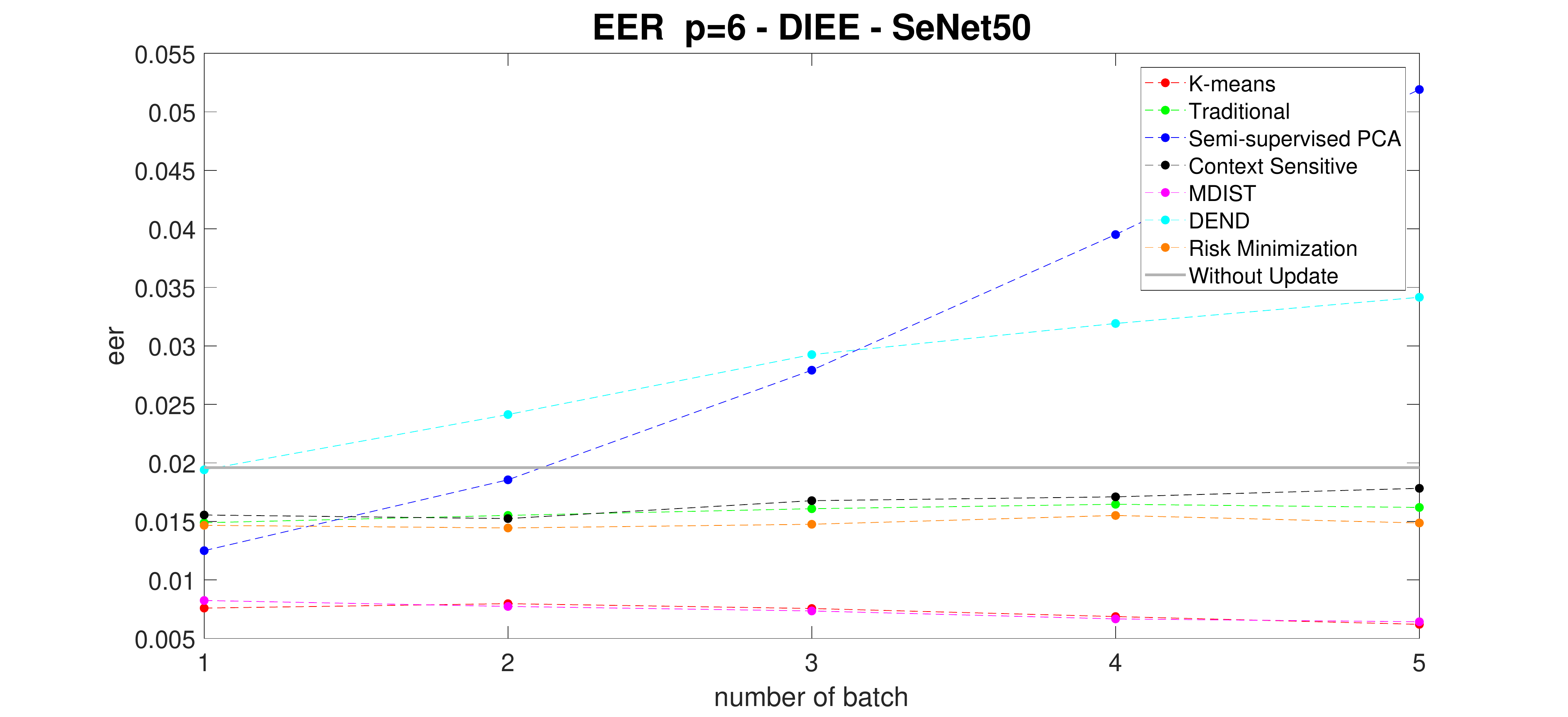}
}
\subfigure{
\includegraphics[scale=0.142]{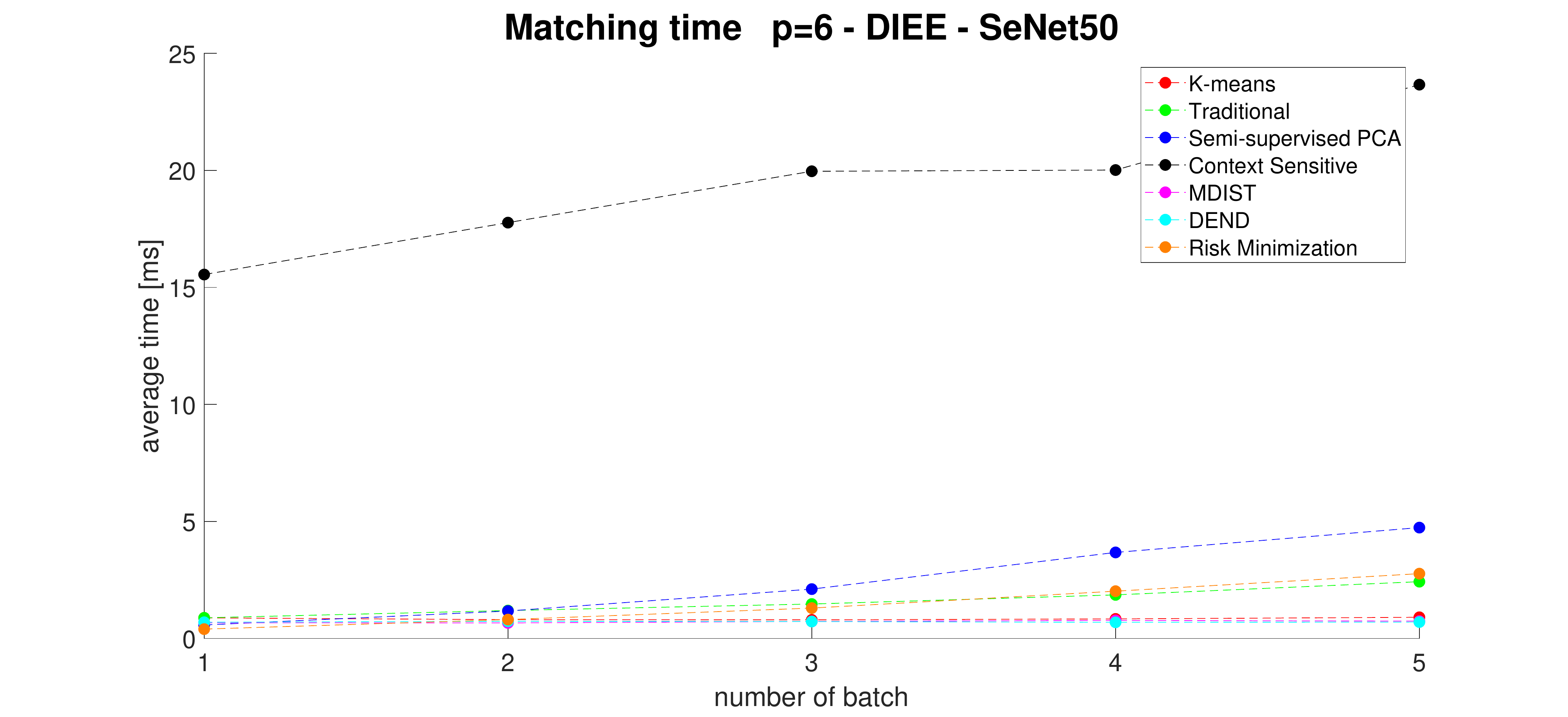}}
\caption{EER, percentage impostors and matching time comparison among the state of the art and the new proposed method with p=6 for Multimodal-DIEE using SeNet50 auto-encoded features. On the x axis is shown the number of the batch and on the y axis the performance index.}
\label{diee6senet}
\end{figure}

Figs. \ref{diee6},\ref{diee6face},\ref{diee6resnet},\ref{diee6senet} show the average values of impostors' percentage, the EER and the processing time for the Multimodal-DIEE data set using, respectively, BSIF, FaceNet, ResNet50 and SeNet50 features. 

By these plots, it is possible to notice that the proposed algorithms exhibit the impostors' percentage in the gallery close to zero for all batches. Therefore, they manage to exploit the initial hypothesis: the closest samples each other are close to the main mode too. As counterproof, the DEND method leads to a high number of impostors. Despite the limited number of templates per user, the EER is low for both the proposed algorithms, especially in the case of the SeNet50 feature set (Fig. \ref{diee6senet}).
For all investigated feature sets, we can notice a significant improvement over the basic performance of the system (in gray). This points out the usefulness of the self-updating approach, beside a very low percentage of impostors into the galleries. Furthermore, the state-of-the-art self-updating methods are outperformed.

\begin{figure}[htbp]
\centering%
\subfigure{
\includegraphics[scale=0.142]{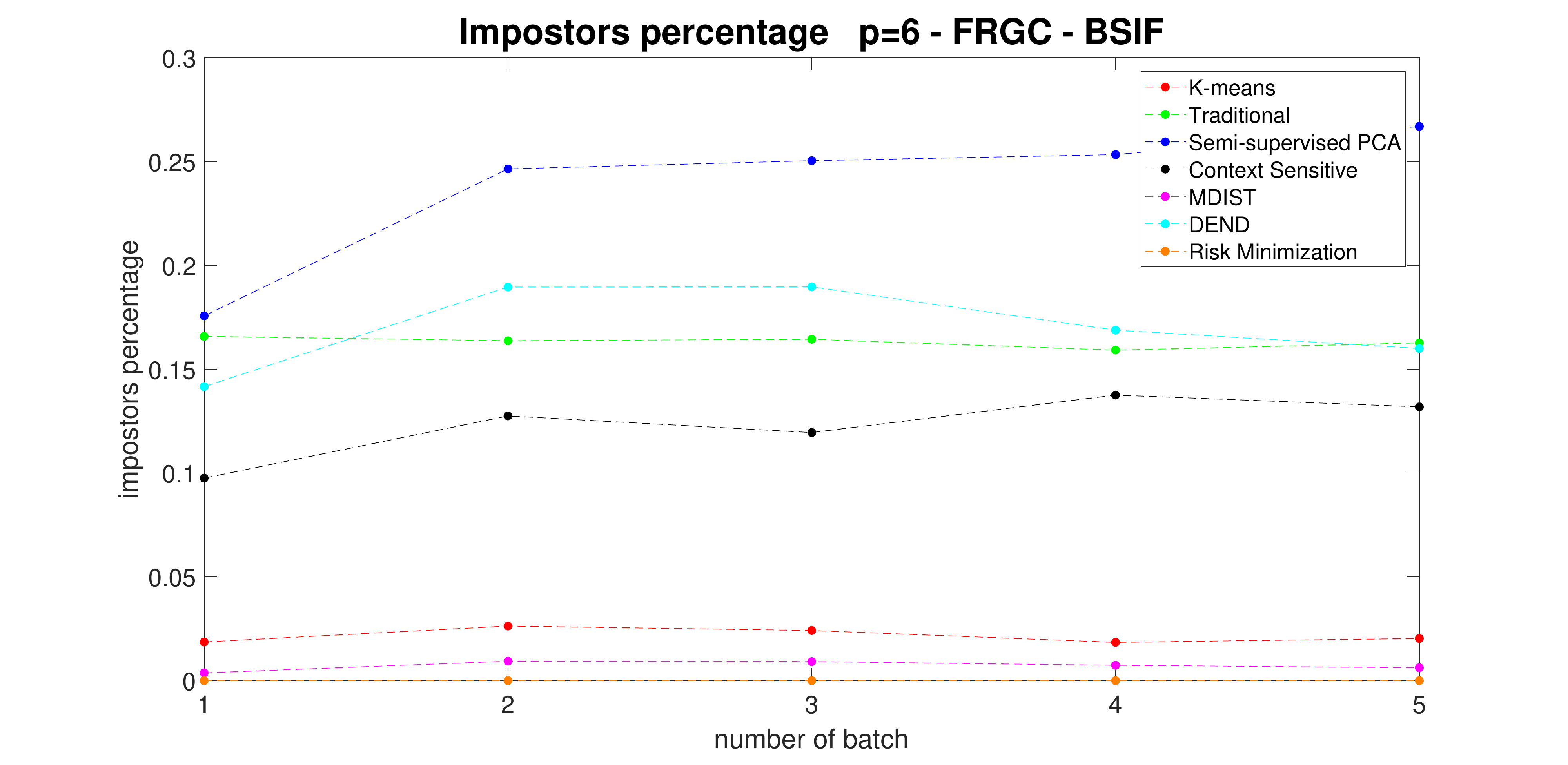}   
}
\subfigure{
\includegraphics[scale=0.142]{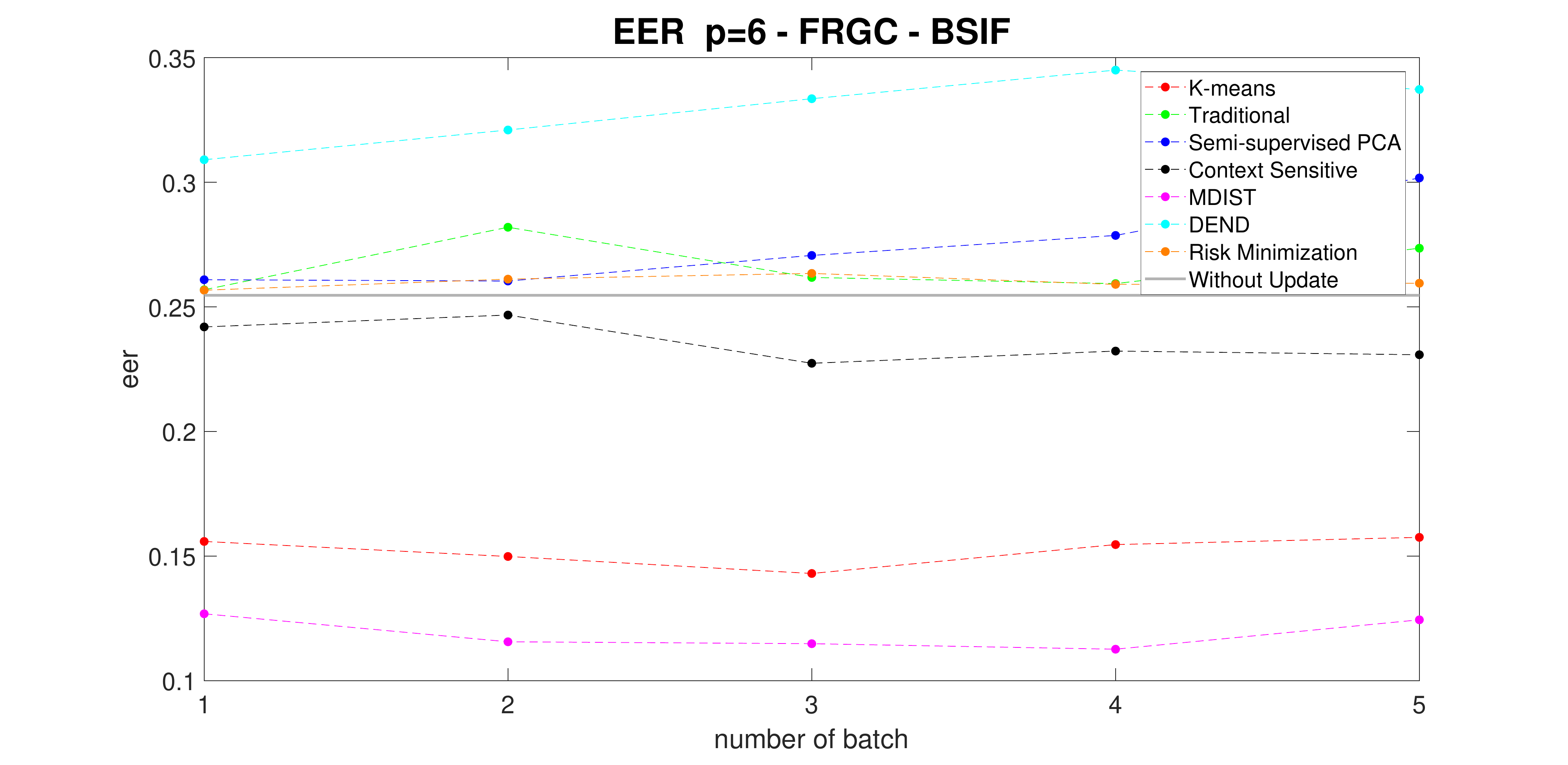}
}
\subfigure{
\includegraphics[scale=0.142]{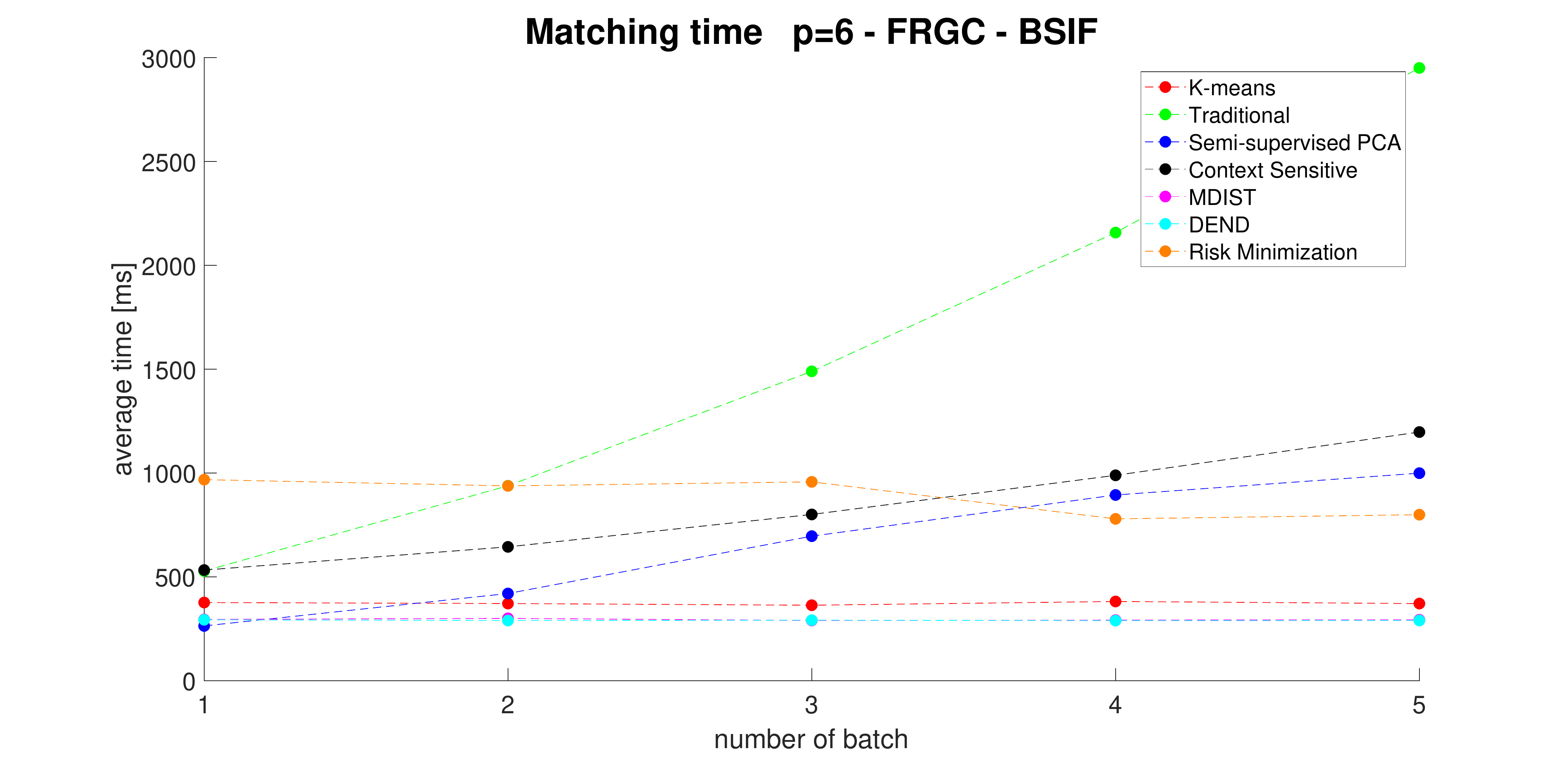}
}
\caption{EER, percentage impostors and matching time comparison among the state of the art and the new proposed method with p=6 for FRGC using BISF handcrafted features. On the x axis is shown the number of the batch and on the y axis the performance index. }
\label{frgc6}
\end{figure}

\begin{figure}[htbp]
\centering%
\subfigure{
\includegraphics[scale=0.142]{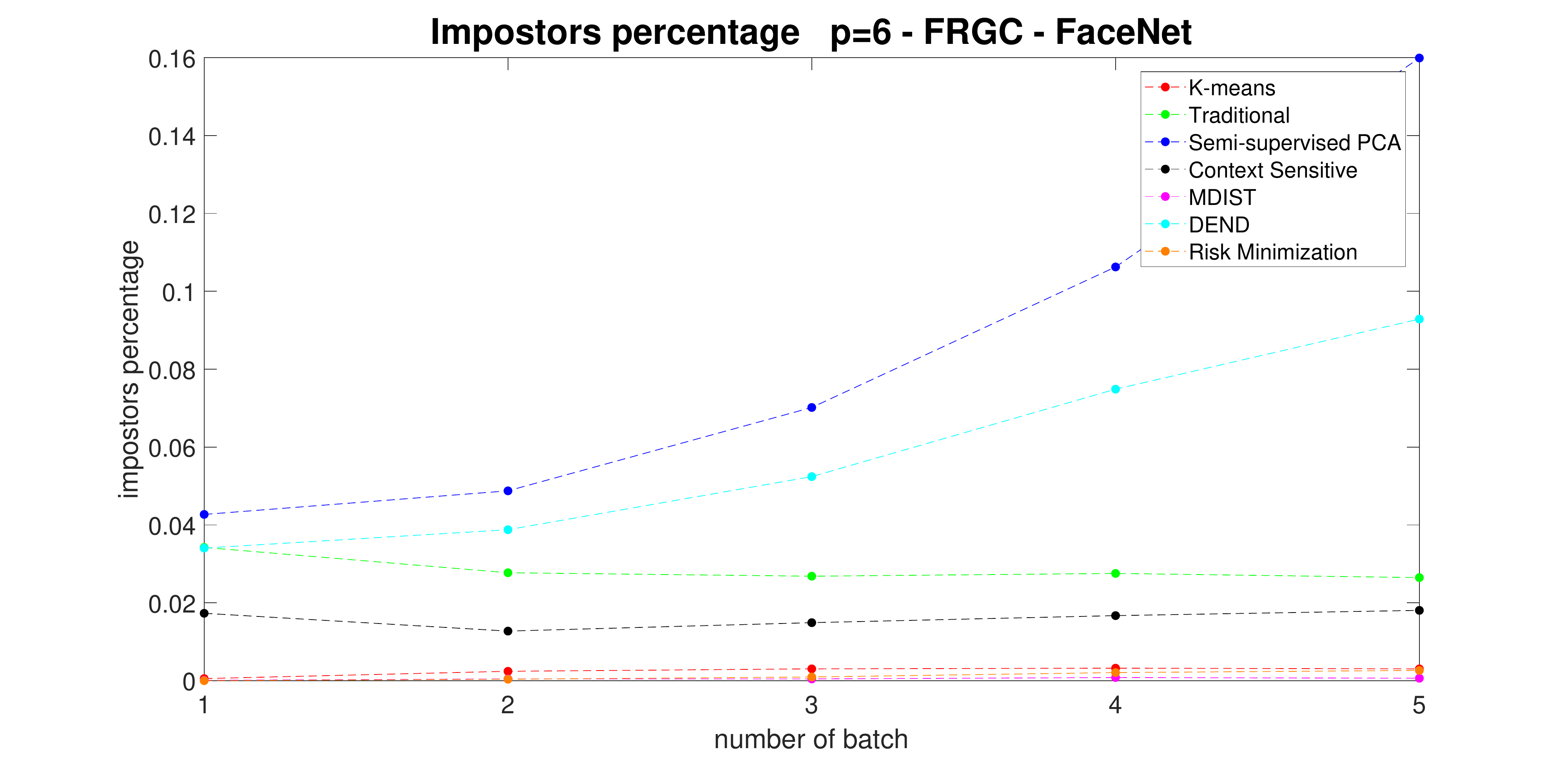}
}
\subfigure{
\includegraphics[scale=0.142]{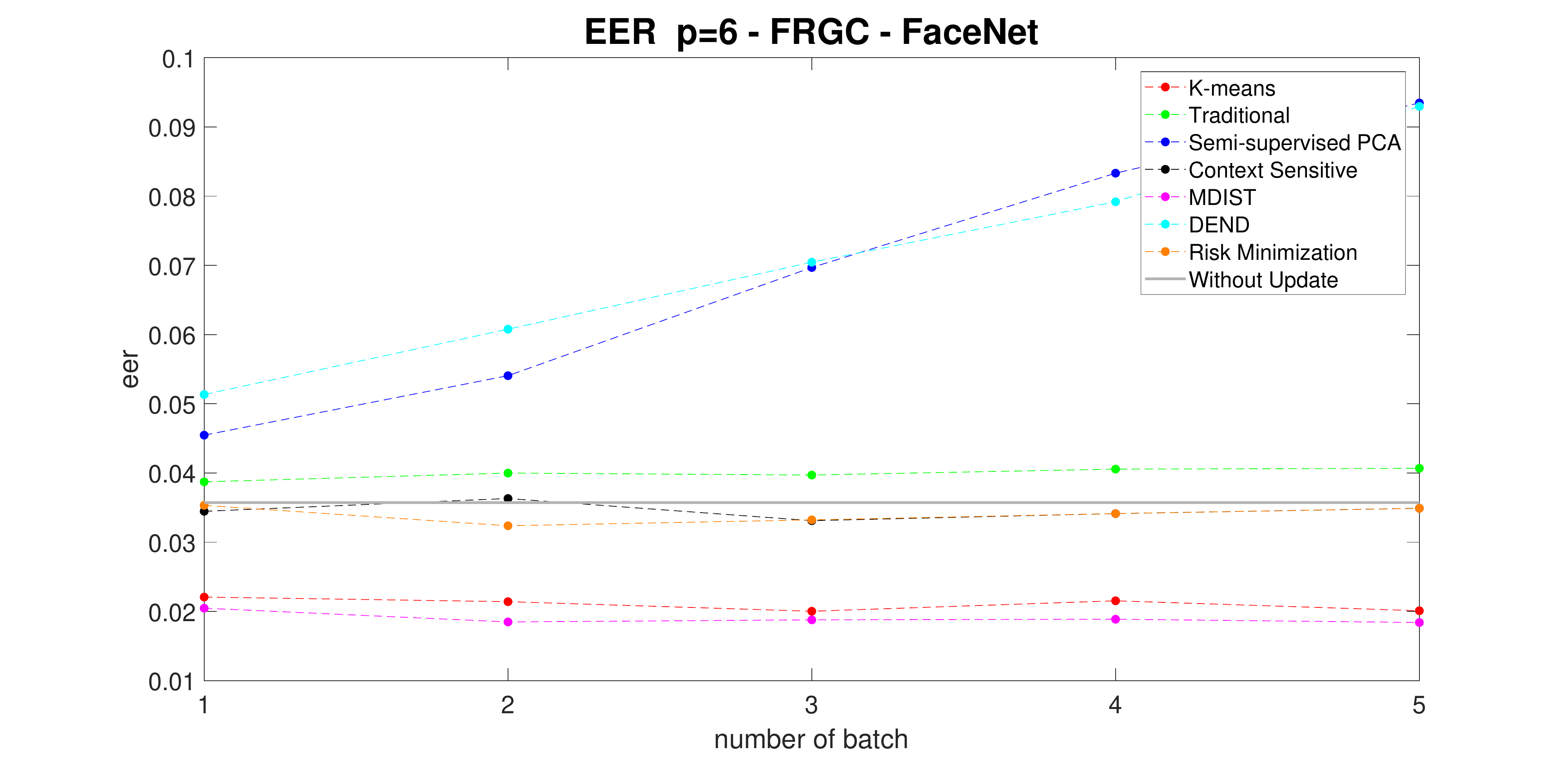}
}
\subfigure{
\includegraphics[scale=0.142]{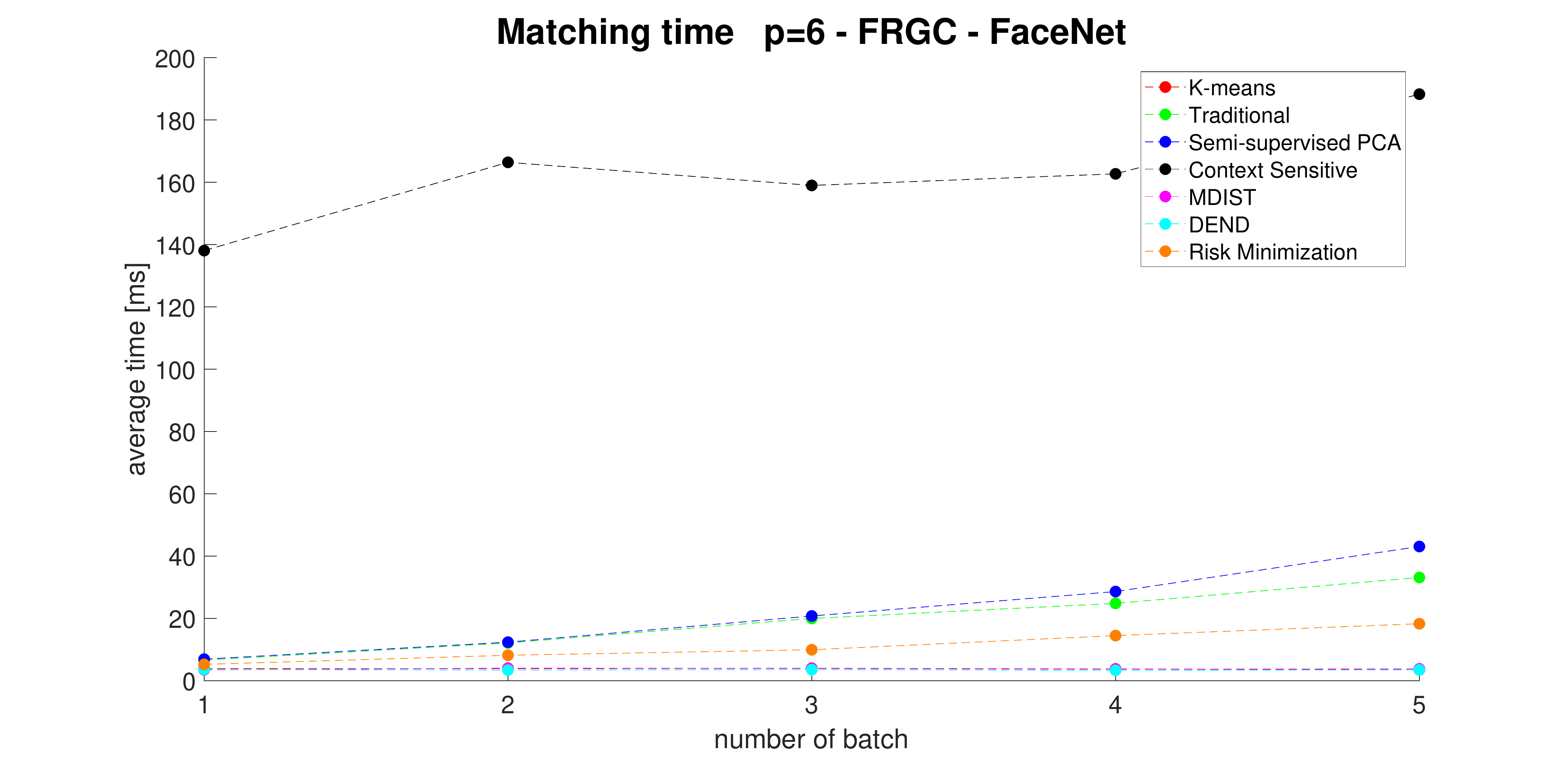}}
\caption{EER, percentage impostors and matching time comparison among the state of the art and the new proposed method with p=6 for FRGC using FaceNet auto-encoded features. On the x axis is shown the number of the batch and on the y axis the performance index.}
\label{frgc6face}
\end{figure}

\begin{figure}[htbp]
\centering%
\subfigure{
\includegraphics[scale=0.142]{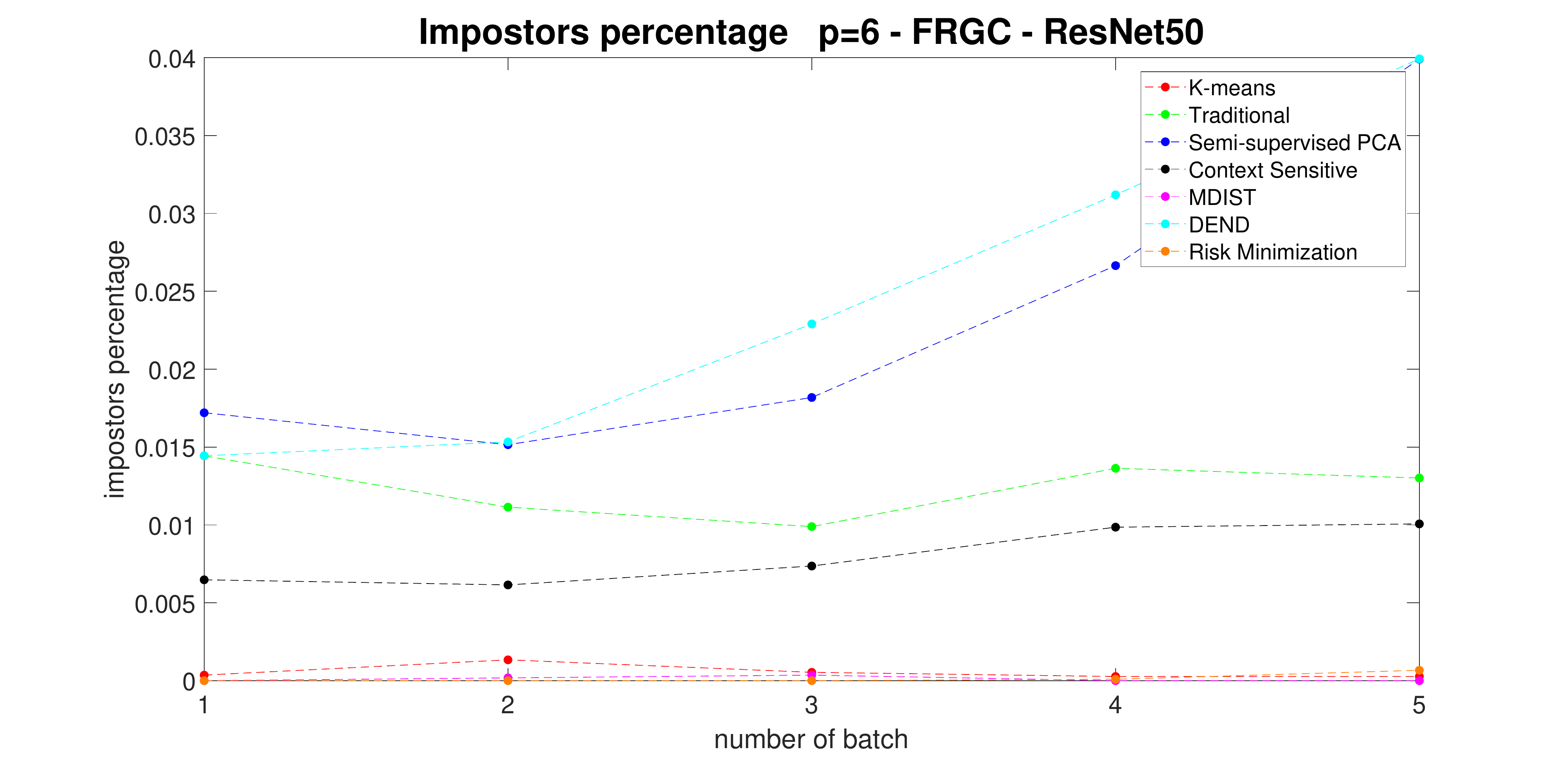}
}
\subfigure{
\includegraphics[scale=0.142]{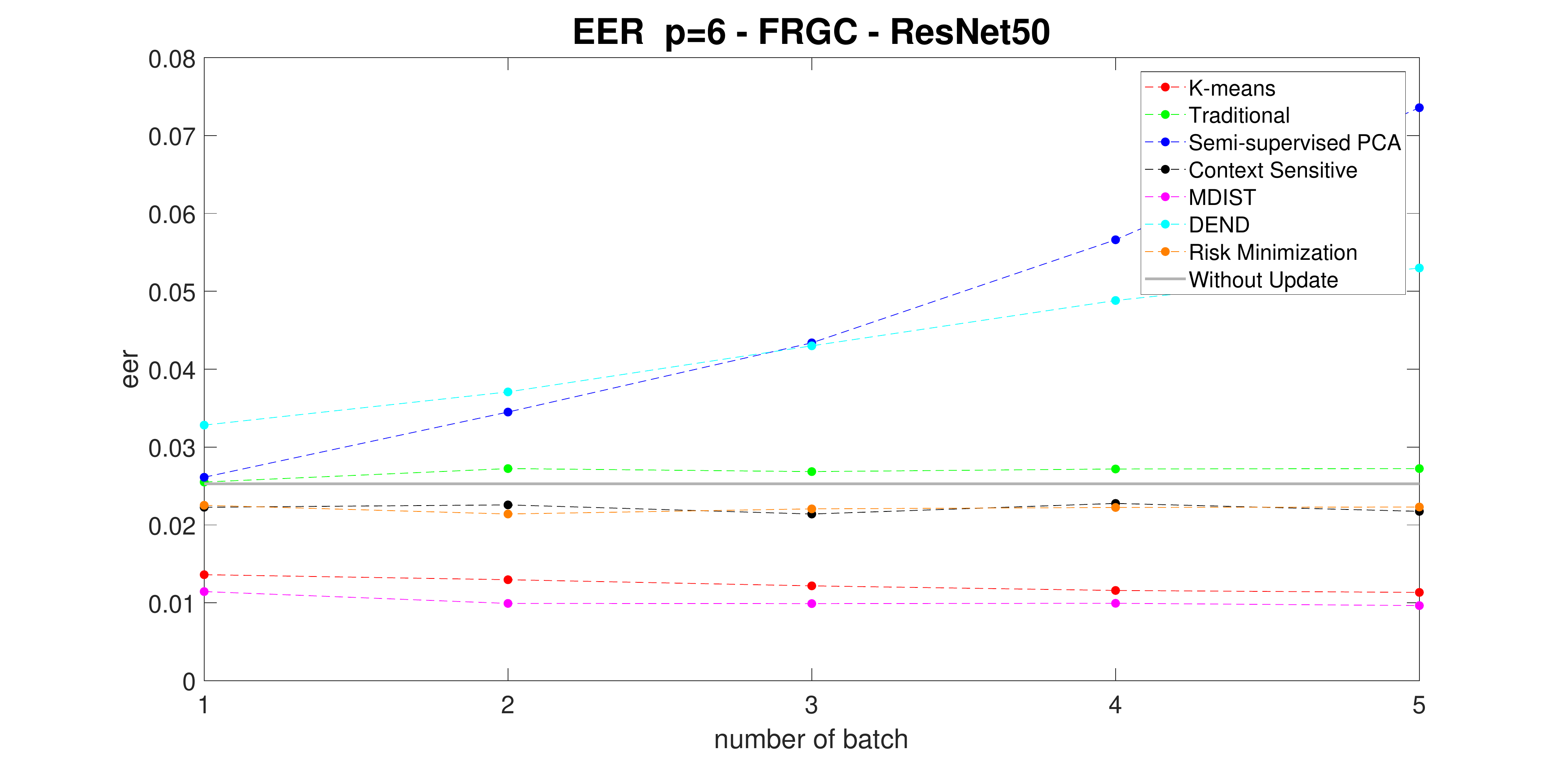}
}
\subfigure{
\includegraphics[scale=0.142]{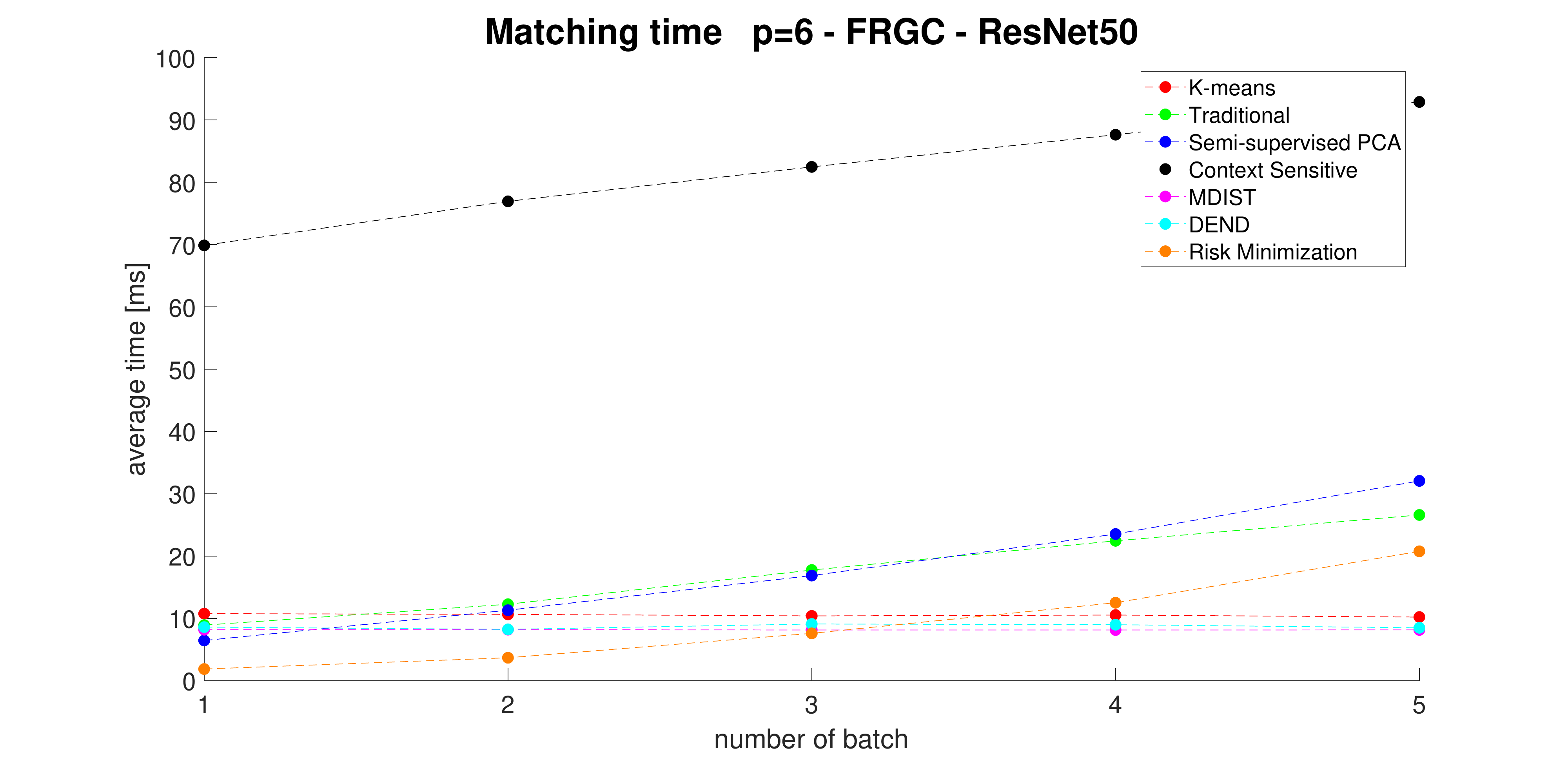}}
\caption{EER, percentage impostors and matching time comparison among the state of the art and the new proposed method with p=6 for FRGC using ResNet50 auto-encoded features. On the x axis is shown the number of the batch and on the y axis the performance index.}
\label{frgc6resnet}
\end{figure}

\begin{figure}[htbp]
\centering%
\subfigure{
\includegraphics[scale=0.142]{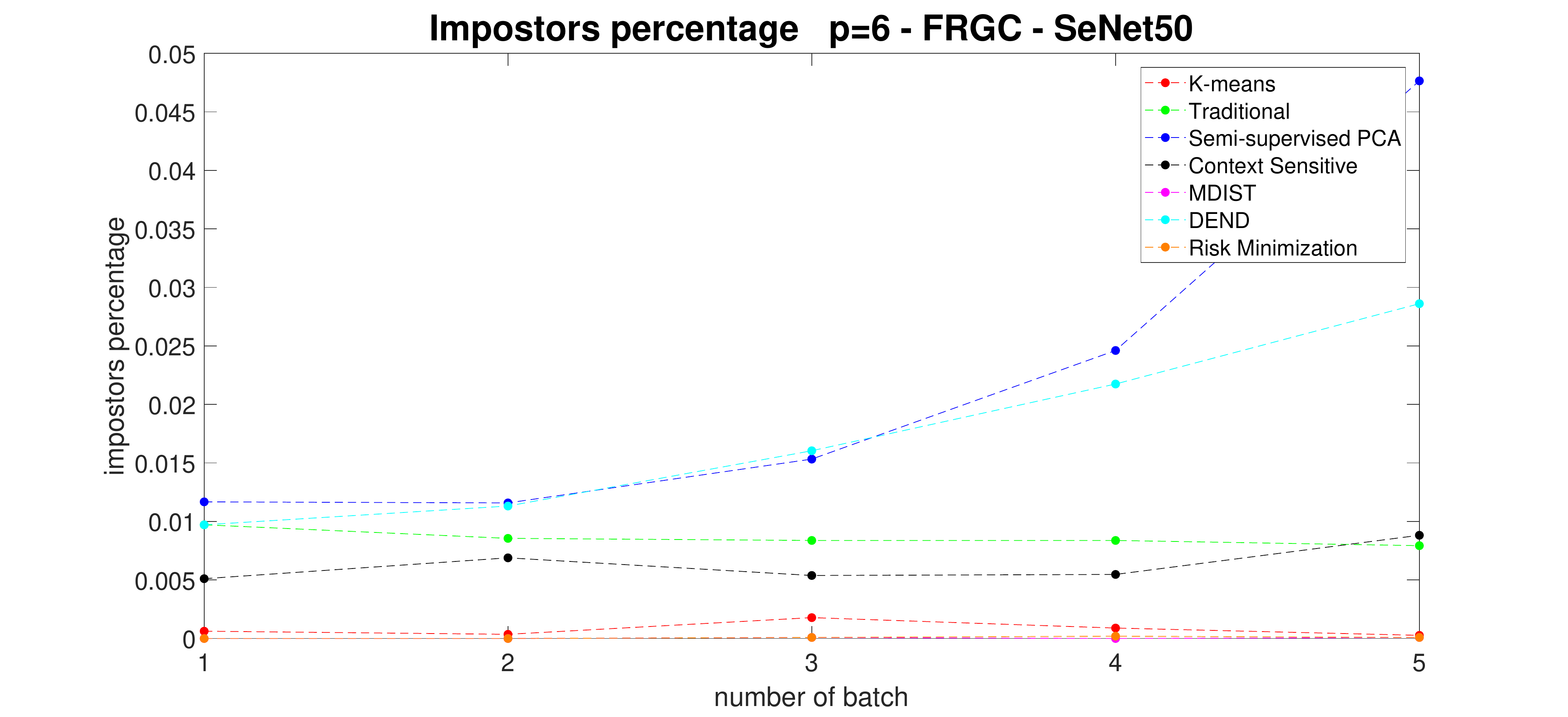}
}
\subfigure{
\includegraphics[scale=0.142]{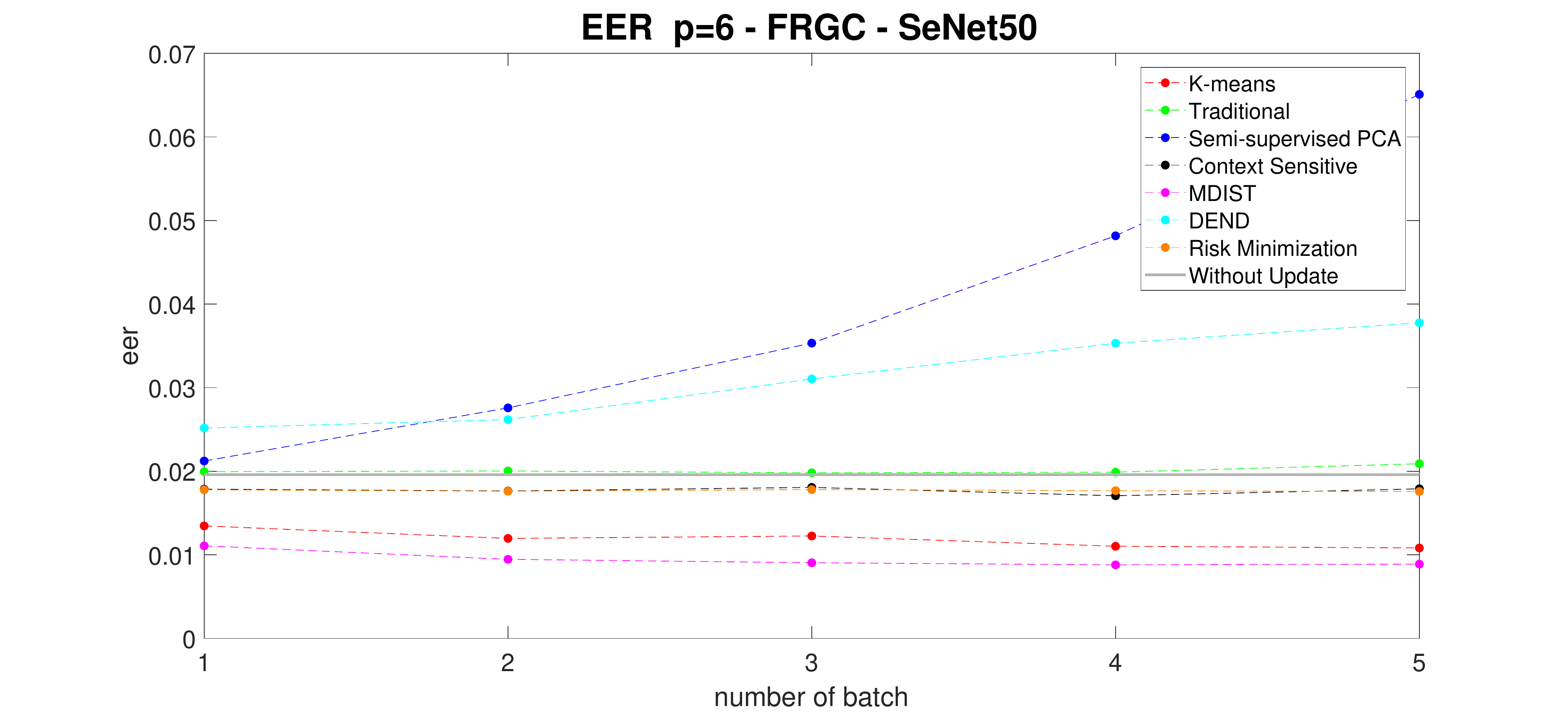}
}
\subfigure{
\includegraphics[scale=0.142]{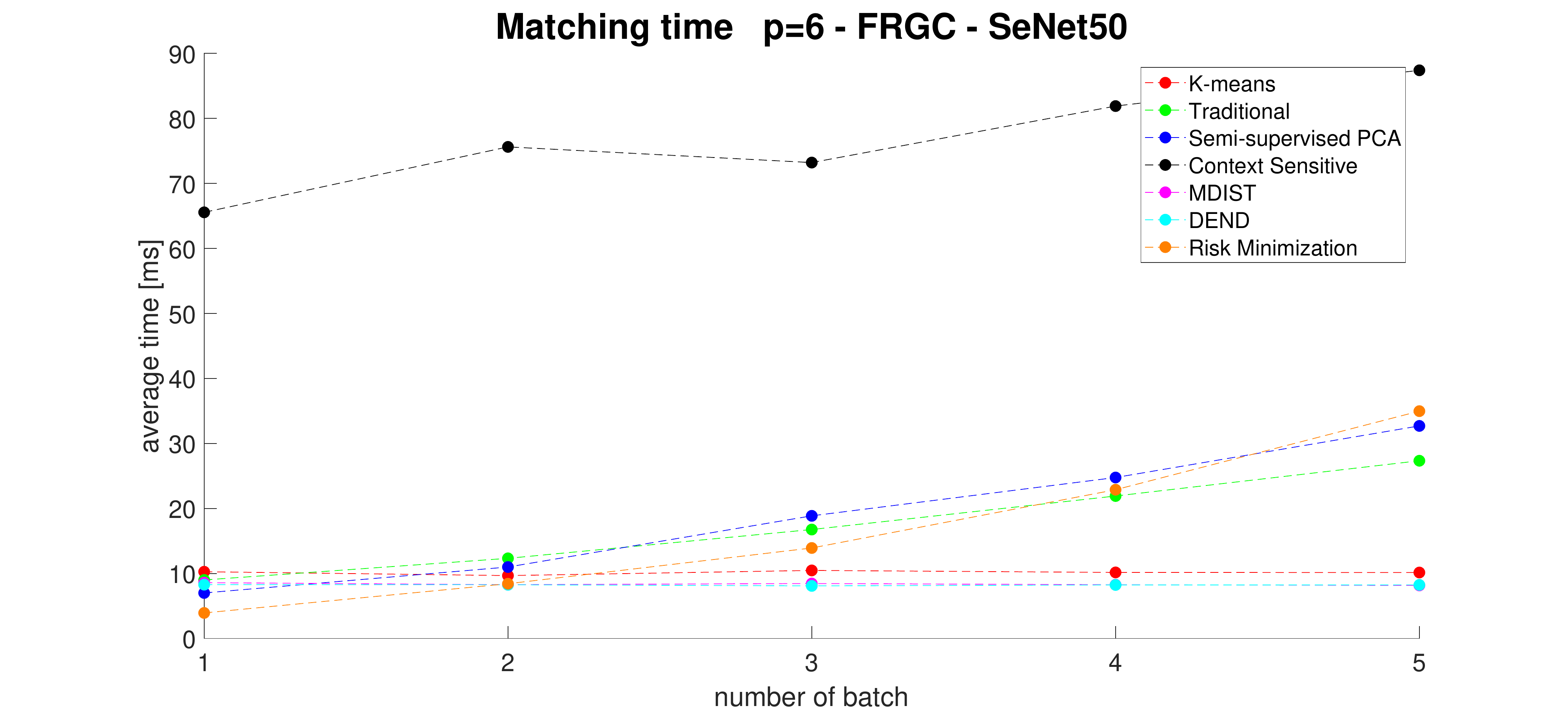}}
\caption{EER, percentage impostors and matching time comparison among the state of the art and the new proposed method with p=6 for FRGC using SeNet50 auto-encoded features. On the x axis is shown the number of the batch and on the y axis the performance index.}
\label{frgc6senet}
\end{figure}

The same metrics are reported in Figs. \ref{frgc6},\ref{frgc6face},\ref{frgc6resnet},\ref{frgc6senet} for the FRGC data set. 
The results confirm a low percentage of impostors and high performance for the proposed methods. As expected, the DEND method is not performing, as confirmation of results achieved on the Multimodal-DIEE. 

\begin{figure}[htbp]
\centering%
\subfigure{
\includegraphics[scale=0.142]{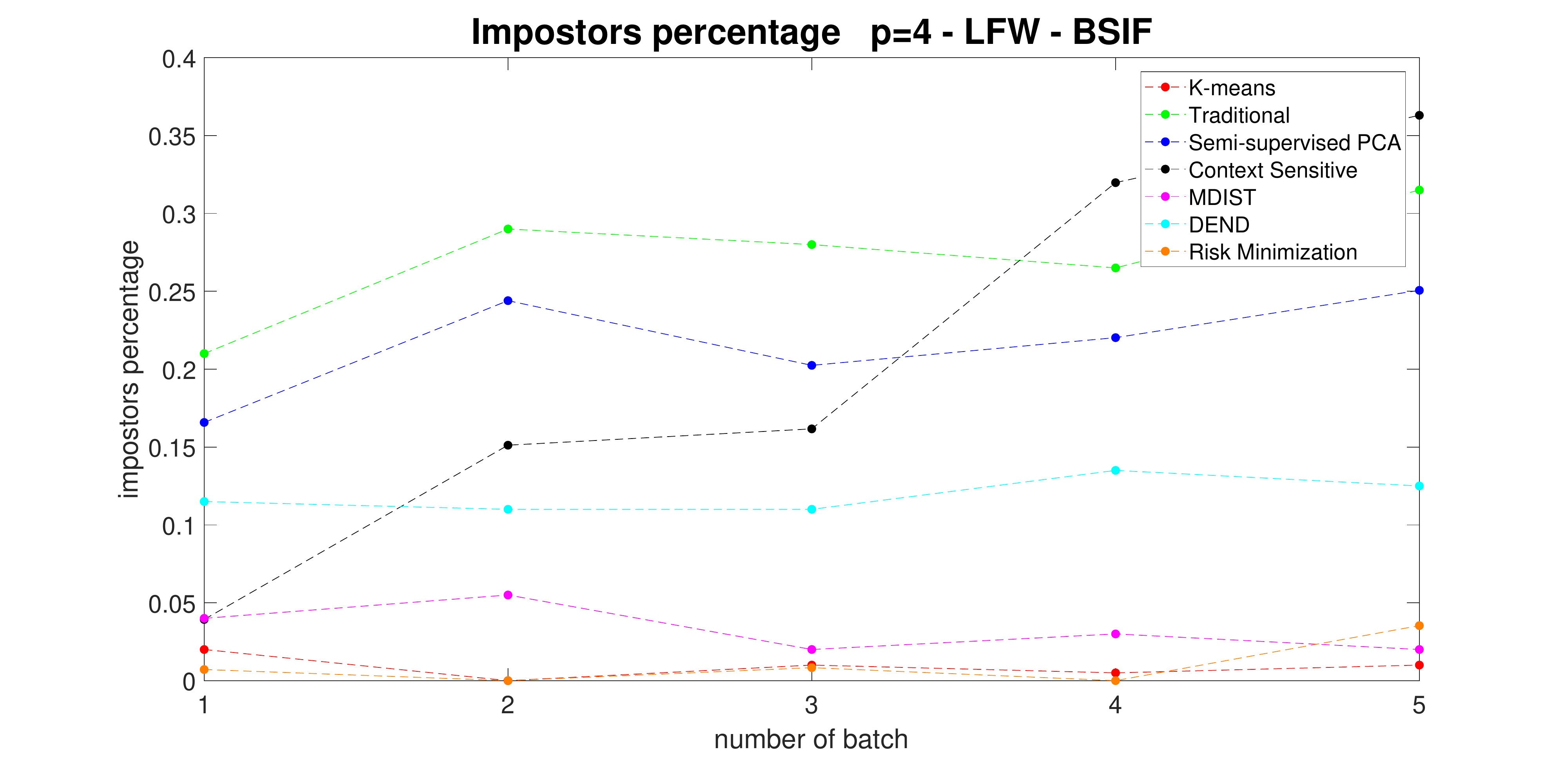}
}
\subfigure{
\includegraphics[scale=0.142]{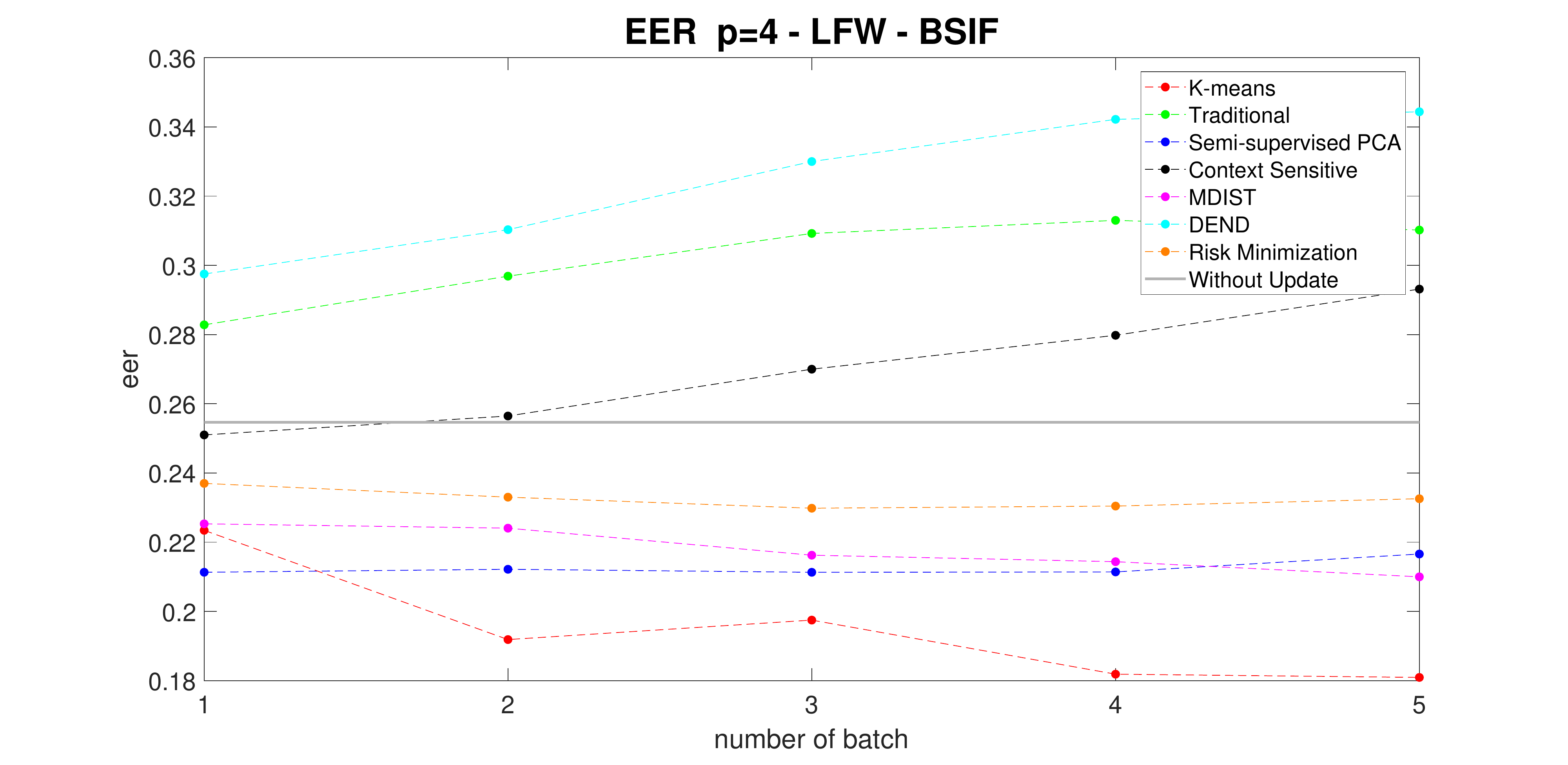}
}
\subfigure{
\includegraphics[scale=0.142]{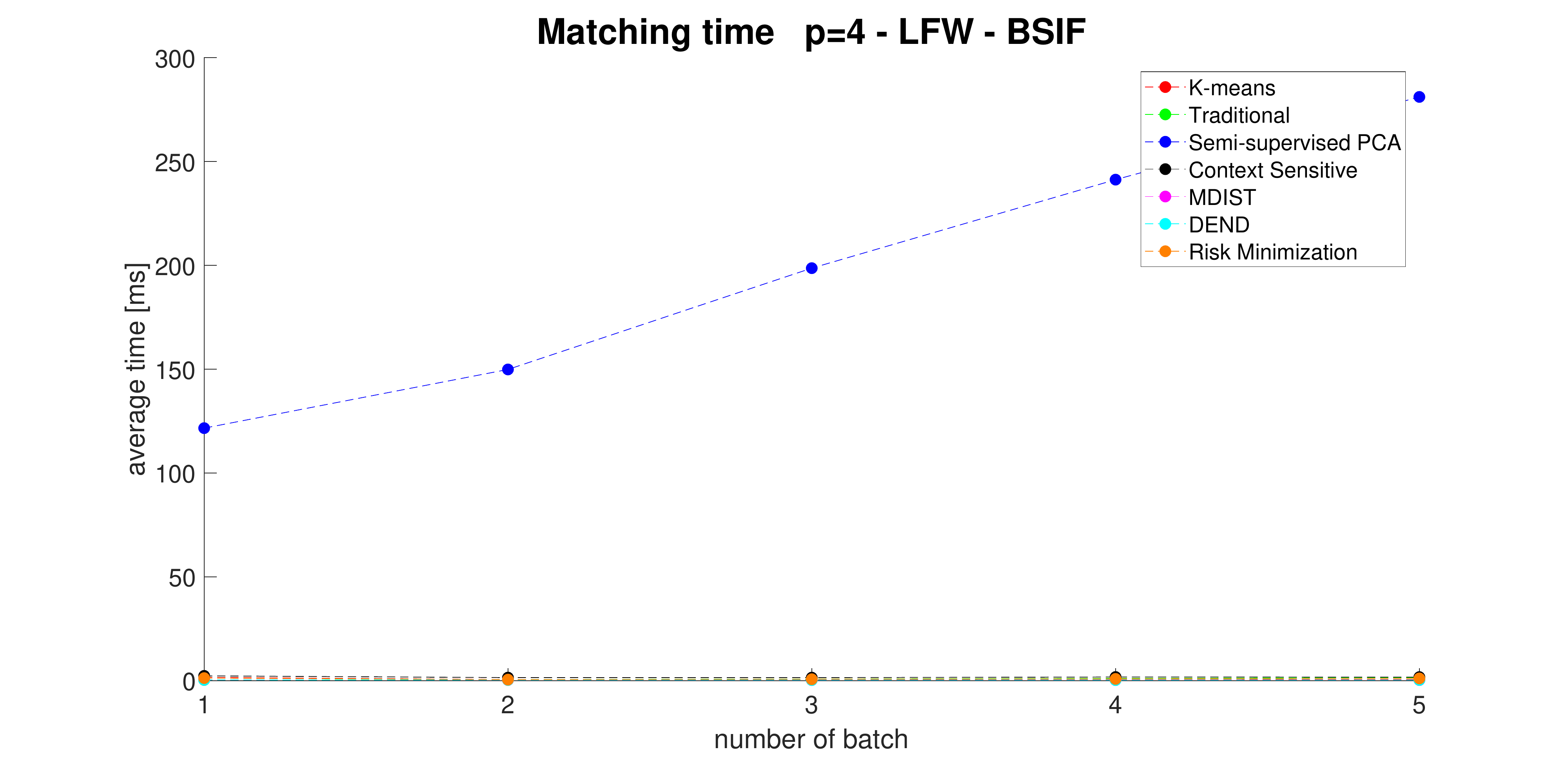}}
\subfigure{
\includegraphics[scale=0.142]{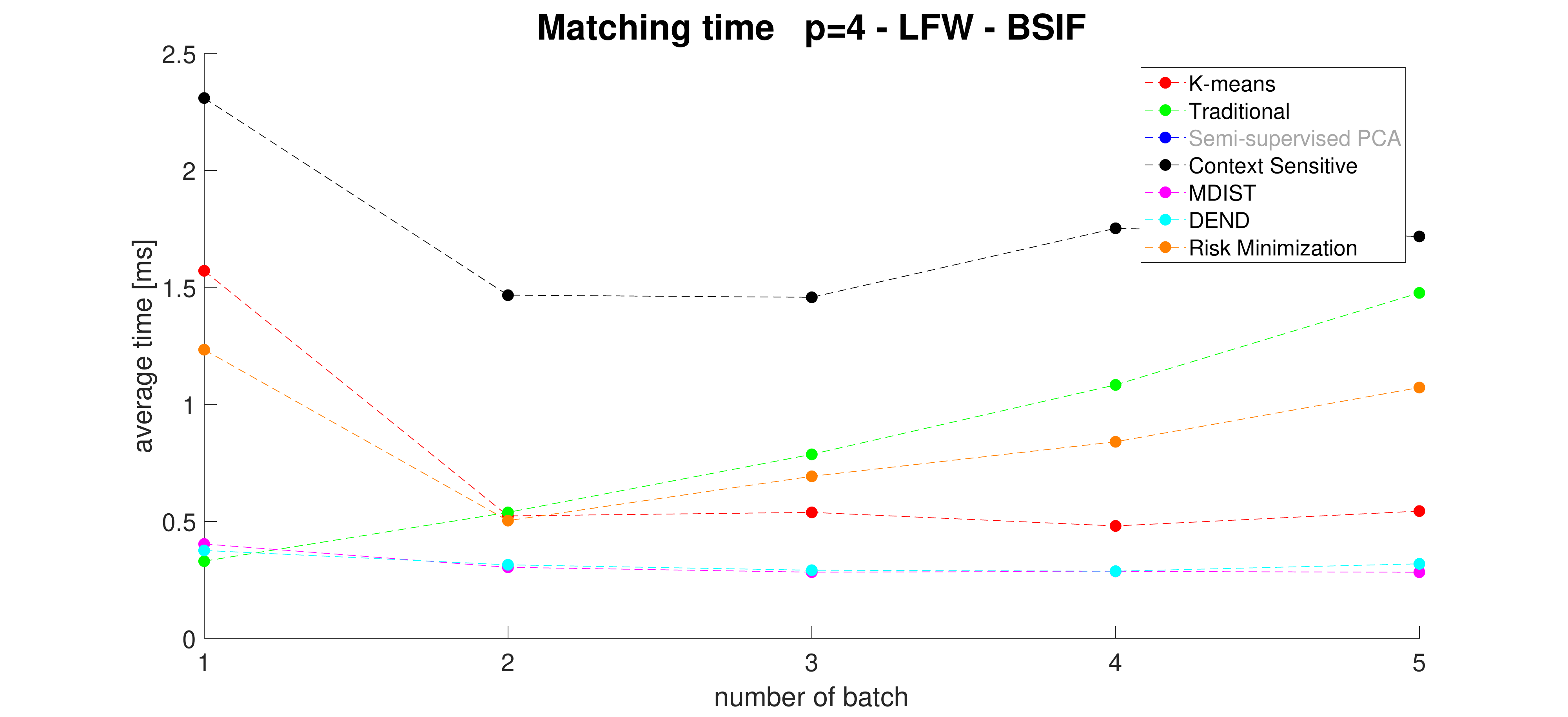}}
\caption{EER, percentage impostors and matching time comparison among the state of the art and the new proposed method with p=4 for LFW using BISF handcrafted features. On the x axis is shown the number of the batch and on the y axis the performance index.}
\label{lfw4}
\end{figure}

\begin{figure}[htbp]
\centering%
\subfigure{
\includegraphics[scale=0.142]{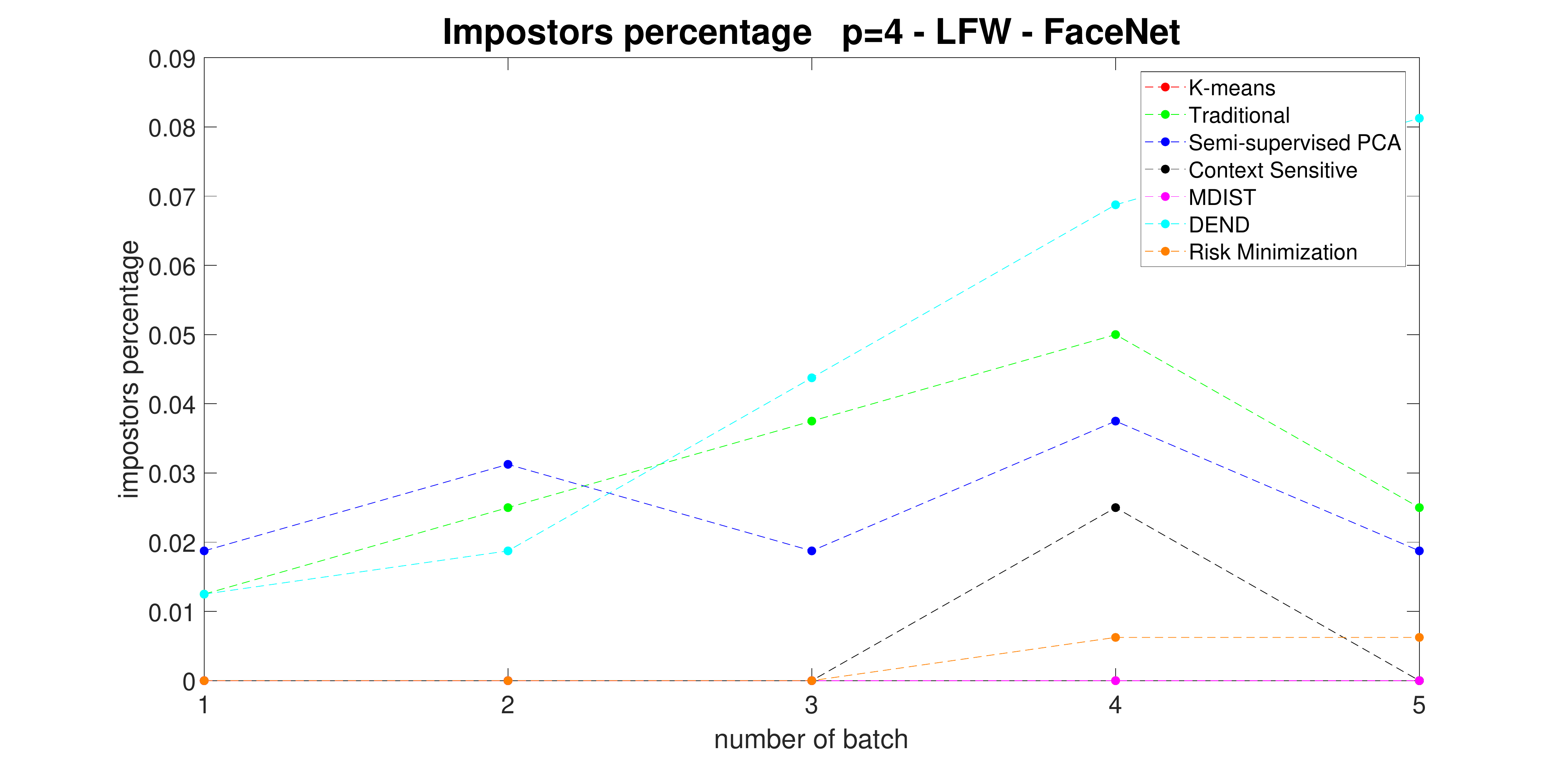}}
\subfigure{
\includegraphics[scale=0.142]{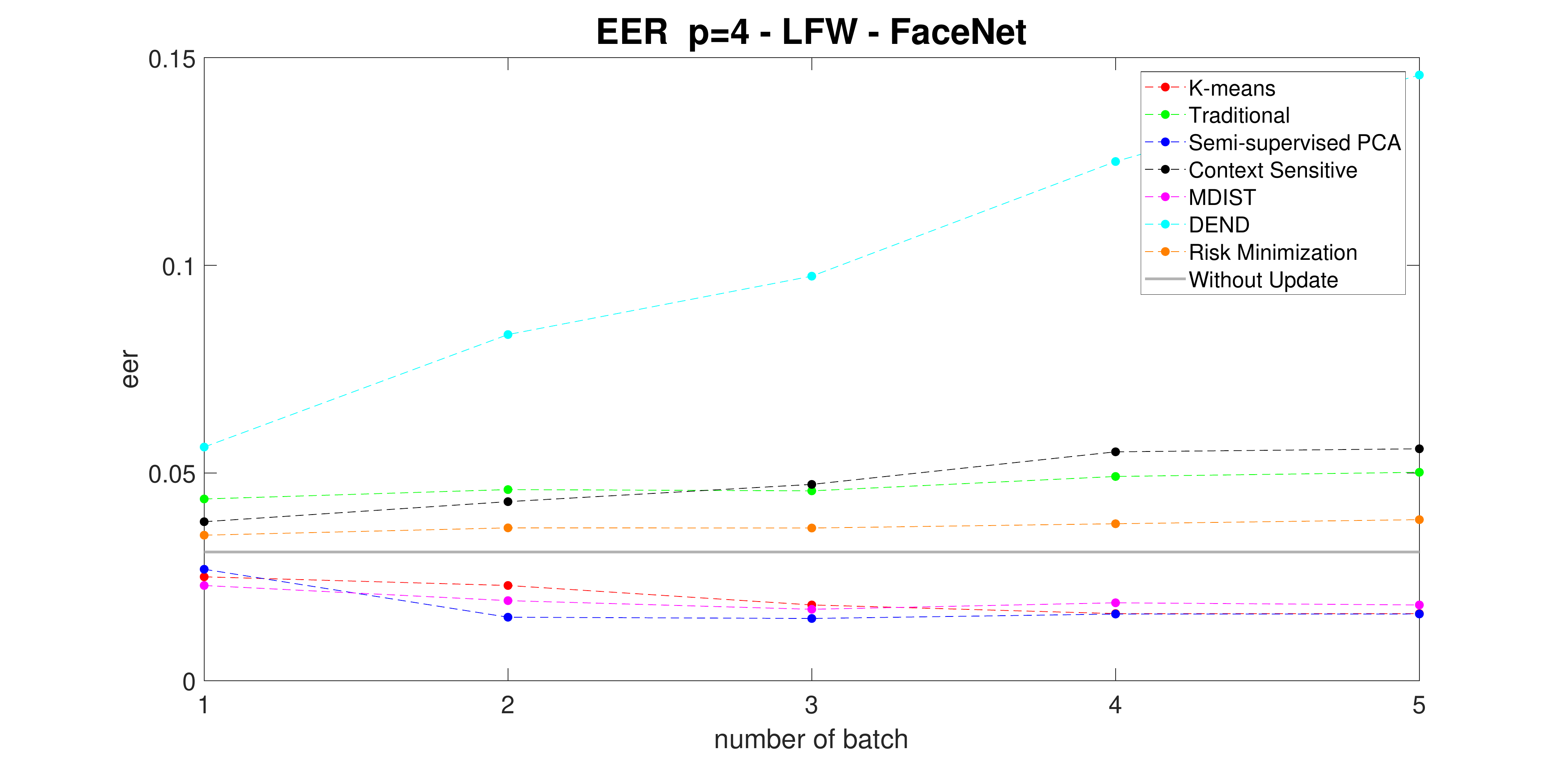}
}
\subfigure{
\includegraphics[scale=0.142]{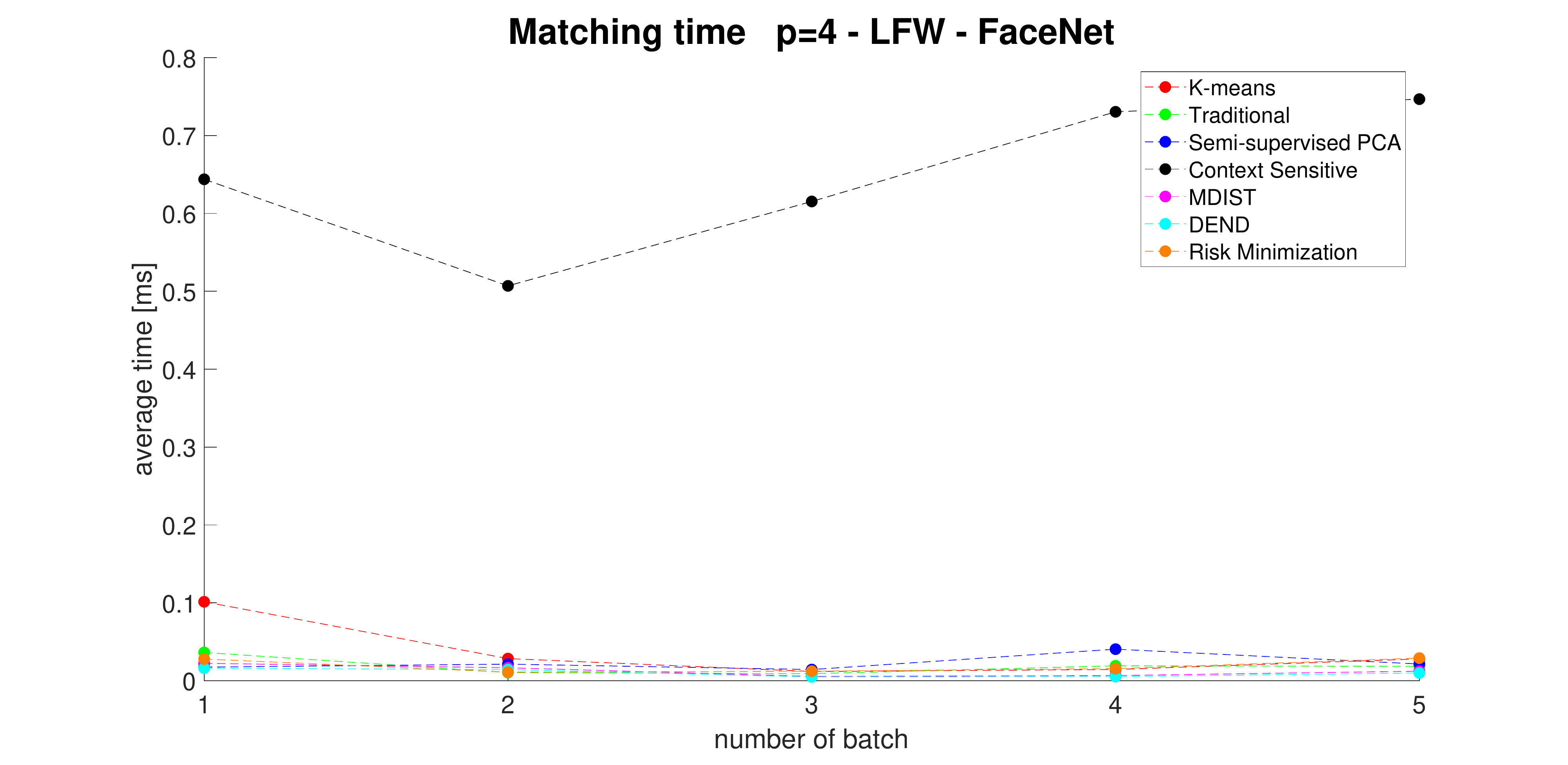}}
\caption{EER, percentage impostors and matching time comparison among the state of the art and the new proposed method with p=4 for LFW using FaceNet auto-encoded features. On the x axis is shown the number of the batch and on the y axis the performance index.}
\label{lfw4face}
\end{figure}

\begin{figure}[htbp]
\centering%
\subfigure{
\includegraphics[scale=0.142]{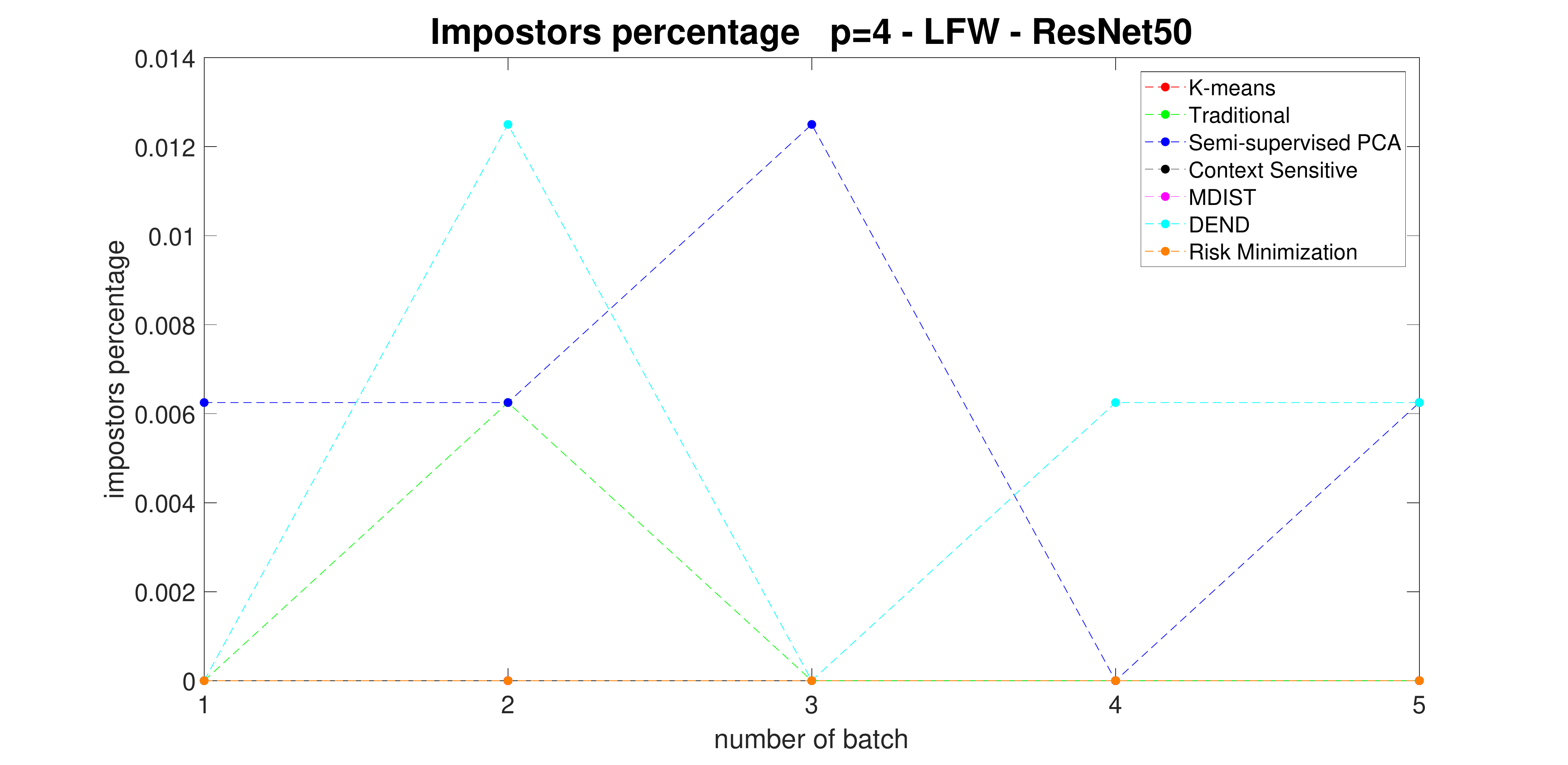}}
\subfigure{
\includegraphics[scale=0.142]{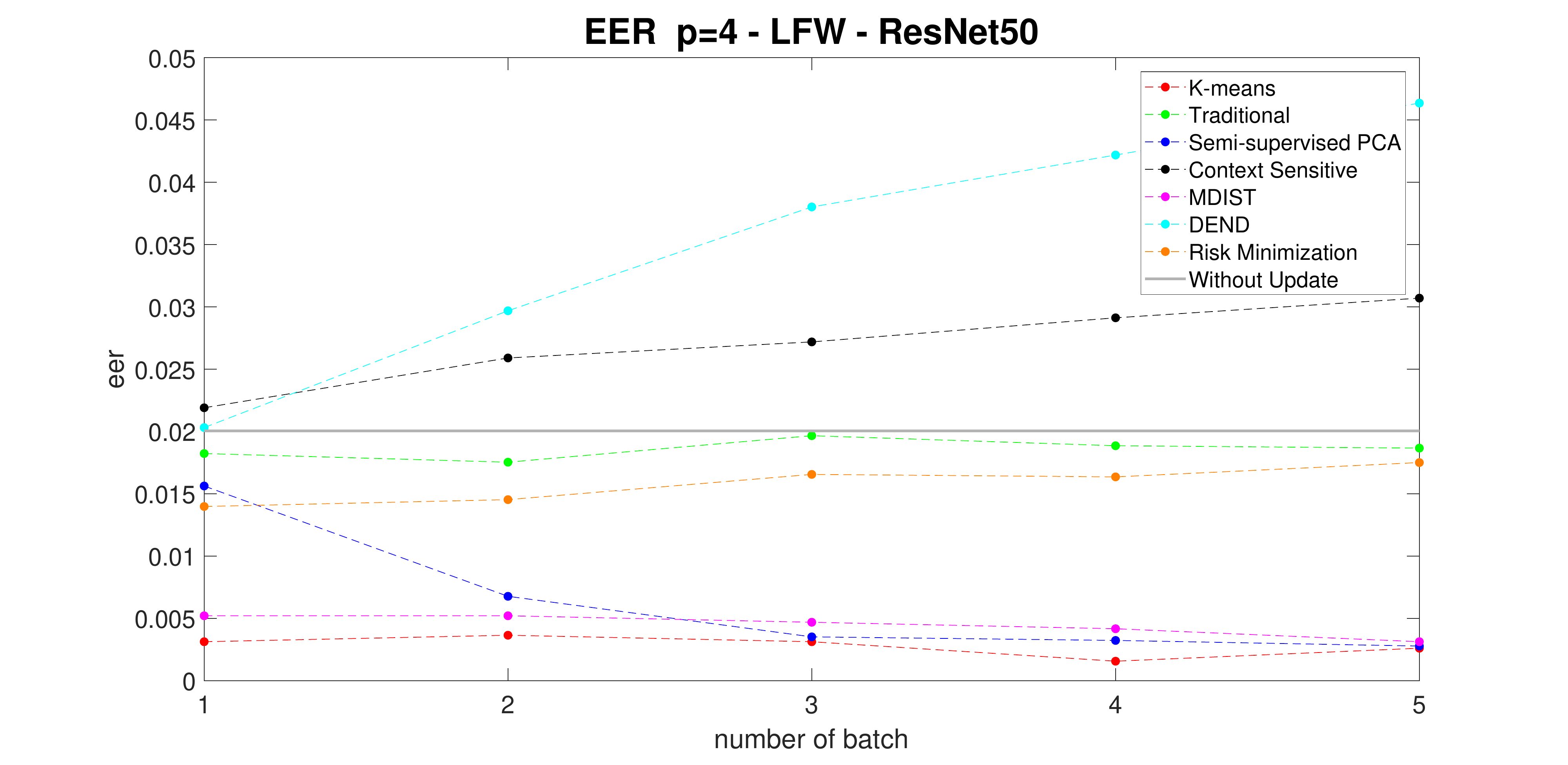}
}
\subfigure{
\includegraphics[scale=0.142]{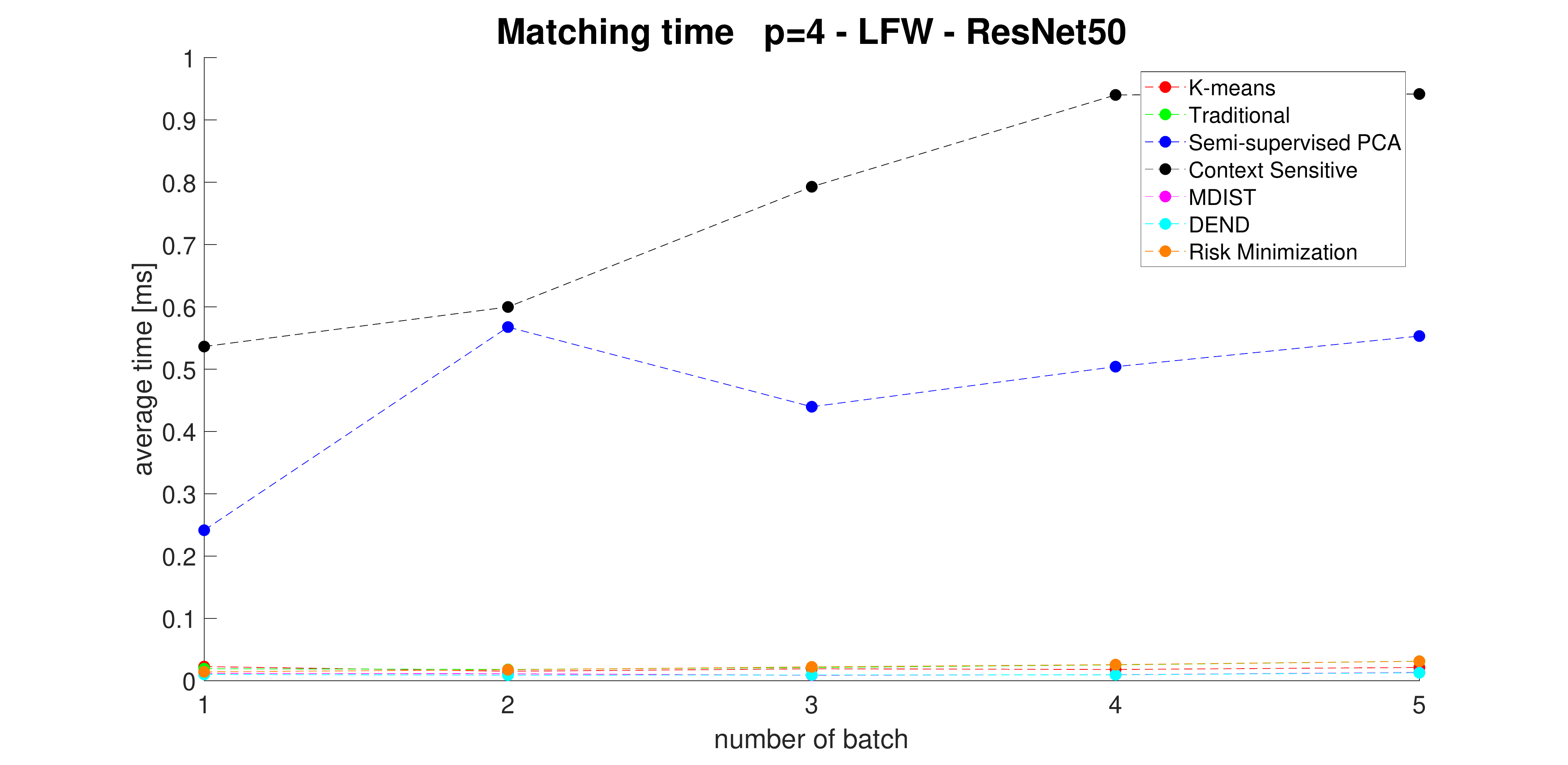}
}
\caption{EER, percentage impostors and matching time comparison among the state of the art and the new proposed method with p=4 for LFW using ResNet50 auto-encoded features. On the x axis is shown the number of the batch and on the y axis the performance index.}
\label{lfw4resnet}
\end{figure}

\begin{figure}[htbp]
\centering%
\subfigure{
\includegraphics[scale=0.142]{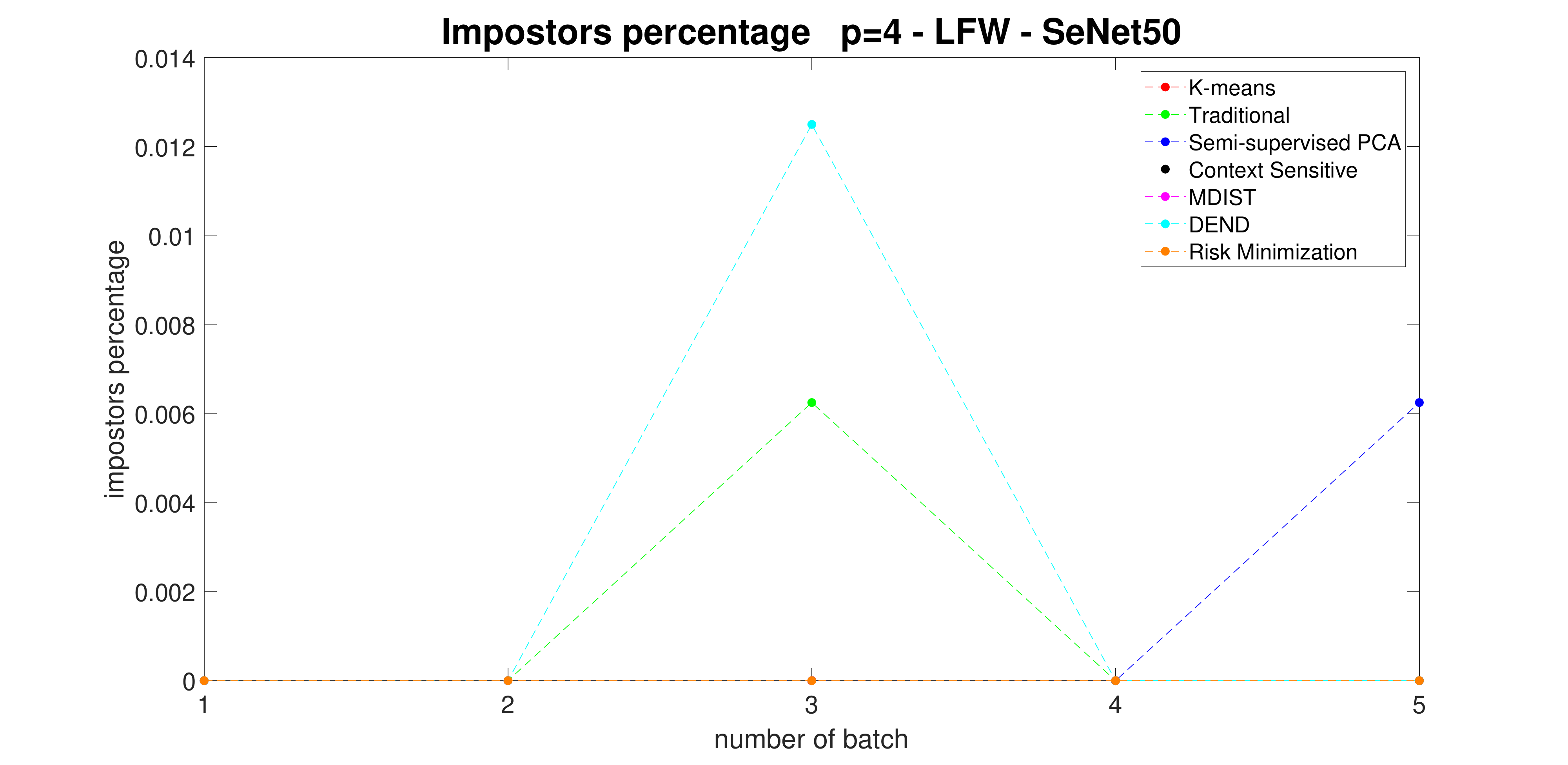}}
\subfigure{
\includegraphics[scale=0.142]{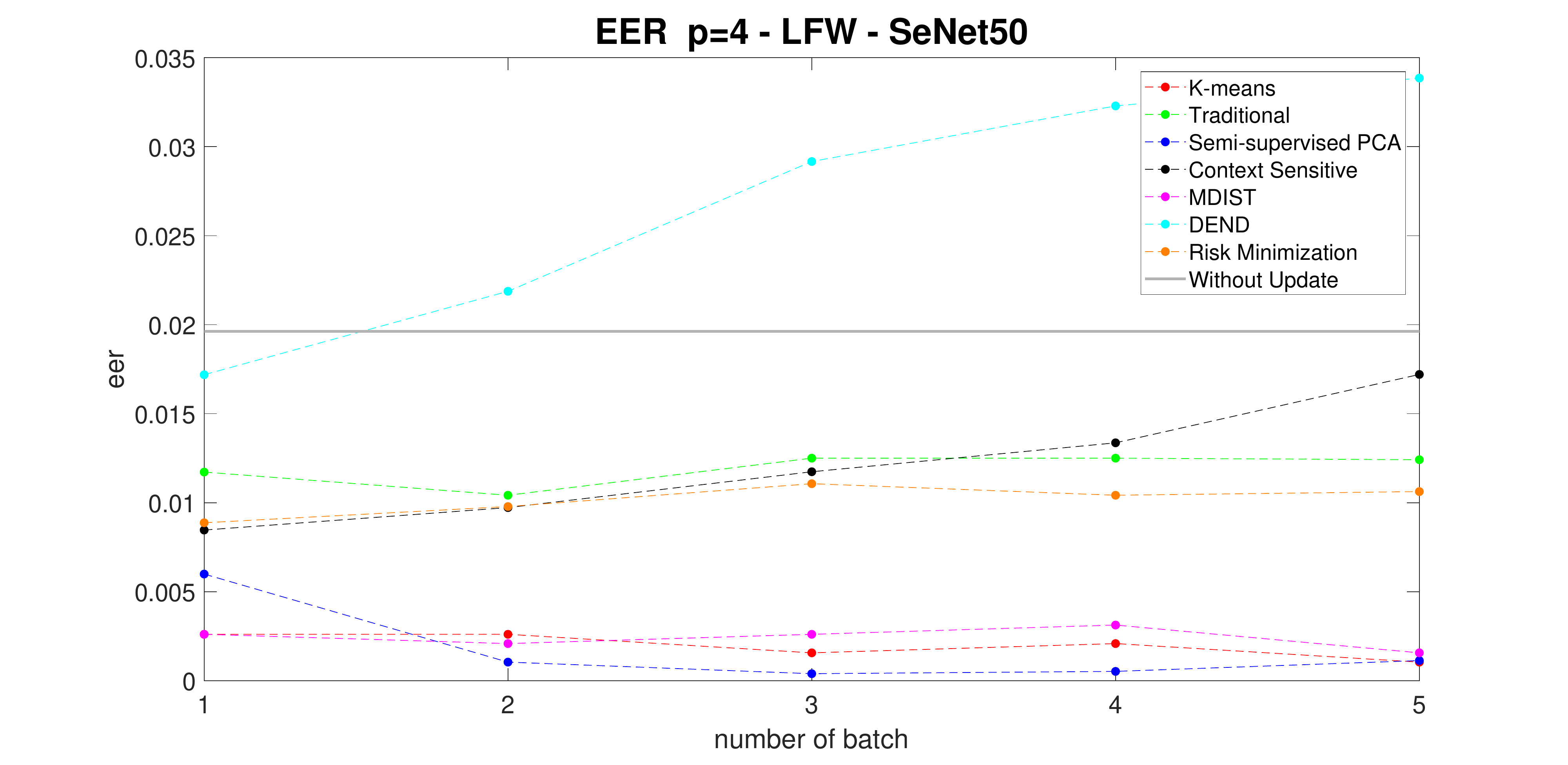}
}
\subfigure{
\includegraphics[scale=0.142]{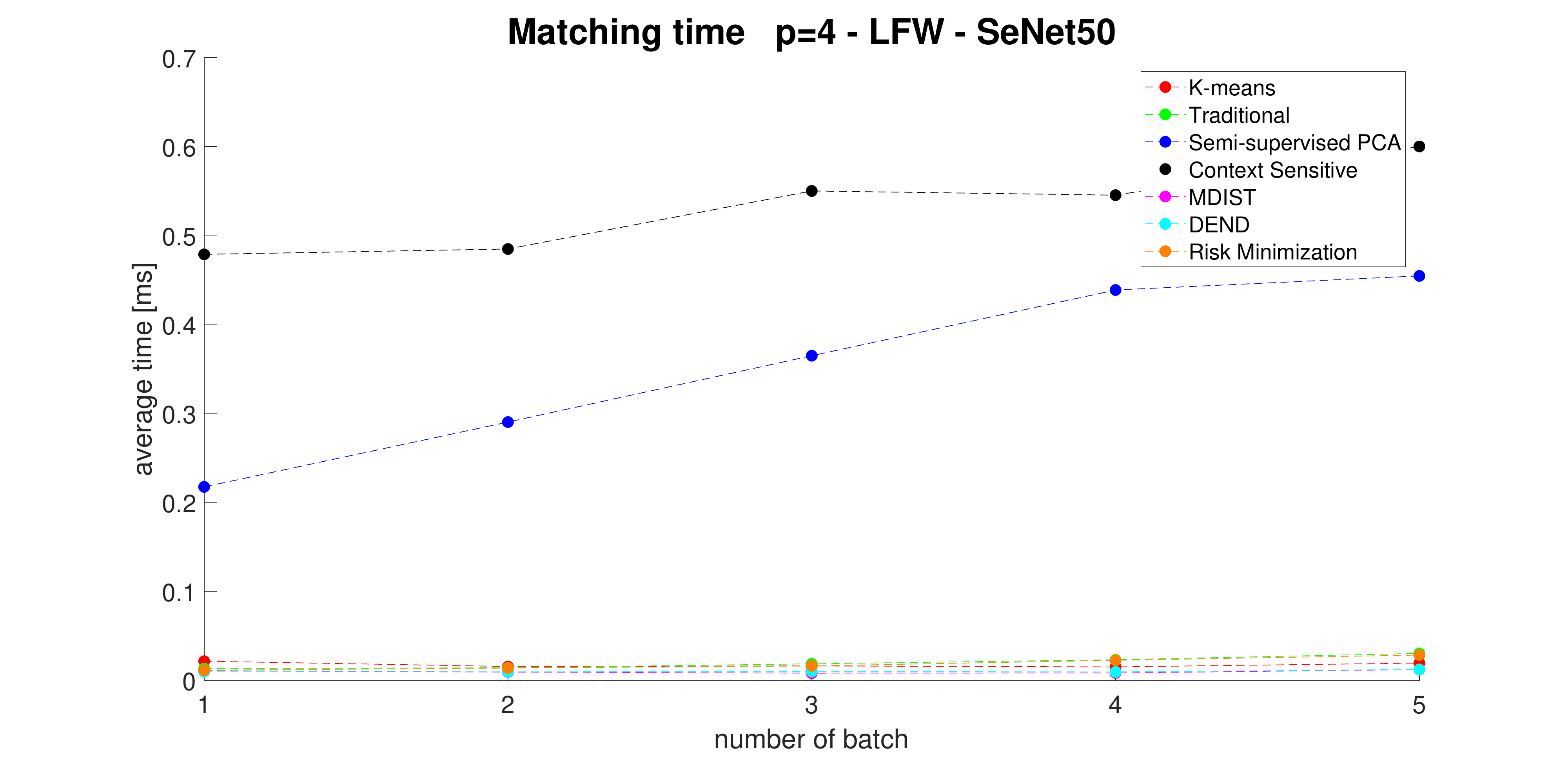}
}
\caption{EER, percentage impostors and matching time comparison among the state of the art and the new proposed method with p=4 for LFW using SeNet50 auto-encoded features. On the x axis is shown the number of the batch and on the y axis the performance index.}
\label{lfw4senet}
\end{figure}

In the case of the LFW data set, due to the small size of the subset, we set $p=4$ (Figs. \ref{lfw4},\ref{lfw4face},\ref{lfw4resnet},\ref{lfw4senet})

This data set presents a completely uncontrolled environment. As a matter of fact, a more significant discrepancy between handcrafted and auto-encoded features is pointed out.
In fact, we have a significant loss in performance for the BSIF features \ref{lfw4}. This drop does not occur with auto-encoded features. However, maybe due to the small user population, K-Means is still the best approach to self-updating for BSIF features. The environment represented by the LFW data set could be similar to that where snapshots are taken everywhere and stored in a mobile device for login by facial recognition. In this case, the reported results suggest that self-updating could help by reducing the cooperation degree of the user without the need of the explicit request of ``enroll's update''.

Here we want to remark that the EER values of the proposed methods are the only ones that never exceed the ``Without Update''. This shows that, although the LFW data set is a sort of hostile environment for self-updating, the hypothesis behind our model appears to be enough for guaranteeing the performance improvement and the good choice of the updated templates.
Even the percentage of impostors in the gallery for the K-Means method and the MDIST method is always close to zero for all batches.

For sake of completeness, we investigated the analysis of processing time of the proposed algorithms. Unlike the methods at the state of the art, we have a constant processing time at each update iteration. This is thanks to the limitation of $p$ templates in the selection phase. The template limit per user means that the storage size can not exceed $(p \cdot k \cdot S)$ B, where $p$ is the maximum number of template per user, $k$ is the number of users in the gallery and $S$ is the size in byte of each feature vector. This value is independent of the iterations number and it is constant after reaching the limit of $p$ template per user. In state-of-the-art classification-selection systems, the storage size depends on the iterations number $i$ and it is $(\beta \cdot i \cdot \bar{m} \cdot k \cdot S)$ B, where $\bar{m}$ is the average number of templates added per updating iteration and $\beta$ is a value between 0 and 1 that indicates the rate of selected templates, which is obviously less than the overall acceptance rate set in terms of the threshold $t^*$. Worth noting, $\beta=1$ for the traditional self-update systems. Without the limit in the number of templates per user, the required storage memory tends to $\infty$ with $i$.
For all the data sets, the processing time curve related to the PCA-based self-updating is not comparable as it refers to a supervised reduction of the feature vectors.

\section{Conclusions}
Compactness and expressiveness are the main characteristics of the most recent feature sets for encoding a facial image. However, they are not still able to embed all possible variations of the users' face. 
In this paper, a classification-selection approach with a maximum number of $p$ templates per user was proposed in order to keep limited the storage and computational requirements. The working hypothesis of this approach is that the statistical distribution of the facial features exhibits a dominating mode around which the templates can be searched.

The proposed method, implemented here in two different ways, showed an excellent performance. Results were confirmed by using two different kinds of state-of-the-art feature sets.
Three facial data sets were used as test. The face recognition performance was superior to that of other state-of-the-art approaches. 

We are aware that further theoretical and experimental investigations are required to draw definitive conclusions, especially concerning the role of the $p$ parameter with respect to the statistical distribution of samples around the hypothesised dominating mode.

However, we believe that a contribution like this was necessary by considering the panorama of the research community, mainly focused on designing deep-learning architectures for facial recognition much more accurate that 15 years ago systems but still affected by the well-known limitations and computationally expensive to be incrementally re-trained. The proposed self-updating methods showed there is no need of re-training for autoencoded features especially, which exhibited impressive results. Our self-updating methods are designed to be applied off-line; they are simple, computationally inexpensive with respect to the re-training of a deep network over the same amount of pseudo-labelled samples. The whole algorithms can be stored and implemented on device, without the need to send the data to a central mainframe for the updating computation. On the basis of these observations, the integration of self-updating algorithms in current face recognition systems is a realistic goal to be carried out.
\section*{Acknowledgement}
This work is supported by the Italian Ministry of Education, University and 
Research (MIUR) within the PRIN2017 - BullyBuster - A framework for bullying and cyberbullying action detection by computer vision and artificial intelligence 
methods and algorithms (CUP: F74I19000370001).

\bibliography{main}

\begin{thebibliography}{10}
\expandafter\ifx\csname url\endcsname\relax
  \def\url#1{\texttt{#1}}\fi
\expandafter\ifx\csname urlprefix\endcsname\relax\def\urlprefix{URL }\fi
\expandafter\ifx\csname href\endcsname\relax
  \def\href#1#2{#2} \def\path#1{#1}\fi

\bibitem{6403227}
N.~Poh, A.~Rattani, F.~Roli, Critical analysis of adaptive biometric systems,
  IET Biometrics 1~(4) (2012) 179--187.
\newblock \href {http://dx.doi.org/10.1049/iet-bmt.2012.0019}
  {\path{doi:10.1049/iet-bmt.2012.0019}}.

\bibitem{Uludag20041533}
U.~Uludag, A.~Ross, A.~Jain, Biometric template selection and update: a case
  study in fingerprints, Pattern Recognition 37~(7) (2004) 1533 -- 1542.
\newblock \href
  {http://dx.doi.org/https://doi.org/10.1016/j.patcog.2003.11.012}
  {\path{doi:https://doi.org/10.1016/j.patcog.2003.11.012}}.

\bibitem{Ryu2005}
C.~Ryu, Y.~Han, H.~Kim, {Super-template Generation Using Successive Bayesian
  Estimation for Fingerprint Enrollment}, Springer Berlin Heidelberg, 2005, pp.
  710--719.
\newblock \href {http://dx.doi.org/10.1007/11527923_74}
  {\path{doi:10.1007/11527923_74}}.

\bibitem{Jiang:2002:OFT:605089.605099}
X.~Jiang, W.~Ser, {Online Fingerprint Template Improvement}, IEEE Trans.
  Pattern Anal. Mach. Intell. 24~(8) (2002) 1121--1126.
\newblock \href {http://dx.doi.org/10.1109/TPAMI.2002.1023807}
  {\path{doi:10.1109/TPAMI.2002.1023807}}.

\bibitem{8114708}
V.~Sze, Y.~H. Chen, T.~J. Yang, J.~S. Emer, Efficient processing of deep neural
  networks: A tutorial and survey, Proceedings of the IEEE ,Vol. 105 (2017)
  2295--2329.
\newblock \href {http://dx.doi.org/10.1109/JPROC.2017.2761740}
  {\path{doi:10.1109/JPROC.2017.2761740}}.

\bibitem{facenet}
F.~Schroff, D.~Kalenichenko, J.~Philbin, Facenet: A unified embedding for face
  recognition and clustering., CoRR abs/1503.03832.

\bibitem{8338413}
G.~Mai, K.~Cao, P.~C. YUEN, A.~K. Jain, On the reconstruction of face images
  from deep face templates, IEEE Transactions on Pattern Analysis and Machine
  Intelligence (2018) 1--1\href {http://dx.doi.org/10.1109/TPAMI.2018.2827389}
  {\path{doi:10.1109/TPAMI.2018.2827389}}.

\bibitem{Balcan2005PersonII}
M.-F. Balcan, A.~Blum, P.~P. Choi, J.~D. Lafferty, B.~Pantano, M.~Robert,
  X.~Zhu, J.~Lafferty, M.~R. Rwebangira, Person identification in webcam
  images: An application of semi-supervised learning, in: ICML 2005 Workshop on
  Learning with Partially Classified Training Data, 2005.

\bibitem{Roli2006}
F.~Roli, G.~L. Marcialis, {Semi-supervised PCA-Based Face Recognition Using
  Self-training}, Springer Berlin Heidelberg, 2006, pp. 560--568.
\newblock \href {http://dx.doi.org/10.1007/11815921_61}
  {\path{doi:10.1007/11815921_61}}.

\bibitem{1699908}
C.~Ryu, H.~Kim, A.~K.~Jain, Template adaptation based fingerprint verification,
  Vol.~4, 2006, pp. 582--585.
\newblock \href {http://dx.doi.org/10.1109/ICPR.2006.1105}
  {\path{doi:10.1109/ICPR.2006.1105}}.

\bibitem{Freni2008}
B.~Freni, G.~L. Marcialis, F.~Roli, {Replacement Algorithms for Fingerprint
  Template Update}, in: Proceedings of the 5th International Conference on
  Image Analysis and Recognition, ICIAR '08, Springer-Verlag, 2008, pp.
  884--893.
\newblock \href {http://dx.doi.org/10.1007/978-3-540-69812-8_88}
  {\path{doi:10.1007/978-3-540-69812-8_88}}.

\bibitem{Pagano2015}
C.~Pagano, E.~Granger, R.~Sabourin, P.~Tuveri, G.~L. Marcialis, F.~Roli,
  {Context-Sensitive Self-Updating for Adaptive Face Recognition}, Springer
  International Publishing, Cham, 2015, pp. 9--34.
\newblock \href {http://dx.doi.org/10.1007/978-3-319-24865-3_2}
  {\path{doi:10.1007/978-3-319-24865-3_2}}.

\bibitem{1247}
P.~Tuveri, V.~Mura, G.~L. Marcialis, F.~Roli, A classification-selection
  approach for self updating of face verification systems under stringent
  storage and computational requirements, in: 18th IAPR Int. Conf. on Image
  Analysis and Processing (ICIAP), Vol. 9280, 2015, pp. 540--550.
\newblock \href {http://dx.doi.org/10.1007/978-3-319-23234-8 50}
  {\path{doi:10.1007/978-3-319-23234-8 50}}.

\bibitem{RattaniDual}
A.~Rattani, G.~L. Marcialis, E.~Granger, F.~Roli, A dual-staged
  classification-selection approach for automated update of biometric
  templates, in: Proceedings of the 21st International Conference on Pattern
  Recognition (ICPR2012), 2012, pp. 2972--2975.

\bibitem{Rattani2013AMD}
A.~Rattani, G.~L. Marcialis, F.~Roli, A multi-modal dataset, protocol and tools
  for adaptive biometric systems: a benchmarking study, International Journal
  of Biometrics , Vol.5 (2013) 266--287.

\bibitem{Roli2008}
F.~Roli, L.~Didaci, G.~L. Marcialis, {Adaptive Biometric Systems That Can
  Improve with Use}, Springer London, 2008, pp. 447--471.
\newblock \href {http://dx.doi.org/10.1007/978-1-84628-921-7_23}
  {\path{doi:10.1007/978-1-84628-921-7_23}}.

\bibitem{Liu20031945}
X.~Liu, T.~Chen, S.~M. Thornton, Eigenspace updating for non-stationary process
  and its application to face recognition, Pattern Recognition , Vol.36~(9)
  (2003) 1945 -- 1959.

\bibitem{IncrPcaZhao}
H.~Zhao, P.~C. Yuen, J.~T. Kwok, J.~Yang, Incremental {PCA} based face
  recognition, in: ICARCV 2004 8th Control, Automation, Robotics and Vision
  Conference, 2004, pp. 687--691 Vol. 1.
\newblock \href {http://dx.doi.org/10.1109/ICARCV.2004.1468910}
  {\path{doi:10.1109/ICARCV.2004.1468910}}.

\bibitem{eigenFace}
M.~A. Turk, A.~P. Pentland, Face recognition using eigenfaces, in: Proceedings.
  1991 IEEE Computer Society Conference on Computer Vision and Pattern
  Recognition, 1991, pp. 586--591.
\newblock \href {http://dx.doi.org/10.1109/CVPR.1991.139758}
  {\path{doi:10.1109/CVPR.1991.139758}}.

\bibitem{Phillips:2005:OFR:1068507.1069015}
P.~J. Phillips, P.~J. Flynn, T.~Scruggs, K.~W. Bowyer, J.~Chang, K.~Hoffman,
  J.~Marques, J.~Min, W.~Worek, {Overview of the Face Recognition Grand
  Challenge}, in: Proceedings of the 2005 IEEE Computer Society Conference on
  Computer Vision and Pattern Recognition - Volume 1, CVPR 2005, IEEE Computer
  Society, pp. 947--954.
\newblock \href {http://dx.doi.org/10.1109/CVPR.2005.268}
  {\path{doi:10.1109/CVPR.2005.268}}.

\bibitem{BOONGOEN20181}
T.~Boongoen, N.~Iam-On, Cluster ensembles: A survey of approaches with recent
  extensions and applications, Computer Science Review 28 (2018) 1 -- 25.
\newblock \href
  {http://dx.doi.org/https://doi.org/10.1016/j.cosrev.2018.01.003}
  {\path{doi:https://doi.org/10.1016/j.cosrev.2018.01.003}}.

\bibitem{6762944}
J.~{Yu}, Y.~{Rui}, D.~{Tao}, Click prediction for web image reranking using
  multimodal sparse coding, IEEE Transactions on Image Processing 23~(5) (2014)
  2019--2032.
\newblock \href {http://dx.doi.org/10.1109/TIP.2014.2311377}
  {\path{doi:10.1109/TIP.2014.2311377}}.

\bibitem{LUMINI2006495}
A.~Lumini, L.~Nanni, A clustering method for automatic biometric template
  selection, Pattern Recognition 39~(3) (2006) 495 -- 497.
\newblock \href
  {http://dx.doi.org/https://doi.org/10.1016/j.patcog.2005.11.004}
  {\path{doi:https://doi.org/10.1016/j.patcog.2005.11.004}}.

\bibitem{LFWTech}
G.~B. Huang, M.~Ramesh, T.~Berg, E.~Learned-Miller, {Labeled Faces in the Wild:
  A Database for Studying Face Recognition in Unconstrained Environments},
  Tech. Rep. 07-49, University of Massachusetts, Amherst (Oct 2007).

\bibitem{BSIF}
J.~Kannala, E.~Rahtu, {BSIF}: Binarized statistical image features, in:
  Proceedings of the 21st International Conference on Pattern Recognition
  (ICPR2012), 2012, pp. 1363--1366.

\bibitem{He2015DeepRL}
K.~He, X.~Zhang, S.~Ren, J.~Sun, Deep residual learning for image recognition,
  2016 IEEE Conference on Computer Vision and Pattern Recognition (CVPR) (2015)
  770--778.

\bibitem{dog}
X.~Tan, B.~Triggs, {Enhanced Local Texture Feature Sets for Face Recognition
  Under Difficult Lighting Conditions}, IEEE Transactions on Image Processing
  19~(6) (2010) 1635--1650.
\newblock \href {http://dx.doi.org/10.1109/TIP.2010.2042645}
  {\path{doi:10.1109/TIP.2010.2042645}}.

\bibitem{guo2016msceleb}
Y.~Guo, L.~Zhang, Y.~Hu, X.~He, J.~Gao, M{S}-{C}eleb-1{M}: A dataset and
  benchmark for large scale face recognition, in: European Conference on
  Computer Vision, Springer, 2016.

\bibitem{Cao2017VGGFace2AD}
Q.~Cao, L.~Shen, W.~Xie, O.~M. Parkhi, A.~Zisserman, Vggface2: A dataset for
  recognising faces across pose and age, 2018 13th IEEE International
  Conference on Automatic Face \& Gesture Recognition (FG 2018) (2017) 67--74.

\bibitem{Hu2017SqueezeandExcitationN}
J.~Hu, L.~Shen, G.~Sun, Squeeze-and-excitation networks, 2018 IEEE/CVF
  Conference on Computer Vision and Pattern Recognition (2017) 7132--7141.

\end{thebibliography}

\end{document}